\documentclass[10pt]{article} % Comment this line out if you need a4paper

\usepackage{scicite}

\usepackage{graphics} % for pdf, bitmapped graphics files
\usepackage{epsfig} % for postscript graphics files
\usepackage{mathptmx} % assumes new font selection scheme installed
\usepackage{times} % assumes new font selection scheme installed
\usepackage{amsmath} % assumes amsmath package installed
\usepackage{amssymb}  % assumes amsmath package installed
\usepackage{xcolor}
\usepackage{algorithm2e}
\makeatletter
\renewcommand{\@algocf@capt@plain}{above}
\renewcommand{\algocf@caption@plain}{\box\algocf@capbox\vskip\AlCapSkip}%
\makeatother

\usepackage{balance}
\usepackage[utf8]{inputenc}
\usepackage[pdfauthor={Bonetto et al.},
            pdftitle={GRADE: Generating Realistic Animated Dynamic Environments},
            pdfsubject={Dynamic environments generation pipeline},
            pdfkeywords={slam,mapping,localization,learning,dynamic,animations,isaac,dynamic environments,dataset}]{hyperref}

\usepackage{enumitem}
\usepackage{subcaption}
\usepackage{cuted}
\usepackage{multirow}
\usepackage{rotating}

\usepackage{environ}
\usepackage{pdflscape}

\usepackage[textsize=tiny]{todonotes}

\hypersetup{
    colorlinks=true,
    linkcolor=blue,
    filecolor=magenta,      
    urlcolor=blue,
}
\usepackage{sepfootnotes}
\usepackage[font=footnotesize,labelfont=bf]{caption}
\NewEnviron{nbp}[1]{%
  \xdef\nbptemp{{#1}{\unexpanded\expandafter{\BODY}}}%
  \aftergroup\donpb
}
\newcommand{\donpb}{\expandafter\sepfootnotecontent\nbptemp}

\usepackage{tikz}
\def\tick{\tikz\fill[scale=0.4](0,.35) -- (.25,0) -- (1,.7) -- (.25,.15) -- cycle;}

\newcommand{\uproman}[1]{\uppercase\expandafter{\romannumeral#1}}

\usepackage{soul}
\usepackage{array}
\newcolumntype{?}{!{\vrule width 2pt}}
\usepackage{booktabs}

\newenvironment{sciabstract}{%
\begin{quote} \bf}
{\end{quote}}

\title{\LARGE\bf{GRADE: Generating Realistic and Dynamic Environments for Robotics Research with IsaacSim}}

\author
{Elia Bonetto$^{1,2}$, Chenghao Xu$^{3,1}$, and Aamir Ahmad$^{2,1}$\\
\\
\normalsize{$^1$Max Planck Institute for Intelligent Systems, Tübingen, Germany.}\\
\normalsize{$^2$Institute of Flight Mechanics and Controls, University of Stuttgart, Stuttgart, Germany.}\\
\normalsize{$^3$Swiss Federal Institute of Technology Lausanne (EPFL), Lausanne, Switzerland.}
\\
\normalsize{$^\ast$To whom correspondence should be addressed; E-mail:  elia.bonetto@tue.mpg.de.}
}

\begin{document}

	\maketitle
    \begin{sciabstract}

Synthetic data and novel rendering techniques have greatly influenced computer vision research in tasks like target tracking and human pose estimation. However, robotics research has lagged behind in leveraging it due to the limitations of most simulation frameworks, including the lack of low-level software control and flexibility, Robot Operating System integration, realistic physics, or photorealism. This hindered progress in (visual-)perception research, e.g. in autonomous robotics, especially in dynamic environments. Visual Simultaneous Localization and Mapping (V-SLAM), for instance, has been mostly developed passively, in static environments, and evaluated on few pre-recorded dynamic datasets due to the difficulties of realistically simulating dynamic worlds and the huge sim-to-real gap. To address these challenges, we present GRADE (Generating Realistic and Dynamic Environments), a highly customizable framework built upon NVIDIA Isaac Sim. We leverage Isaac's rendering capabilities and low-level APIs to populate and control the simulation, collect ground-truth data, and test online and offline approaches. Importantly, we introduce a new way to precisely repeat a recorded experiment within a physically enabled simulation while allowing environmental and simulation changes. Next, we collect a synthetic dataset of richly annotated videos in dynamic environments with a flying drone. Using that, we train detection and segmentation models for humans, closing the syn-to-real gap. Finally, we benchmark state-of-the-art dynamic V-SLAM algorithms, revealing their short tracking times and low generalization capabilities. We also show for the first time that the top-performing deep learning models do not achieve the best SLAM performance. Code and data are provided as open-source at~\url{https://grade.is.tue.mpg.de}.
    \end{sciabstract}
	%\thispagestyle{empty}
	%\pagestyle{empty}
	%%%%%%%%%%%%%%%%%%%%%%%%%%%%%%%%%%%%%%%%%%%%%%%%%%%%%%%%%%%%%%%%%%%%%%%%%%%%%%%%
	
	\section*{INTRODUCTION}
\label{sec:intro}
%\todo{I don't think it's just DL-based perception that creates problems. noisy depth data, Optical flow, or even clustering of dynamic points can give issues. Even if the EKF or the MPC is wrongly set up directly in real world you have high chance of mess up}
Directly conducting real-world robotics experiments to test new methods, particularly in dynamic environments, is not safe. There are significant risks related to unforeseen failure cases that can induce harm to people, animals, or infrastructures. The problem is further exacerbated when those methods rely on perception-based methods to perform their tasks. Those are indeed highly susceptible to noise and, when deep learning (DL) is involved, often do not include formal performance guarantees or uncertainty quantification. Pre-recorded datasets can then be used to develop and evaluate new approaches. However, relying solely on those for robotic applications is not straightforward due to differences in the form factors of the robots (e.g. placement of the sensors), the sensor settings (e.g. focal length), or noise models. Datasets are also fixed in time, and one cannot record new sensors or change light conditions directly. At the same time, recording and labeling ground truth data from the real world is complex, time-consuming, expensive, and when involving robots, a dangerous process. For example, a VICON system is limited in the number of people it can track% and the variability of scenarios it can handle
, and there exist only a handful of real-world SLAM benchmark datasets with ground-truth information such as TUM-RGBD~\cite{turgbd}, EuRoC~\cite{euroc}, and KITTI~\cite{kitti}. 
Then, although datasets collected for computer vision research like~\cite{cocodataset,airpose,surreal} are visually appealing due to the use of real-world images or rendering engines like Blender~\cite{blender} and Unreal Engine~\cite{unrealengine} (UE), they are limited in several aspects. For example, the absence of basic sensor readings (like IMU/LiDAR) and camera state (position, speed), makes them often unusable for robotics research. Moreover, due to the different research scopes, they generally lack temporal information and, when synthetically generated, contain objects or humans that are flying, incoherently placed, or stitched over panoramic and hallucinated backgrounds~\cite{peoplesanspeople,3dpeople}. This lowers their realism and often limits their usability in robotic contexts. %Some robotic-focused synthetic datasets have begun to appear, especially towards the training of deep visual odometry methods. One example is TartanAir~\cite{tartanair}, which comprises several sequences generated with a teleporting camera, varying visual settings, and some dynamic subjects. 7
Finally, an offline dataset cannot be used for evaluating methods where a robot needs to make real-time decisions, such as obstacle avoidance, environment interaction, or active SLAM, and are limited in the motion patterns and scenarios they provide~\cite{tartanair}. %However, their generation code is not public, and they lack any physics simulation. Moreover, their approach cannot be used for online applications (crucial in robotics), and, in most cases, simulated data usually has little visual adherence to reality, thus making generalization for visual-related tasks hard. 

The robotics community often uses simulation software like Gazebo~\cite{gazebo} or WeBots~\cite{Webots04} for developing navigation, grasping, mapping, and other methods before testing those in the real-world. However, there are often difficulties in i) obtaining and controlling animated assets, ii) simulating dynamic environments, iii) fully customizing and controlling those simulation engines, and iv) bridging the gap between the simulated visual data and the real world, due to the limitations of the underlying engines. 
\begin{figure*}[!ht]
  \includegraphics[width=0.245\textwidth]{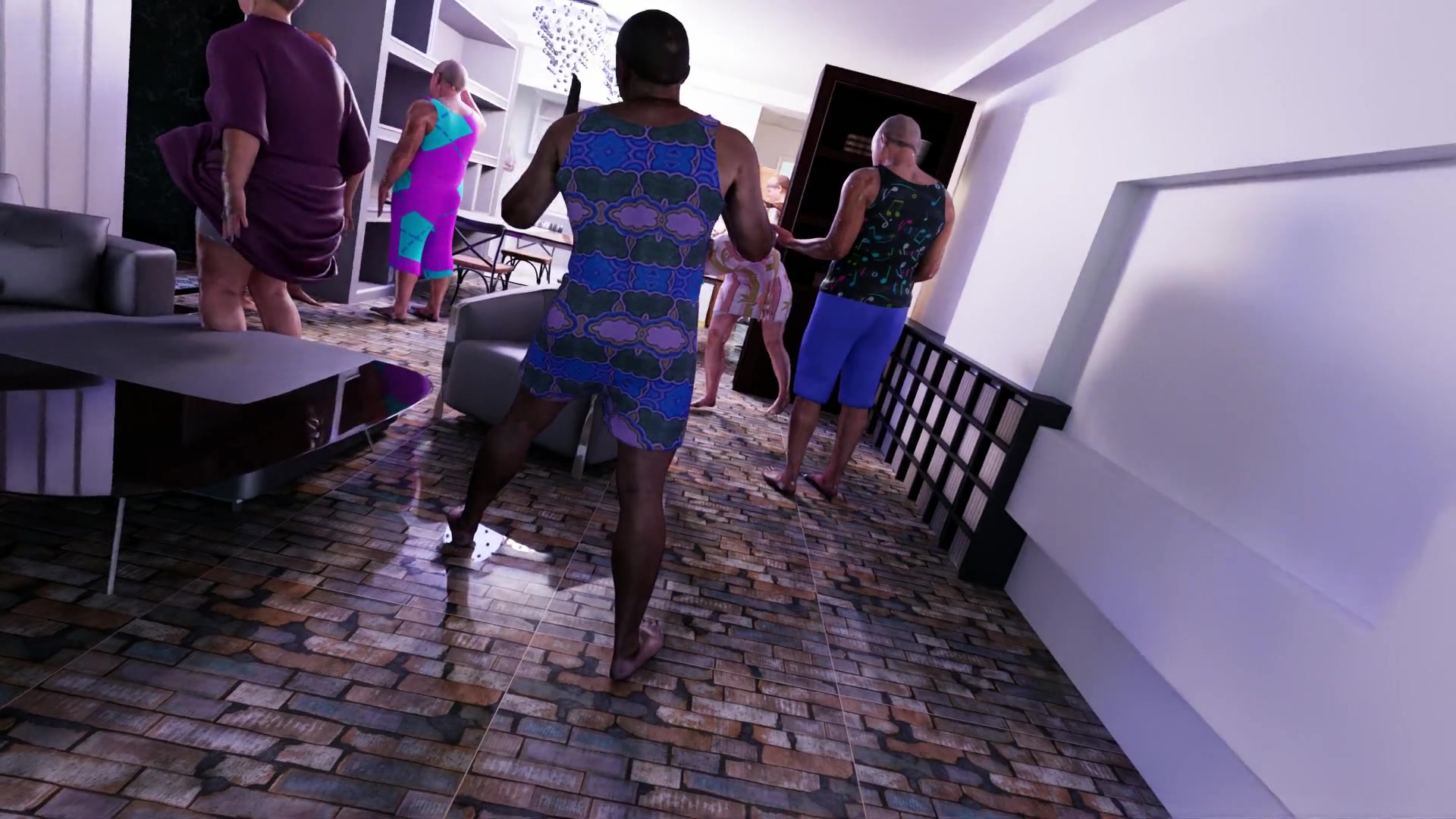}
  \includegraphics[width=0.245\textwidth]{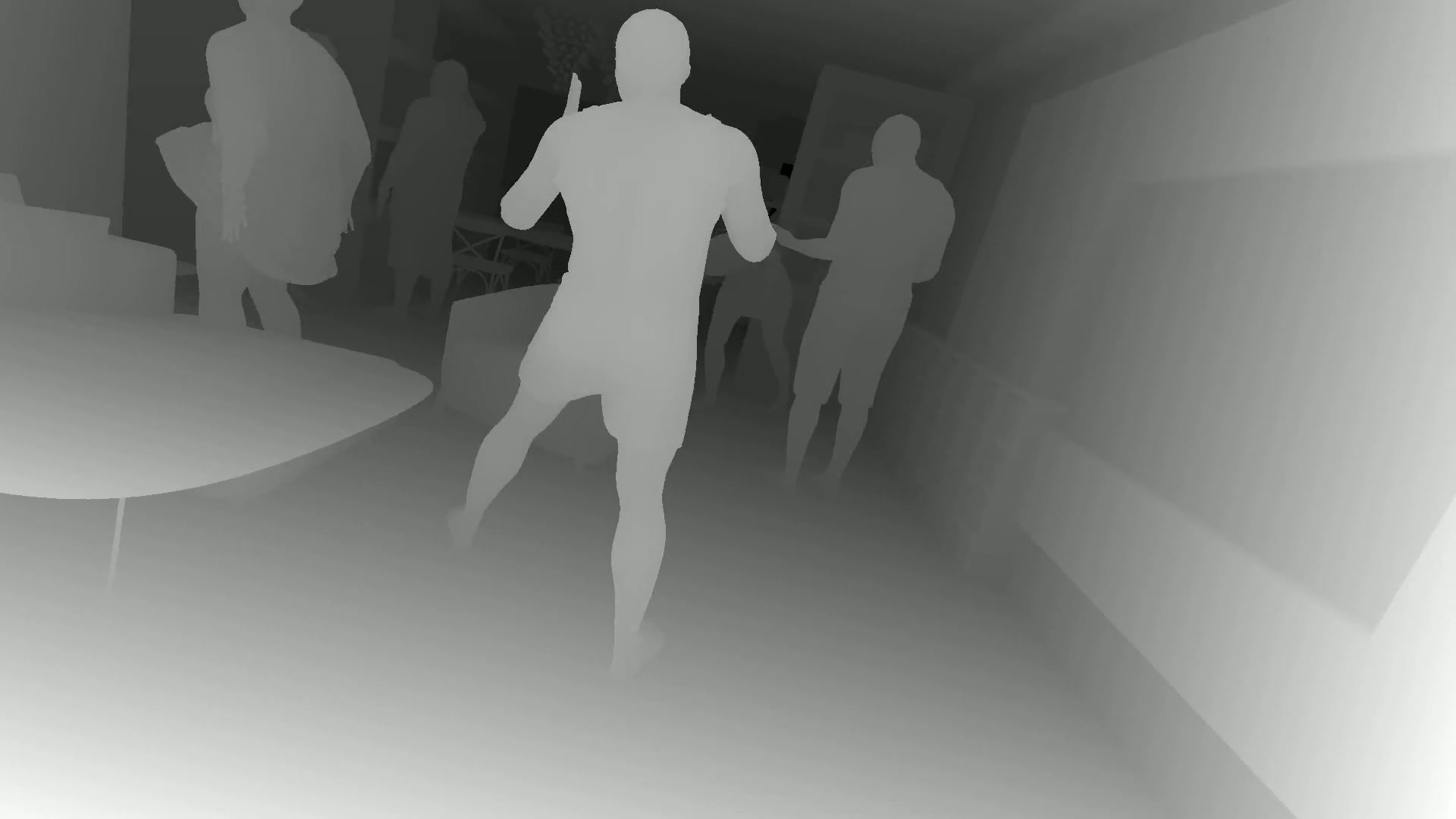}
  \includegraphics[width=0.245\textwidth]{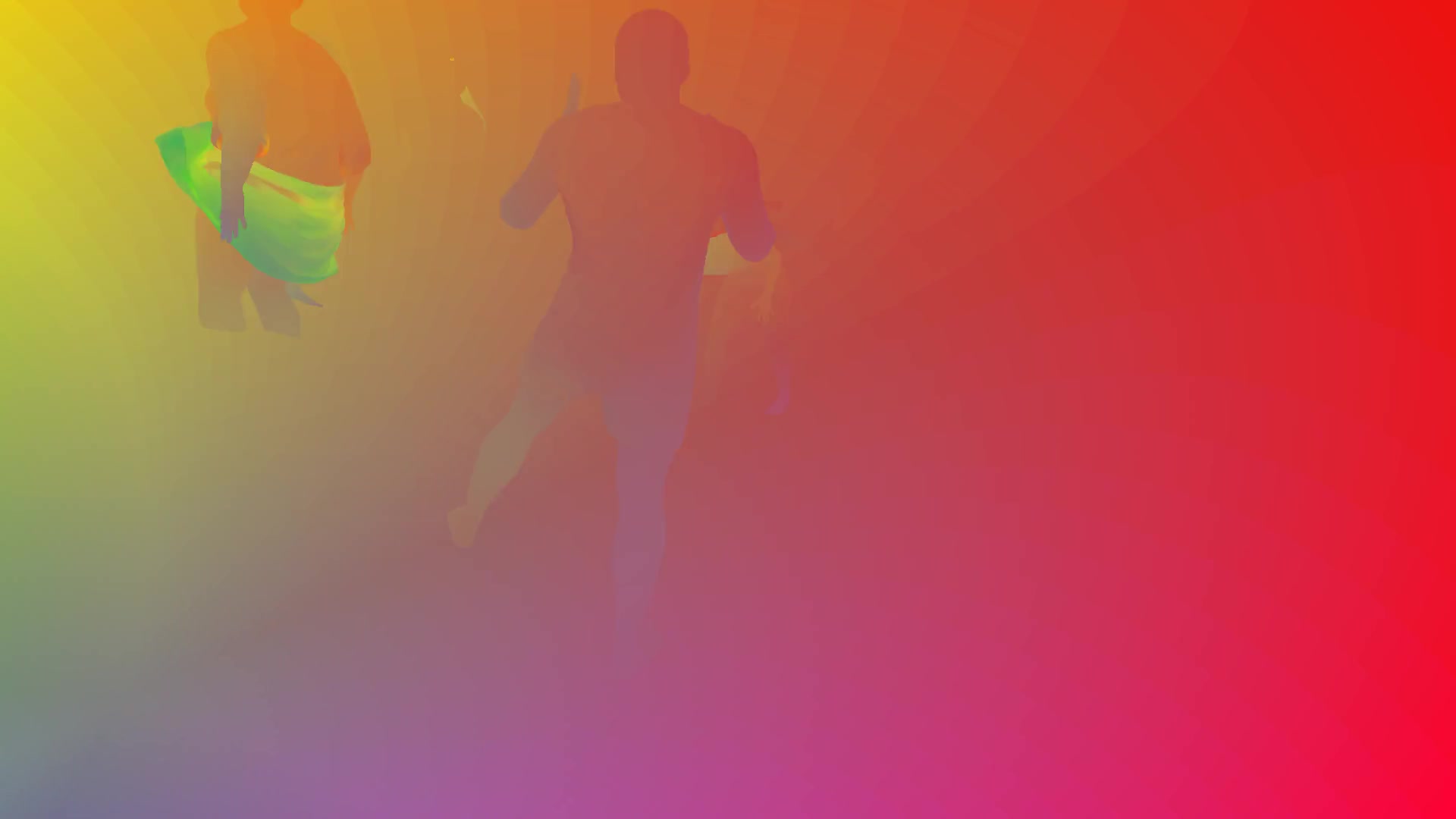}
  \includegraphics[width=0.245\textwidth]{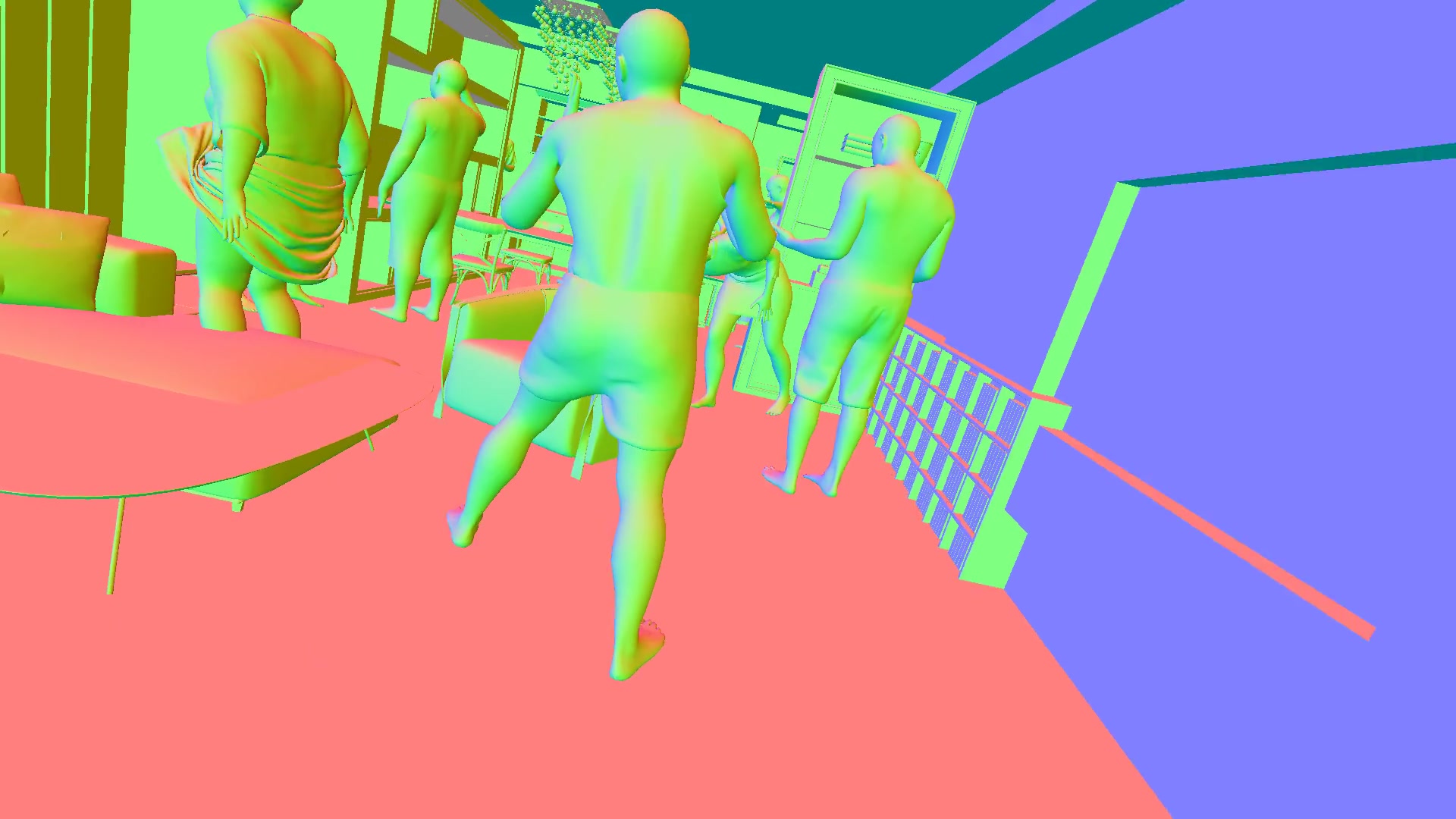}
  
  \vspace{3pt}
  
  \includegraphics[width=0.245\textwidth]{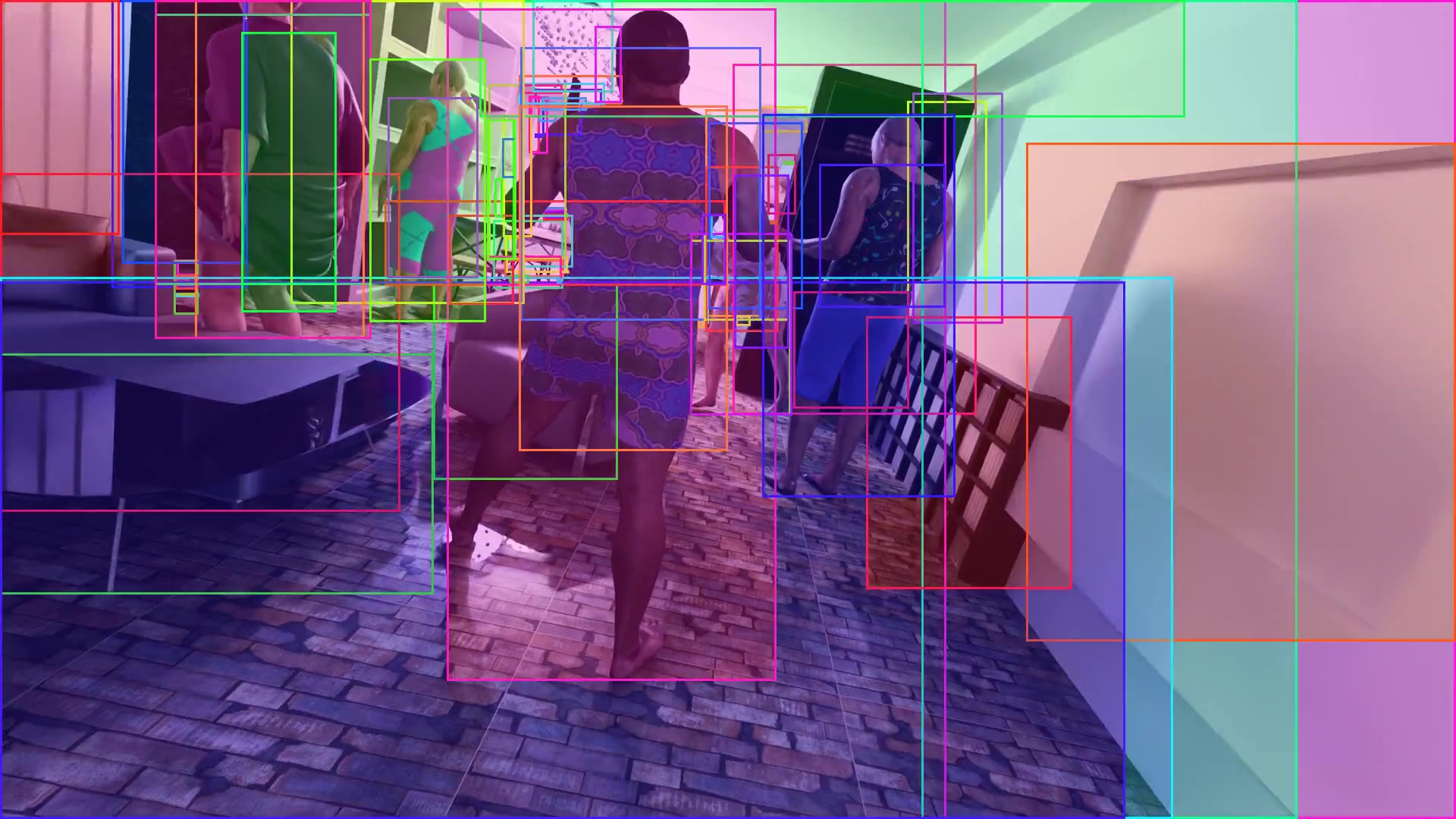}
  \includegraphics[width=0.245\textwidth]{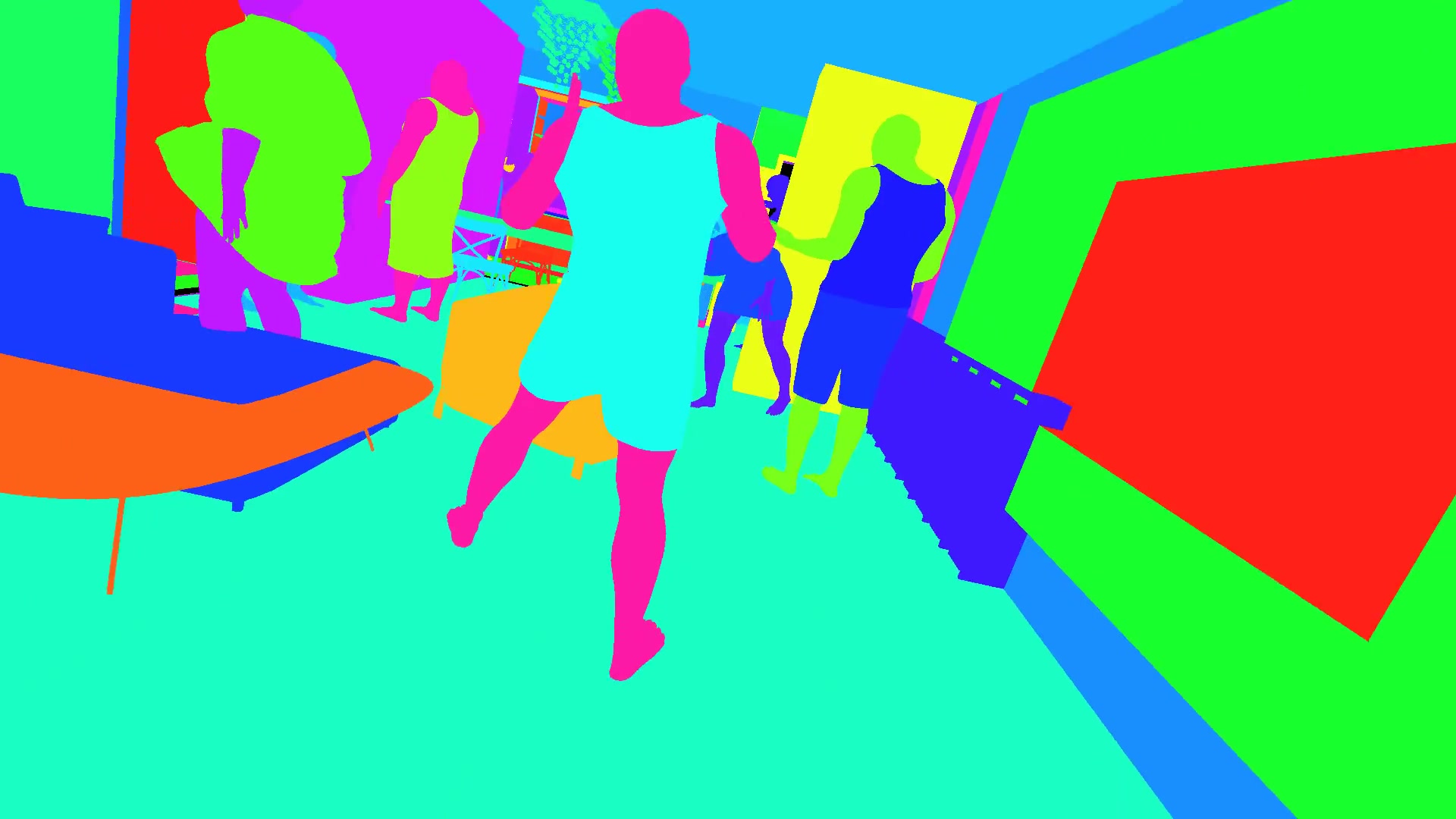}
  \includegraphics[width=0.245\textwidth]{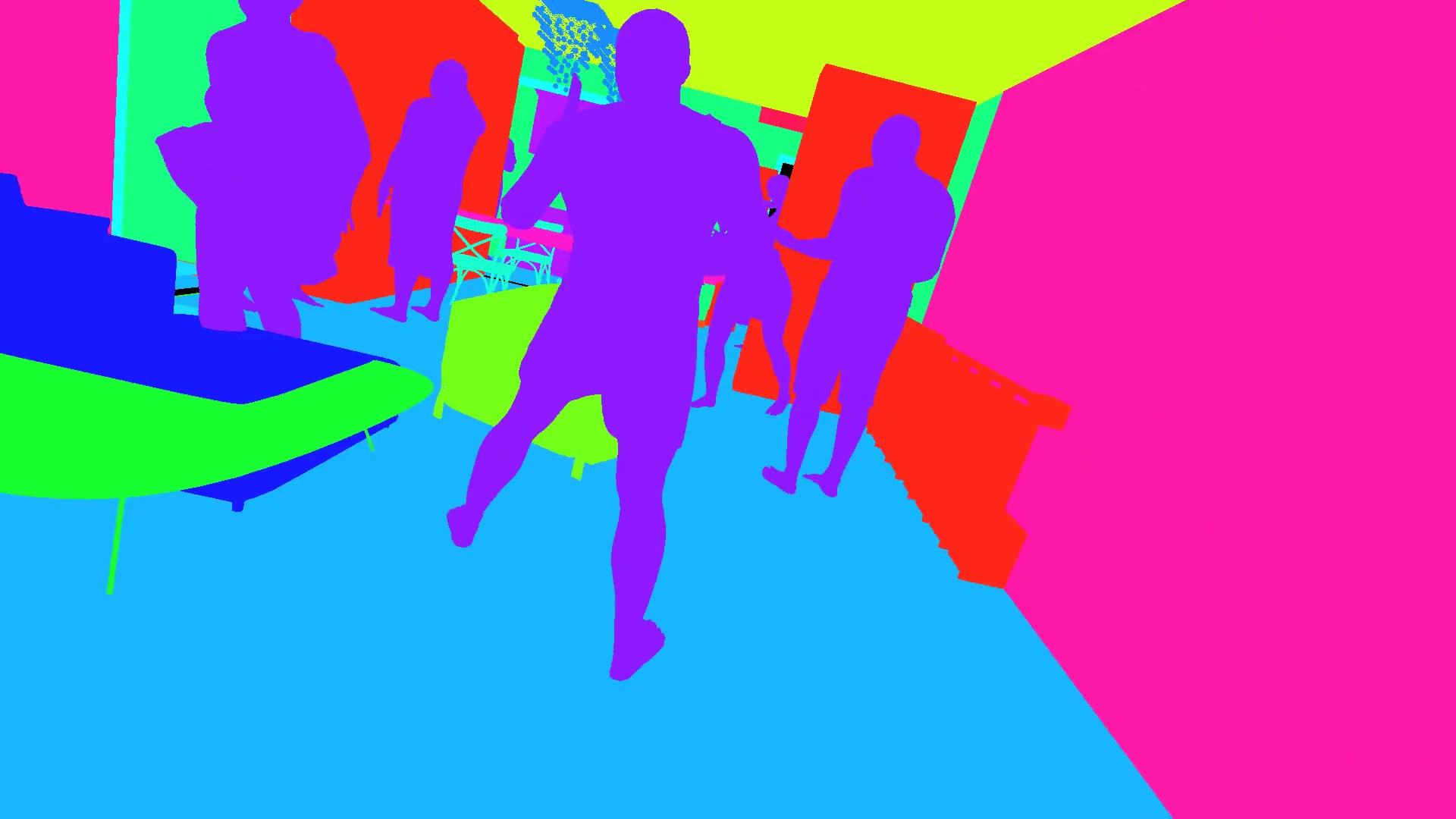}
  \includegraphics[width=0.245\textwidth]{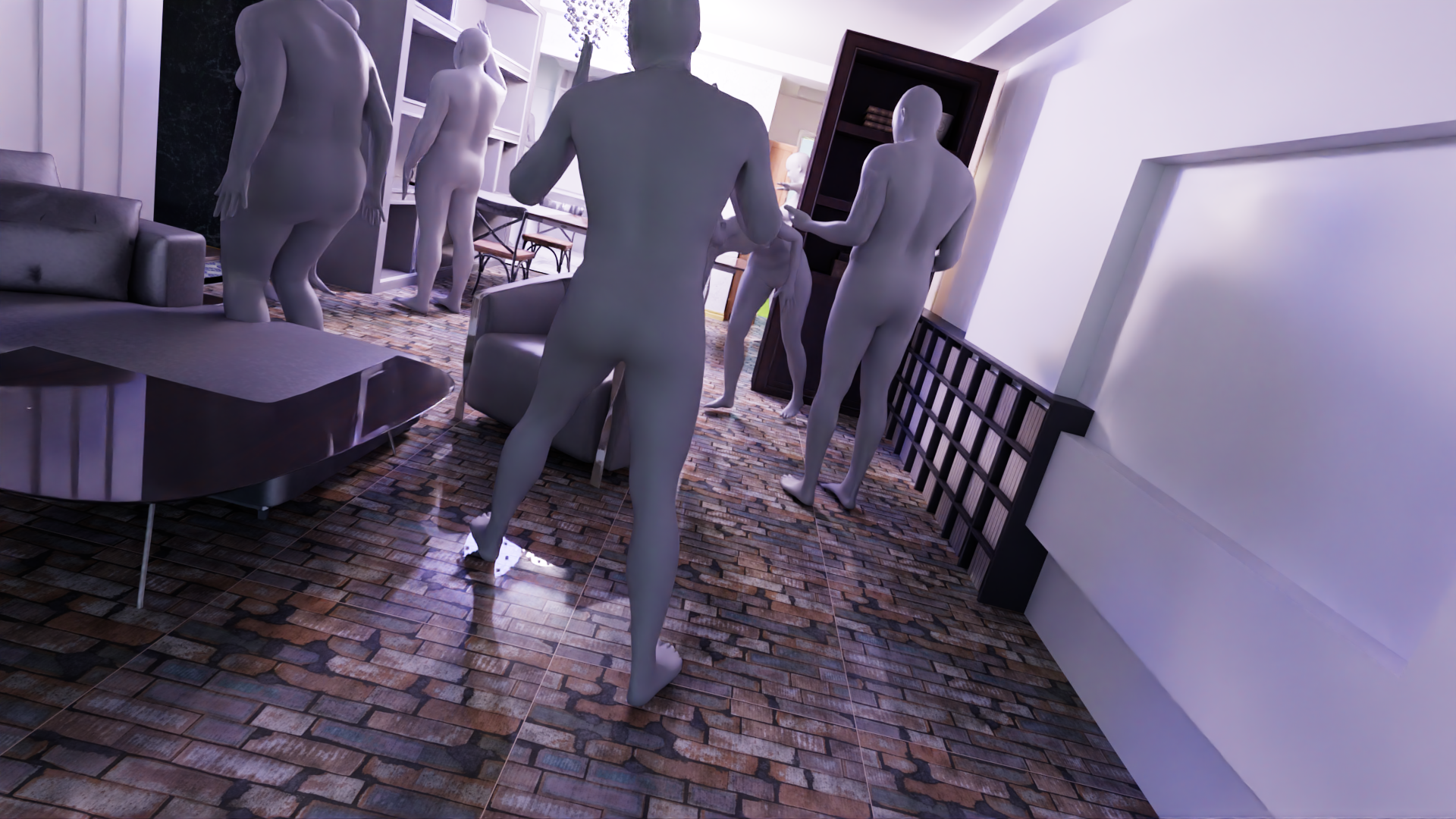}
  \caption{An example of the data generated using our simulation framework GRADE, assets from \textit{Cloth3D humans}~\cite{cloth3d} and one of the environments from \textit{3D-Front}~\cite{3d-front,3d-future}. Top row, left to right: Rendered RGB image, corresponding depth map, optical flow, and surface normals. Bottom row, left to right: 2D bounding boxes, semantic instances, semantic segmentation, and SMPL~\cite{smpl} shapes. Best viewed in color.}
  \label{fig:gendata}
\end{figure*}
Gazebo~\cite{8921035,FARLEY2022102629} is often the to-go choice thanks to its simplicity, reliable physics engine, and ROS~\cite{ros} integration. However, it lacks photorealism and full simulation control, only limited assets/worlds can be easily loaded, and real-time performance is often inadequate even for single robots in simple worlds with limited rendering necessities~\cite{9196621,8088134,Platt2022}. Alternatives, such as BenchBot~\cite{benchbot}, AirSim~\cite{airsim}, Ai2Thor~\cite{ai2thor}, iGibson~\cite{igibson}, AI-Habitat~\cite{habitat19iccv}, and Sim4CV~\cite{Mller2018} all lack either low-level controllability of the simulation, ROS integration, or realistic physics and appearance. Additionally, some simulate \textit{only} rigid objects~\cite{gazebo,habitat19iccv} or do not include dynamic assets which would introduce challenges like their placement, management, or generation. Applications like Sim4CV or Kubric~\cite{greff2021kubric} are focused only on CV-related applications and are hardly usable for robot simulations. AirSim sought to bridge the rendering gap by working on top of UE. However, it provides limited APIs, does not support custom or multiple heterogeneous robots, or direct joint control, and has a loose and incomplete integration with ROS in its native version~\cite{airsim-arbitrary, airsim-arbitrary2, airsim-ros}. Ai2Thor is not customizable for general robotics purposes, due to its focus on AI and visual tasks, lacks sensor interfaces like IMUs and LiDARs~\cite{ai2thor-lidar}, and has no direct ROS integration~\cite{ai2thor-ros}. The same is true for AI-Habitat, which is mainly focused on navigation, although a community's plugin for ROS integration in this case exists~\cite{ros-x-habitat}. Still, as an external package not directly integrated into the primary simulator, it creates overhead, lack of support, and thus features. Indeed, as of today, \texttt{ros\_x\_habitat} does not support ROS2 or Habitat2.0~\cite{ros-x-habitat-support}. Notably, Habitat3.0~\cite{habitat3} was recently released concurrently with the development of this work, allows the integration of only human animations based on the SMPL-X model~\cite{smpl-x}, having thus, as opposed to us, a quite limited scope. Moreover, both Habitat2.0~\cite{habitat2} and 3.0 support only rigid objects, limiting the realism of the simulation, a problem not present in Isaac Sim. iGibson focuses instead on interactive environments with enhanced characteristics like temperature or wetness. However, its initial version was not customizable or expandable to different tasks, such as navigation, and offered limited visual realism. To solve the realism issue, it was recently ported to Isaac Sim with OmniGibson~\cite{li2022behavior}. It still acts majorly as a plug-in to add characteristics such as temperature and dirtiness control focusing on indoor activities, without any simulated dynamic human~\cite{omnigibson-humans}, and it was published at the same time our work was first released. As also explained in Behavior1K~\cite{li2022behavior}, the Omniverse Isaac Sim allows countering several of the limitations of other engines, e.g. Habitat or THOR-based simulations, while allowing for higher flexibility. Finally, BenchBot and its extension BEAR~\cite{bear} are two solutions aimed at introducing a procedural way to benchmark (active) SLAM methods using Isaac Sim. However, they do not include any dynamic assets natively and, as a benchmark suite, are a close system by nature. They also employ control policies that could be unrealistic for most robots (e.g. 1 cm and 1$^\circ$ goal position accuracies). Then, due to their limited scope and the addition of additional APIs and software layers between the user and the simulator itself, they lack the desirable customization possibilities. For example, the integration of already-developed methods to control the robot or the easy adaptation to different platforms or tasks is inherently hard as they only provide a limited set of actions for the platforms (i.e. \texttt{move\_[next,angle,distance]}). Therefore, although (non-)rigid moving objects are common in real life, most research in robotics still assumes a (semi-)static world, and is developed in very controlled scenarios, limiting its widespread deployment. For example, the presence of dynamic entities affects core building blocks of vision-based SLAM/odometry methods like feature extraction, place recognition, and relocalization~\cite{surveyVodom,surveySlam,robust-slam-survey}. Commonly, dynamic SLAM methods rely on semantic segmentation or optical clues built on top of static SLAM frameworks to mask the extracted visual features. However, due to the low visual fidelity of many robotics simulators, the sim-to-real transfer gap is often difficult to address, making evaluation hard and based only on a few real-world sequences. This, jointly with the aforementioned problems, is detrimental to the development and evaluation of methods strongly affected by the dynamic components of the environment. Therefore, it is imperative to have a simulation framework that has, at least, the following key characteristics: i) physical realism - to correctly simulate the dynamics, ii) photorealism - to lower the perception gap, iii) full controllability and iv) the ability to simulate dynamic entities - to allow a widespread deployment. The integration with ROS, although optional, would also help the wider adoption of such a framework as it is commonly used to develop higher-level software working simultaneously in simulation and real robots.

To address the open issues discussed so far, we present a solution for Generating Realistic And Dynamic Environments --- GRADE. GRADE is a flexible, controllable, customizable, photorealistic, ROS-integrated framework. Like the concurrent OmniGibson work, we use NVIDIA Isaac Sim~\cite{isaac-sim} as the main simulation engine, leveraging its rendering and PhysX engines. We use Blender~\cite{blender}, the Omniverse~\cite{omniverse} suite, and Unreal Engine~\cite{unrealengine} to prepare our assets. We do this by creating a flexible pipeline that can produce visually realistic data in physically enabled environments. We develop and make available a set of functions, tools, and examples to give the community an entry point with low-level access to the Isaac Sim capabilities. In contrast to existing methods, we build \textit{directly} upon Isaac Sim by showing how researchers can easily adapt the simulation to their needs, including interactions with objects, and visual settings, further closing the gap between simulation and real-world scenarios.  Moreover, our work is not just aimed at a new benchmarking method or application-specific platform like some of the above simulators, but rather an open system that can be easily expanded towards different goals.

We employ GRADE to perform heterogeneous multi-robot simulations, data generation, on-line simulations, and precise experiment repetition, highlighting its versatility. We generate an indoor dynamic environment dataset to demonstrate the visual realism of the simulation and use it to perform extensive experiments with YOLOv5~\cite{yolov5} and Mask R-CNN~\cite{maskrcnn}. We obtain compelling syn-to-real performance on both the COCO dataset and some dynamic sequences of the TUM-RGBD dataset~\cite{turgbd}. With this, we show that pre-training with GRADE-generated data can outperform the baseline models on the COCO dataset, while synthetic data alone can be at par without \textit{any} fine-tuning on the TUM data. We then extensively benchmark state-of-the-art indoor dynamic V-SLAM algorithms. Our evaluations provide evidence of their poor generalization capabilities by failing to either correctly or completely track the trajectories, a metric often overlooked. Finally, our results show how, and different from the common belief, using the best-performing deep model does not always yield the best performance in trajectory error or tracking rate.

\section*{RESULTS}
\label{sec:exp}
% Here we first briefly introduce our data generation procedure, further expanded in both the \textit{Materials and Methods} section and in the Supplementary Materials. Then we extensively demonstrate i) the precision of the experiment repetition system of GRADE, ii) GRADE's syn-to-real transfer capabilities, and iii) the limitations of state-of-the-art dynamic V-SLAM methods using GRADE.

 % Then, we analyze the performance of the experiment repetition tool in terms of the precision of the repeated poses and the quality of the re-rendered images and depth data. Finally, we extensively evaluate both syn-to-real performance and representative dynamic V-SLAM methods.
% \subsection*{Application of GRADE in different scenarios}
% \label{subsec:scenarios}
% These scenarios emphasize GRADE's versatility and effectiveness in various tasks by addressing diverse research themes including ground-truth data generation for visual and robotics tasks, online/offline testing of robotics applications, and customizable experiment repetition. We accomplished this by manipulating simulations by crafting diverse tools and APIs based on our developed backbone over IsaacSim. Those address various aspects like robot control, asset placement, environments, robot platforms, control techniques, and the inclusion of SIL. 

\subsection*{Data Generation in GRADE}%(Multi-)Robot simulations and Active SLAM}
\begin{figure*}[!ht]
  \includegraphics[width=0.195\textwidth, height=2.5cm]{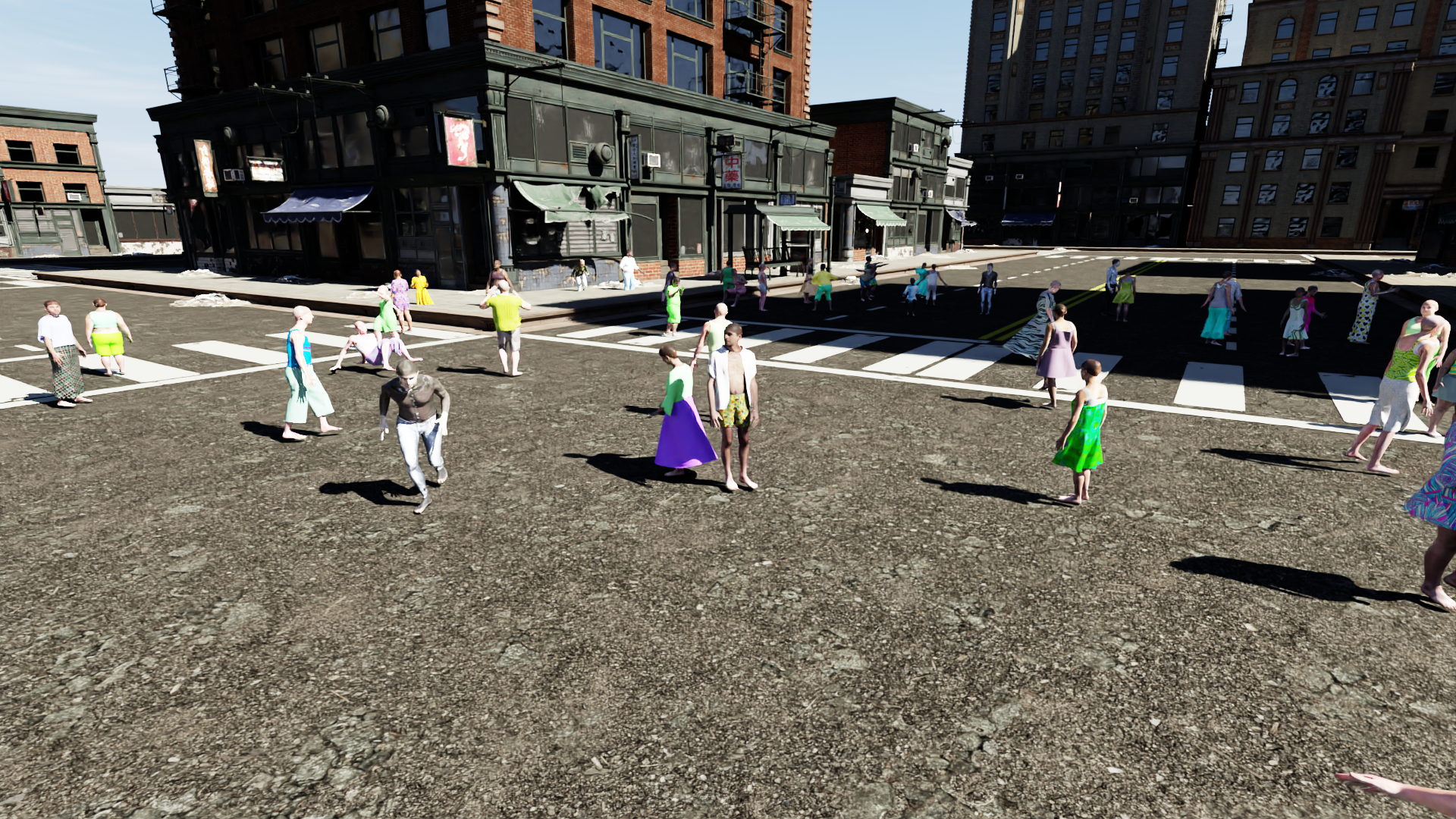}
  \includegraphics[width=0.195\textwidth, height=2.5cm]{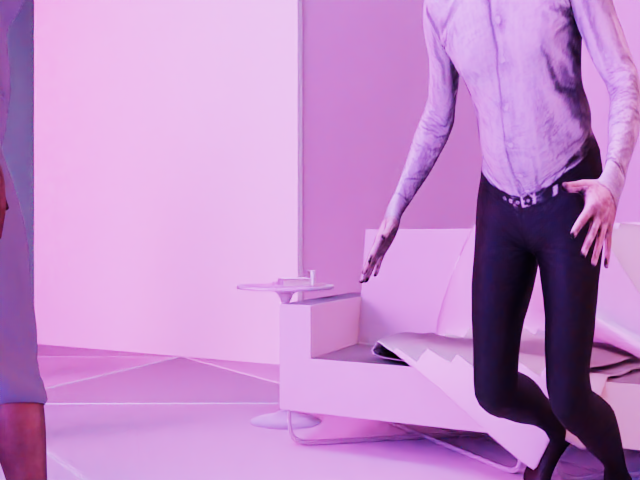}
  \includegraphics[width=0.195\textwidth, height=2.5cm]{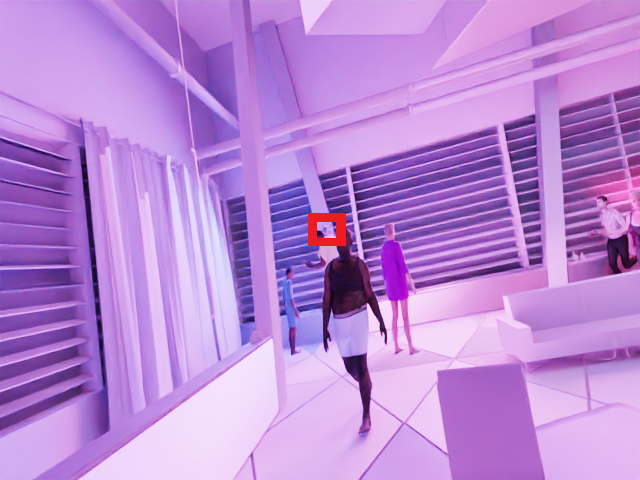}
  \includegraphics[width=0.195\textwidth, height=2.5cm]{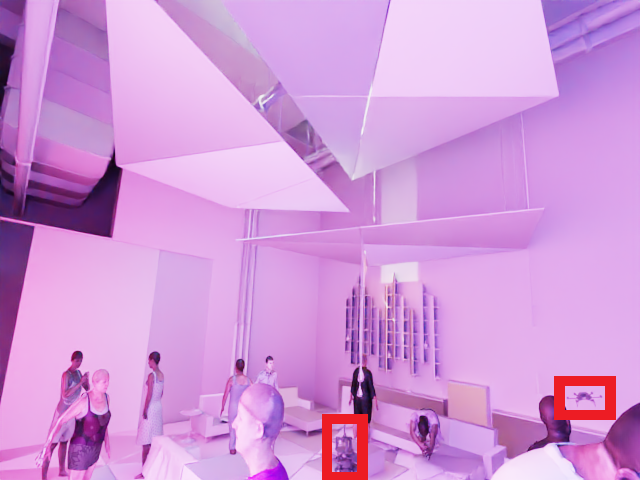}
  \includegraphics[width=0.195\textwidth, height=2.5cm]{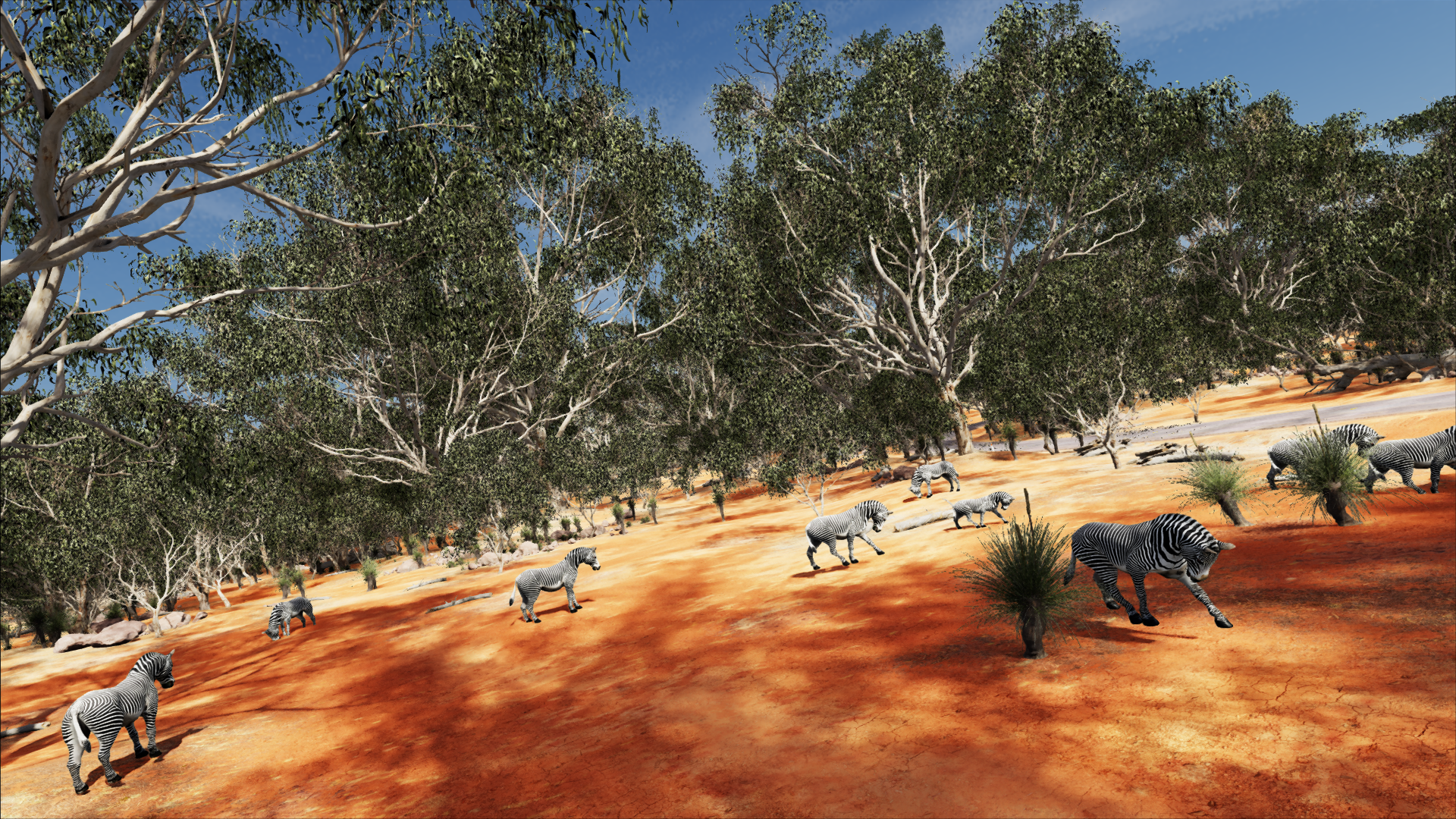}
  
    \vspace{3pt}
    
  \subcaptionbox{People in a city\label{img:scen_city}}{\includegraphics[width=0.195\textwidth, height=2.5cm]{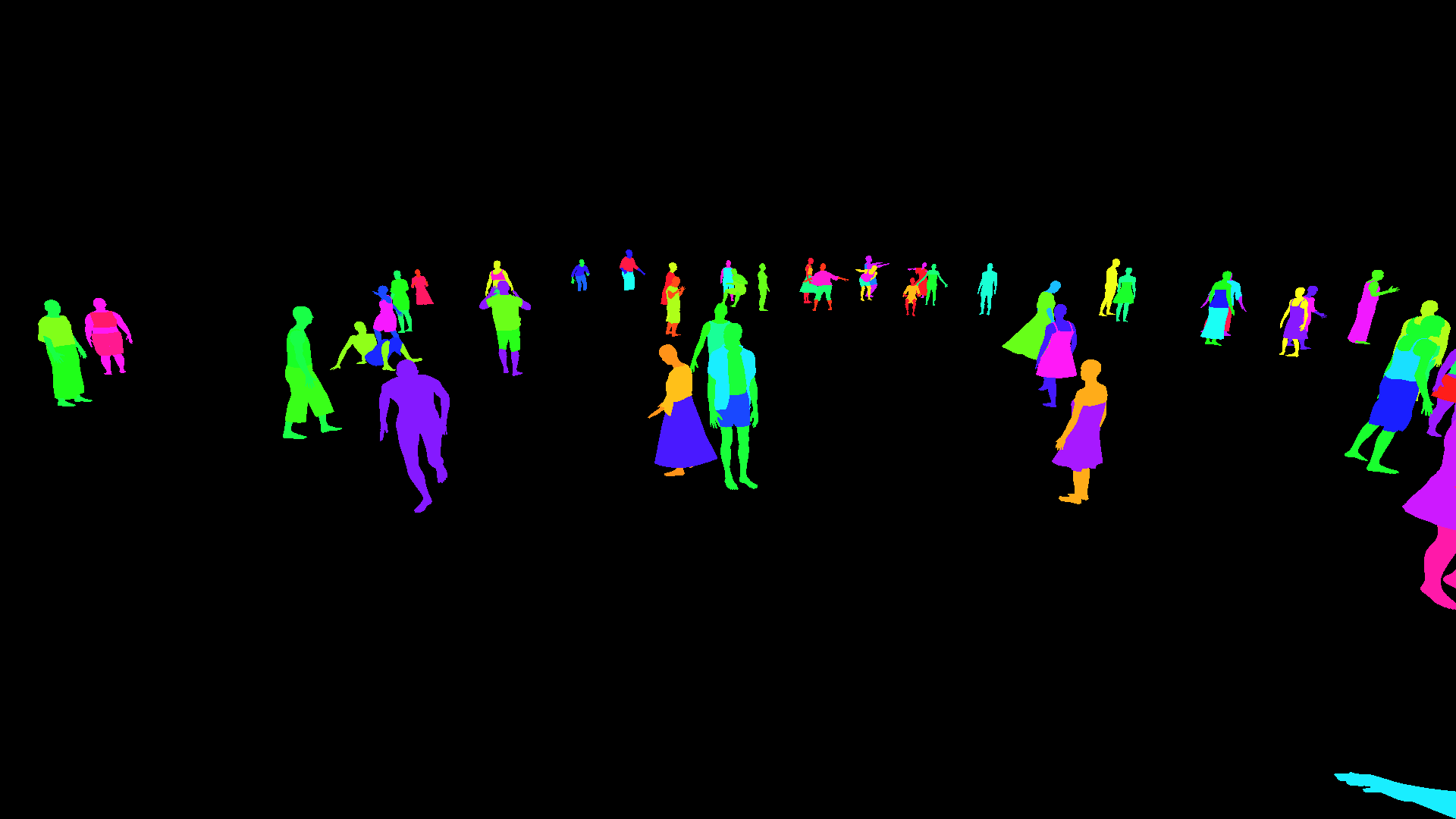}}\hfill
  \subcaptionbox{Multi-robot: ground robot\label{img:scen_gr1}}{\includegraphics[width=0.195\textwidth, height=2.5cm]{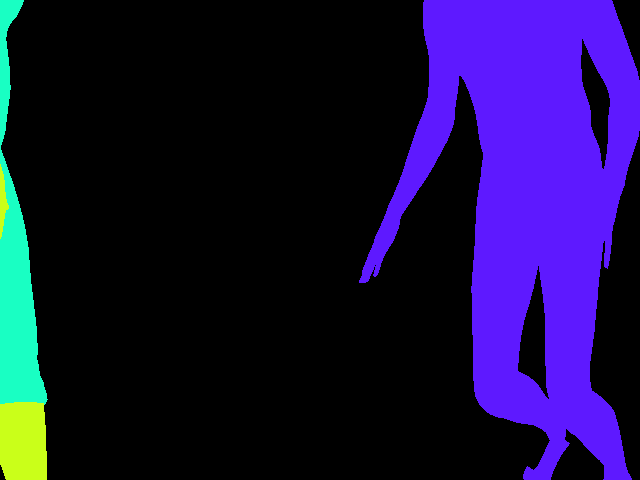}}\hfill
  \subcaptionbox{Multi-robot: UAV \#1\label{img:scen_gr2}}{\includegraphics[width=0.195\textwidth, height=2.5cm]{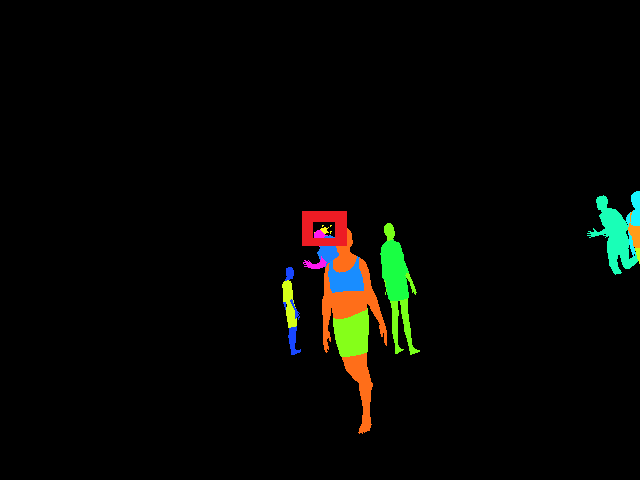}}\hfill
  \subcaptionbox{Multi-robot: UAV \#2\label{img:scen_gr3}}{\includegraphics[width=0.195\textwidth, height=2.5cm]{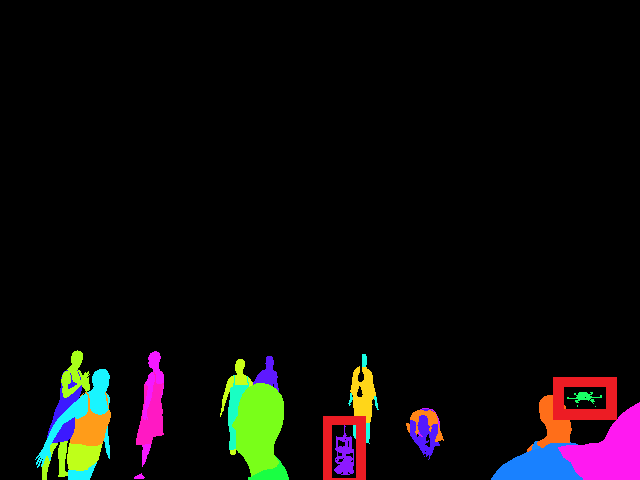}}\hfill
  \subcaptionbox{Zebras in a savanna\label{img:scen_zebra}}{\includegraphics[width=0.195\textwidth, height=2.5cm]{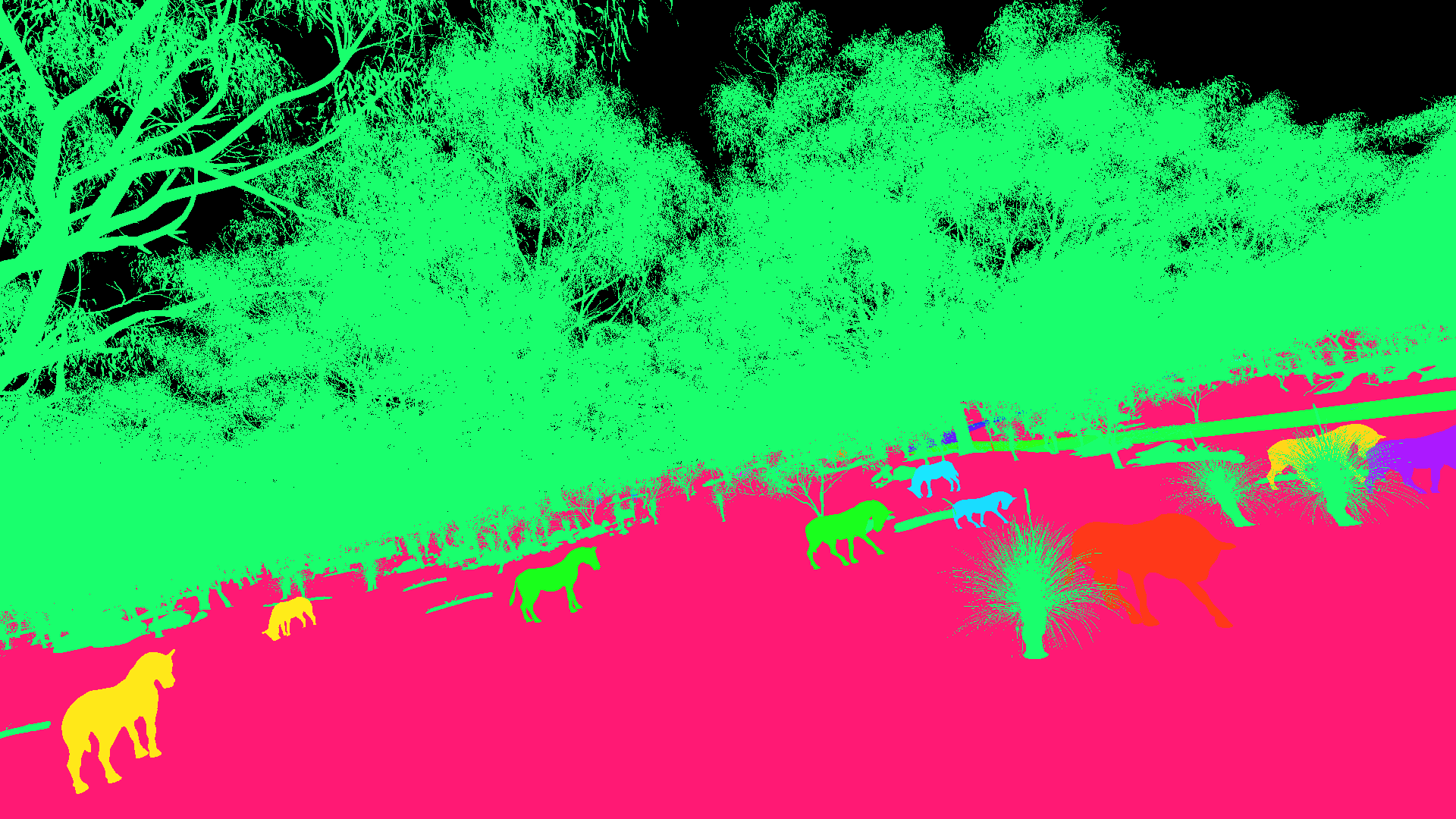}}
  
  \vspace{8pt}
  
  \caption{The RGB images are in the top row, with the associated instance segmentations (randomly colored) below. For the multi-robot UAV images, we highlight the other robots in the field of view with a red box. Best viewed in color.}
  \label{img:scenarios}
\end{figure*}

\label{subsec:indoor}
We use GRADE to generate the synthetic data for our evaluations. The environments are taken from the 3D-Front~\cite{3d-front} dataset and enhanced with random textures and lighting conditions. Additionally, we collect data once in an outdoor city-like environment~\cite{outdoorcity} (Fig.~\ref{img:scen_city}), and once in an indoor~\cite{interior-sketch} multi-robot scenario with one UGV robot (Fig.~\ref{img:scen_gr1}) and two UAVs (Fig.~\ref{img:scen_gr2} and Fig.~\ref{img:scen_gr3}). The dynamic components in the scenes are \textit{animated} humans and, in some experiments, random flying objects. We then use unmanned aerial and ground vehicles (UAV/UGV) to explore environments by using Active SLAM methods. The UAV model is taken from the RotorS package~\cite{rotors} and controlled with a custom 6DOF PID controller in the loop with the main simulation software receiving position goals from a non-linear MPC linked to an adaptation of the FUEL~\cite{zhou2021fuel} framework capable of working on simulators different from the original one. The ground platform is a three-wheeled omnidirectional UGV with an independently rotating camera~\cite{independent-camera}. It is controlled using velocity setpoints from the iRotate package~\cite{irotate} for Active SLAM exploration of indoor environments. An example of the richly annotated data generated can be seen in Fig.~\ref{fig:gendata}. 

% With this experiment, we show the scalability of the system to crowds and outdoor scenes. 

% By leveraging our APIs we introduce a parameter to identify the ``type'' of robot to then, in a single pass, dynamically import and place those in the environment, equip them with their specific sensor suite, attach to each one unique ROS topics, and launch the corresponding control software.

% While to running these experiments we resort to and adapt several online and autonomous approaches, we use the data generated for different downstream offline tasks. Indeed, since the data generated here is used for both the syn-to-real learning experiments and the \textit{offline} SLAM benchmarking, we will give a more thorough description of both the simulation preparation and the data collection strategy in the method section and the Supplementary Materials.

\subsection*{Experiment repetition}
% Visual SLAM, and several other robotics applications, primarily rely on image data. However, precisely varying experiments' conditions in terms of lighting, simulated/visible assets, or location of camera or other sensors in the same physics-enabled environment is often not possible, especially in the real-world. 

In GRADE we provide a way to precisely replay any experiment as is or by modifying a previously recorded scenario in which any number of conditions can be selectively altered. This includes the addition of new sensors to the robot, changes in light conditions, or the inclusion of new robots, humans, animals, or other objects, etc., in a previously recorded experiment. This, while keeping the physics simulation enabled (more details and figures in the Supplementary Materials). This allows us to study the robustness of the developed methods by changing the robot's surrounding conditions or to expand previously collected datasets under the \textit{exact} same settings. For example, as described further in our experiments, we will use this to re-render dynamic scenes into static ones to evaluate the influence of moving entities.

To test the exactness of the experiment repetition procedure we employ a 60-second experiment and compare the precision of i) position and orientation values on 3601 poses, and of the re-rendered ii) RGB, and iii) depth images (1800 each). The average deviation between the position ($x,y,z$ in meters) and orientation ($roll,pitch,yaw$ in radians) of the robot are $[-3.5\text{e}{-6}, -9.00\text{e}{-7}, -1.94\text{e}{-7}]$ and $[-4.61\text{e}{-7}, -8.70\text{e}{-7}, -4.48\text{e}{-7}]$ with standard deviations $[2.75\text{e}{-5}, 3.06\text{e}{-5}, 2.71\text{e}{-5}]$ and $[2.93\text{e}{-5}, 2.90\text{e}{-5}, 2.90\text{e}{-5}]$ respectively. The average RGB image structural similarity~\cite{1284395} on all the frames is $99.6\%$ with a $0.15\%$ deviation. Finally, the mean of the average difference between the original and repeated depth maps is $0.0015$ meters and the standard deviation is $0.0019$ meters, over 1800 frames. Note that this is driven mostly by aliasing effects and pixel values that lie alongside the borders of the objects. Finally, the corresponding depth images have a structural similarity attested at $99.8\%$ with a $0.23\%$ standard deviation.

\subsection*{Syn-to-real transfer learning}
\label{subsec:vrealism}
% Humans are arguably one of the most common dynamic entities in many scenarios, especially indoors. Indeed, detecting and segmenting humans are two of the core techniques applied to visual odometry and dynamic V-SLAM methods to filter out noisy features~\cite{surveyVodom, surveySlam}. Therefore, for approaches that rely on visual data, 
To enable quick and reliable real-world deployment of various methods in robotics, a simulation that closely resembles the real world with the smallest sim-to-real gap is needed. To this end, here we showcase the results of GRADE's syn-to-real transfer capabilities using two popular neural networks, i.e. YOLOv5~\cite{yolov5} and Mask R-CNN~\cite{maskrcnn}. We do so by using both synthetic and real-world data, and evaluating the network performances using the COCO dataset and some popular dynamic sequences of the TUM-RGBD dataset~\cite{turgbd}. Our objective is to demonstrate that synthetic data generated with GRADE successfully captures real-world features and enables the training of models that generalize well over real images. We train the networks in three modalities: 1) from scratch with both synthetic and real data, 2) fine-tuning with real-world images the networks pre-trained on synthetic data, and 3) using datasets of mixed synthetic and real-world data. 

Using GRADE, we automatically generate data by running several 60-second experiments as described in the previous \textit{Data Generation} section, further expanded in \textit{Materials and Methods} and in the Supplementary Materials. To train YOLO and Mask R-CNN we use both a subset of this dataset, which we will refer to as \textbf{S-GRADE}, and the complete one, \textbf{A-GRADE}. Images with a high probability of being occluded by flying objects due to peculiar depth and/or color information are automatically discarded. S-GRADE consists of 18K frames, with dynamic humans and without flying objects. Of those, 16.2K have humans in them and 1.8K are only background. A-GRADE is made of all available data, i.e. 591K images, out of which 362K have humans.  
% The details data generation procedure and of the dataset are thoroughly explained in both the Method section and in the Supplementary Materials. 

The real data come from the COCO and the \textit{fr3/walking} sequences of the TUM-RGBD~\cite{turgbd} datasets. From COCO we utilize only the subset of data containing humans in the frame and will call this dataset \textbf{CH} (\textbf{C}OCO-\textbf{H}umans) from now on. CH contains $64115$ training and $2693$ validation images. From those, we randomly sample 1256 training and 120 validation images, totaling $\sim 2\%$ and $\sim 4\%$ of the corresponding training/validation set of CH. In this work, we call this subset \textbf{S-CH} and use it to understand how the networks perform with limited real data. The \textit{fr3/walking} sequences consist of $3579$ images with people, $5362$ instances, and $130$ background samples. We manually label those with precise bounding boxes and segmentation masks using the free version of Roboflow~\cite{roboflow}, and release this data publicly. We will use \textbf{TH} to indicate this data in our tests. Differently from CH, the TH dataset is more related to the synthetic data we generate with GRADE, i.e. indoor dynamic sequences, as opposed to COCO's broader variability of scenarios like outdoor scenes or crowds. Our baseline models (called BASELINE henceforth) are the networks trained on the full COCO dataset. We evaluate the performance with the COCO standard metric (mAP@[.5, .95], mAP in this work) and the PASCAL VOC's metric (mAP@.5, mAP50 in this work). Note that, differently from PeopleSansPeople~\cite{peoplesanspeople}, we save the best checkpoint based on the training dataset's own validation set, i.e. not CH. We do not perform any hyperparameters search and use the default network settings. The Supplementary Materials contains a recap of all the datasets in Tab.~\ref{tab:str_dataset_humans}.
 
\subsubsection*{Human Detection with YOLOv5}
We trained YOLOv5s for 300 epochs and report the results in Tab.~\ref{tab:yolo}. The model trained from scratch using the S-GRADE data shows lower precision compared to the one trained using exclusively S-CH when evaluated on CH. Testing the same models on TH data revealed instead comparable performances, with the network trained solely with S-GRADE data achieving a $\sim 5\%$ lower mAP50 but a $6\%$ higher mAP. Furthermore, the network pre-trained first with S-GRADE and then fine-tuned on S-CH shows an increased mAP for both metrics of around 6\% on the CH validation set, and of $8\%$ in mAP50 and $12\%$ in mAP on the TH dataset w.r.t. the model using only S-CH. Similarly, fine-tuning S-GRADE with the full CH dataset overcomes the baseline by $\sim 5\%$ for both metrics on the CH validation set, and $\sim +2\%$ mAP50 and $\sim +5\%$ mAP on the TH dataset. Moreover, models that use the A-GRADE dataset during training exhibit comparable or slightly worse performance on the CH validation set than the ones that use S-GRADE data. At the same time, using A-GRADE rather than S-GRADE data yields better performance on the TH dataset. We can then say that the models that use A-GRADE or S-GRADE during training tend to perform well on indoor human detection and do not generalize well to CH. This is likely due to the lack of examples in the synthetic dataset that can correctly represent the data distribution of CH.

Evaluating models using the 50th training epoch checkpoint, identified as E50 in Tab.~\ref{tab:yolo}, consistently outperformed the corresponding A-GRADE and S-GRADE in all metrics and datasets. Moreover, A-GRADE-E50, when tested on TH data, performs better than models trained from scratch on both synthetic and S-CH data, as well as models fine-tuned on S-CH, achieving a remarkable $79.8\%$ mAP50. This indicates that there would be an advantage of using the real-world data during validation, as done in PeopleSansPeople~\cite{peoplesanspeople}. However, that would not be fair and would prevent us from correctly evaluating the performance of using solely the synthetically generated images.

Moreover, we notice how training with a mixed dataset seems more advantageous than using a pre-train and fine-tuning strategy, showing considerable improvements on the TH dataset, up to $\sim 11\%$ mAP50 and $\sim 16\%$ mAP of the model trained on the combined S-GRADE+S-CH dataset. The only noticeable improvement on the tests with the CH validation set is related to the S-GRADE+S-CH training set, likely due to the difference in cardinality of the combined datasets used. Overall, the best-performing model that we obtained is S-GRADE+CH with an improvement of $\sim 5\%$ on both metrics on the CH dataset, and $2\%$ and $6\%$ on the TH dataset, compared to the baseline. These results indicate how the data generated with GRADE can be used both to address the syn-to-real gap and improve the results of the models trained only on real-world data. %Moreover, also considering the relatively tight margin that we observe in the TH validation between A-GRADE-E50 and the baseline.

\begin{table}[h]
    \centering
    \resizebox{0.7\columnwidth}{!}{
\begin{tabular}{cc|cc|cc}
  \multicolumn{2}{c|}{} & \multicolumn{2}{c|}{CH} & \multicolumn{2}{c}{TH} \\ 
(Pre-)Training Set & Fine-Tuning Set & AP50 & AP & AP50 & AP \\ \hline
 BASELINE & --- & 0.753 & 0.492 & 0.916 & 0.722 \\
 S-CH & --- & 0.492 & 0.242 & 0.661 & 0.365 \\
 S-GRADE & --- & 0.206 & 0.109 & 0.616 & 0.425 \\
 S-GRADE-E50 & --- & 0.234 & 0.116 & 0.683 & 0.431 \\
 A-GRADE & --- & 0.176 & 0.093 & 0.637 & 0.459 \\
 A-GRADE-E50 & --- & 0.282 & 0.154 & 0.798 & 0.613 \\ \hline
  S-GRADE & S-CH & 0.561 & 0.302 & 0.744 & 0.488 \\
 A-GRADE & S-CH & 0.540 & 0.299 & 0.762 & 0.514 \\
 S-GRADE & CH &  \textbf{0.801} & \textit{0.544} & 0.931 & \textit{0.778} \\
 A-GRADE & CH &  \textit{0.797} & 0.542 & 0.932 & \textbf{0.786} \\ \hline
 S-GRADE + S-CH & --- & 0.590 & 0.334 & 0.855 & 0.648 \\
 A-GRADE + S-CH & --- & 0.527 & 0.289 & 0.801 & 0.597 \\
 S-GRADE + CH & --- & \textbf{0.801} & \textbf{0.547} & \textbf{0.938} & \textbf{0.786} \\
 A-GRADE + CH & --- & 0.764 & 0.503 & \textit{0.936} & \textit{0.778} \\
\end{tabular}}
\caption{YOLOv5s bounding box evaluation results. We put in bold the best result and in italics the second best. The baseline is the officially released model of YOLOv5s trained on the full COCO dataset. The `+' sign indicates that the two datasets are mixed.}
    \label{tab:yolo}
\end{table}

\subsubsection*{Human Detection and Segmentation with Mask R-CNN} 

We use the \textit{detectron2}~\cite{wu2019detectron2} implementation of Mask R-CNN, using a 3x training schedule~\cite{detectron2-config} and a ResNet50 backbone. We use the default steps (210K and 250K) and maximum iterations (270K) parameters with four images per batch when training A-GRADE and CH. We reduced those to 60K, 80K, and 90K when training S-GRADE and to 80K, 108K, and 120K and 2 images per batch for S-CH due to their relatively small size. %Note that the number of iterations does not correspond to the number of epochs which is computed as \begin{math}{\mathrm{epoch} = \frac{\mathrm{iteration}\cdot\mathrm{batch\_size}}{\mathrm{number\_of\_images}}}\end{math}. 
We evaluate the models every 2K iterations and save the best one by comparing the mAP50 metric on each of the two tasks, detection and segmentation. Due to A-GRADE size, we evaluate the model trained from scratch on this data every 3K iterations. We save the best model separately for each task and evaluate their accuracy using 0.70 and 0.05 confidence thresholds. Since the training and evaluation schedules greatly impact the performance of this network~\cite{detectron2-lr} we also train from scratch the same network with the CH data using our configuration. Our findings, depicted in Tab.~\ref{tab:mask}, are similar to those discussed earlier for YOLO. Differently than before, fine-tuning using the whole CH does not yield improvements over the official baseline model. However, if we consider the model trained by us using CH for a fairer comparison, we see that using A-GRADE or S-GRADE combined with CH often results in better performance, evidencing the usefulness of our synthetic data. 
%\todo{point is: baseline depends highly on schedule, validation set etc. So we re-train using COCO. Comparing with "baseline" (official data) can be tricky. Thus, we compare also with model trained with CH. A/S-GRADE+CH and pretrain with synthetic and fine-tuning with CH yield better performance. Is it clearer now?}

Furthermore, these tests show that, when using A-GRADE, we consistently outperformed the corresponding model that uses S-GRADE in both datasets and both tasks. At the same time, the model trained only with A-GRADE synthetic data performs worse than the one trained only with S-CH real data when evaluated on CH. Still, the result is the opposite when we evaluate the models using the TH dataset, with the model trained only with A-GRADE performing significantly better (up to $\sim 25\%$) than the one trained only with S-CH data. This is true considering both tasks and both confidence thresholds. %in detecting people in CH using a threshold of 0.7 with a $\sim 14\%$ lower mAP50 and $4.6\%$ lower mAP.  the model trained with A-GRADE has a higher $\sim 20\%$ mAP50 and $\sim 25\%$ mAP . The same holds for the instance segmentation task, and also with a threshold of 0.05. 

The best model is obtained by pre-training on A-GRADE and fine-tuning on CH. This training strategy yields an improvement of $\sim 2\%$ and $3-5\%$ on CH and TUM datasets, respectively. The second-best model is the one trained on S-GRADE and fine-tuned on CH. The models trained on mixed data, in general, perform worse than the corresponding fine-tuned counterparts on the CH dataset while achieving similar results on the TUM dataset. These differences, also w.r.t. YOLO findings, are likely due to imbalanced synthetic and real data, and the scheduling which greatly affects the number of epochs. Indeed, the model trained on the mixed S-GRADE+CH data performs similarly to the model pre-trained on S-GRADE and fine-tuned on CH.

% Finally, we compare our results with PeopleSansPeople~\cite{peoplesanspeople} on the detection accuracy using a threshold of 0.05. Our training results on S-CH and CH show comparable results w.r.t. theirs on both mAP and mAP50. The differences lie likely in the dataset size when comparing S-CH and the number of training steps when considering CH. We can also see that both S-GRADE and A-GRADE greatly outperform corresponding PeopleSansPeople synthetic data with +3-8\% and +7-12\% accuracies. However, unlike them, we focus on indoor environments, use shorter training procedures, i.e. 270K iterations for A-GRADE instead of 4M of their biggest synthetic set, and do not use real-world data in validation. Then, while their increment when using a subset of CH as a fine-tuning dataset is much greater than ours, this is probably driven by the chosen validation set: in PeopleSansPeople it is the \textit{full} CH dataset, while our is still S-CH (120 images). Indeed, the improvement obtained when using the full CH is just 0.7\%, almost half of our 1.3\%. Finally, while PeopleSansPeople images were generated with the clear purpose of generalizing well in diverse data by generating images with high variability in poses, background color, and human locations, our synthetic dataset remained focused on indoor dynamic scenes.

\begin{table}[h]
    \centering
    \resizebox{\columnwidth}{!}{
\begin{tabular}{cc|cc|cc|cc|cc?cc|cc|cc|cc}
\multicolumn{2}{c|}{} & \multicolumn{8}{c?}{Threshold 0.7} & \multicolumn{8}{c}{Threshold 0.05}\\\cline{3-18}
\multicolumn{2}{c|}{}  & \multicolumn{4}{c|}{Detection} & \multicolumn{4}{c?}{Segmentation} & \multicolumn{4}{c|}{Detection} & \multicolumn{4}{c}{Segmentation} \\\cline{3-18}
\multicolumn{2}{c|}{}  & \multicolumn{2}{c|}{CH} & \multicolumn{2}{c|}{TH}  & \multicolumn{2}{c|}{CH} & \multicolumn{2}{c?}{TH} & \multicolumn{2}{c|}{CH} & \multicolumn{2}{c|}{TH}  & \multicolumn{2}{c|}{CH} & \multicolumn{2}{c}{TH} \\ \cline{3-18}
(Pre-)Training Set & Fine-Tuning Set & mAP50 & mAP & mAP50 & mAP & mAP50 & mAP & mAP50 & mAP & mAP50 & mAP & mAP50 & mAP & mAP50 & mAP & mAP50 & mAP \\ \hline

% BASELINE & --- & 0.716 & 0.495 &  0.886 & 0.716 & 0.705 &  0.432 &  0.887 &  0.674 & 0.841 & 0.556 &  0.920 &  0.736  & 0.817 & 0.479 &  0.922 &  0.692 \\ \hline
BASELINE & --- & 0.727 & 0.504 & 0.860 & 0.709 & 0.723 & 0.445 & 0.870 & 0.652 & 0.852 & 0.567 & 0.910 &  0.738  & 0.828 & 0.493 & 0.913 &  0.675 \\ \hline

CH & --- & \textit{0.693} & 0.471 & 0.829 & 0.653  & 0.681 & 0.410 & 0.838 & 0.584 & 0.829 & 0.537 & 0.898 & 0.692  & 0.801 & 0.461 & 0.890 & 0.611 \\
S-CH & --- &  0.340 &  0.161  &  0.526 &  0.250  & 0.351 & 0.155 &  0.543 & 0.231 & 0.439 & 0.195 & 0.610 &  0.282 & 0.392 & 0.168 & 0.568 & 0.241 \\
S-GRADE & --- &  0.128 & 0.064 &  0.563 &  0.312 & 0.100 & 0.043 &  0.509 & 0.264 & 0.167 & 0.077 & 0.637 & 0.343 & 0.117 & 0.048 & 0.561 & 0.283 \\
A-GRADE & --- &  0.202 & 0.115 & 0.727 & 0.502 & 0.178 & 0.088 & 0.709  & 0.408 & 0.269 & 0.140 & 0.784 & 0.531  & 0.214 & 0.100  & 0.749 & 0.425 \\ \hline
S-GRADE & S-CH &  0.428 &  0.232  &  0.708 &  0.412 & 0.401 &  0.195 &  0.665 &  0.374 & 0.518 & 0.265 &  0.748 & 0.432 & 0.465 &  0.216 &  0.694 &  0.387  \\
A-GRADE & S-CH &  0.450 & 0.262 & 0.736 & 0.489 & 0.460 & 0.231 & 0.758 & 0.449 & 0.560 & 0.303 & 0.788 & 0.515 & 0.515 & 0.247 & 0.780 & 0.458 \\
S-GRADE & CH &  \textit{0.693} &  \textit{0.474} & 0.858 &  \textit{0.679} & \textit{0.682} & \textit{0.415}  & \textit{0.858} & \textit{0.611} & \textit{0.833} & \textit{0.539} & \textit{0.916} & \textit{0.713} & \textit{0.805} & \textit{0.467} &  \textit{0.905} & \textit{0.633} \\
A-GRADE & CH &  \textbf{0.714} & \textbf{0.489} & \textbf{0.869} & \textbf{0.696} & \textbf{0.710} & \textbf{0.430} & \textbf{0.869} & \textbf{0.638} & \textbf{0.843} & \textbf{0.550} & \textit{0.916} & \textbf{0.728} & \textbf{0.813} & \textbf{0.476} & \textbf{0.908} & \textbf{0.660} \\ \hline
S-GRADE + S-CH & --- & 0.297 & 0.154 & 0.650 & 0.381 & 0.268 & 0.126 & 0.608 & 0.321 & 0.408 & 0.194 & 0.724 & 0.417 & 0.344 & 0.149 & 0.661 & 0.346 \\
A-GRADE + S-CH & --- & 0.300 & 0.168 & 0.791 & 0.561 & 0.283 & 0.138 & 0.746 & 0.467 & 0.384 & 0.200 & 0.842 & 0.588 & 0.335 & 0.155 & 0.779 & 0.483 \\
S-GRADE + CH & --- & 0.683 & 0.463 & \textit{0.859} & 0.676 & 0.671 & 0.401 & 0.849 & 0.603 & 0.821 & 0.528 & \textbf{0.917} & 0.709 & 0.790 & 0.452 & 0.896 & 0.626 \\
A-GRADE + CH & --- & 0.563 & 0.356 & 0.846 & 0.659 & 0.540 & 0.306 & 0.846 & 0.587 & 0.713 & 0.422 & 0.903 & 0.689 & 0.669 & 0.355 & 0.888 & 0.608 \\ 
    \end{tabular}}
    \caption{Mask R-CNN detection and segmentation results using both score thresholds \textbf{0.7} and \textbf{0.05}. We put in bold the best results and in italics the second best. Our baseline is the officially released model of Mask R-CNN trained on the full COCO dataset. For these tests, we also compare against the model trained on CH using our training and validation schedules for a fair comparisons.}
    \label{tab:mask}
\end{table}

\subsection*{Dynamic Visual SLAM}
\label{sec:evalSLAM}
With these evaluations, we pursue the following objectives. First, we must ensure that the synthetic data generated with GRADE can be successfully used to evaluate V-SLAM approaches. Then, we proceed to benchmark current state-of-the-art methods using simulated runs in dynamic indoor environments. Finally, we will study the impact that the performance of the underlying detection and segmentation models have on two different Dynamic V-SLAM methods using both synthetic and real-world sequences. 

The simulated data is collected by a UAV autonomously exploring an environment using the same procedure described previously. Each 60 seconds long sequence is formed by RGB-D (30 fps), IMU (240 Hz), and ground truth pose (60 Hz) data. In total, we use eight different runs. Those are divided and labeled as follows: two are recorded in static environments (\textbf{S}), two contain dynamic people and no flying objects (\textbf{D}), two have both people and flying objects (\textbf{F}), and two present an occlusion of the camera (\textbf{WO}). The occlusion creates some challenging completely black frames during the experiments. For each one of these kinds (S, D, F, WO), the UAV is either: kept horizontal, in which case it cannot perform roll and pitch movements, or is free to move, in order to increase variability in our evaluations. If the robot is not free to move we post-fix the label of the run with the letter \textbf{H}, i.e. SH, DH, FH, and WOH. To complete our evaluations, we also use our experiment repetition procedure to re-render all the dynamic sequences into their static counterparts by disabling all the dynamic assets and re-rendering RGB and depth data. We indicate those by post-fixing \textbf{-static}. This allows us to study the effects of the dynamic entities in our experiments. %We use the static sequences as a way to show that the visual information is usable as-is for the visual odometry task.

% \todo{these edits are performed AFTER the experiment collection. We collect the data in GT form (infinite depth etc), THEN we limit the depth to 3.5 and 5 meters, and use both GT data and images/depth with added noise.}
The ground-truth data generated by the simulator is then post-processed in different ways before performing the evaluations. Depth data is limited to $3.5$ meters, a reasonable value when using a RealSense D435i, and to $5$ meters (presented in the Supplementary Materials), to study the effect of the depth range on the SLAM methods. We then evaluate the SLAM frameworks using the same data but enhanced with additional noise. The noise applied to the depth values is based on the model described in~\cite{intelnoisemodel}. The RGB data is modified with random rolling shutter noise ($\mu=0.015$, $\sigma=0.006$), and blur following~\cite{deblurDatasetIMU}. The IMU drift and noise parameters are taken from~\cite{rotors}. 

We use two static V-SLAM methods and four Dynamic V-SLAM approaches. Those are: RTABMap~\cite{labbe2019rtab} and ORB-SLAMv2~\cite{murORB2}, which do not explicitly address dynamic entities, DynaSLAM~\cite{dynaslam}, which uses Mask R-CNN to segment dynamic content, DynamicVINS~\cite{dynamic-vins} (in both its VO and VIO variations, abbreviated to DynaVINS here), which uses YOLO to detect it, StaticFusion~\cite{staticfusion}, a non-learning method that performs RGB-D based clustering, and TartanVO~\cite{tartanvo}, i.e. a learned visual odometry system developed specifically for challenging scenarios. We chose not to modify parameters on any of the SLAM approaches, except for the number of extracted features in DynaSLAM and ORB-SLAMv2 which was increased to 3000, to have more stable results. We also modified the source code of DynaVINS, taking inspiration from VINSFusion~\cite{qin2019a}, to keep the system running in case of a tracking failure. Finally, we study the relationship between the performance of the detection and segmentation models and the corresponding Dynamic V-SLAM methods using our YOLO and MaskRCNN models combined with DynaVINS and DynaSLAM.

We report the absolute trajectory error (ATE)~\cite{robust-slam-survey} and the total time a considered V-SLAM framework can successfully track the trajectory. The latter, expressed as tracking rate (TR), is a critical quantity to be considered. It helps the reader put ATE values in perspective whenever the tested method fails due to some featureless frames or occlusions, as the ATE alone cannot completely quantify the robustness~\cite{robust-slam-survey}. The ATE has been computed using the standard TUM-RGBD evaluation tool. For each test, we perform ten different trials and report the mean and standard deviation of both metrics.
\begin{table*}[!ht]
\centering
\resizebox{\textwidth}{!}{
    \begin{tabular}{lr|cc|cc|cc|cc|cc|cc|cc}
     & {} & \multicolumn{2}{c|}{DynaVINS --- VO} & \multicolumn{2}{c|}{DynaVINS --- VIO} & \multicolumn{2}{c|}{StaticFusion} & \multicolumn{2}{c|}{TartanVO} & \multicolumn{2}{c|}{DynaSLAM} & \multicolumn{2}{c|}{ORB-SLAMv2} & \multicolumn{2}{c}{RTABMap} \\
    {} & {} & ATE &  TR & ATE &  TR & ATE &  TR & ATE &  TR & ATE &  TR & ATE &  TR & ATE &  TR \\ \hline
    \multirow{2}{*}{D-static} & mean & 0.366 & 0.992 &0.158 & 0.989 &2.878 & 0.999 &    1.732 &   1.000 & 0.047 & 0.985 & 0.020 &0.984 &  0.090 & 0.198 \\
        & std & 0.330 & 0.002 &0.012 & 0.000 &0.000 & 0.000 &    0.000 &   0.000 & 0.051 & 0.021 & 0.003 &0.020 &  0.000 & 0.000 \\ \hline
    \multirow{2}{*}{DH-static} & mean &  1.548 & 0.737  &6.768 & 0.993 &2.745 & 0.999 &    1.173 &   1.000 & 0.008 & 0.107 & 0.005 &0.189 &  0.086 & 0.655 \\
     & std & 0.192 & 0.220 &0.202 & 0.000 &0.000 & 0.000 &    0.000 &   0.000 & 0.006 & 0.038 & 0.000 &0.007 &  0.000 & 0.000 \\ \hline
    \multirow{2}{*}{F-static} & mean & 1.176 & 0.825 &1.344 & 0.982 &2.823 & 0.999 &    3.765 &   1.000 & 0.621 & 0.841 & 0.327 &0.904 &  0.049 & 0.317 \\
        & std & 0.404 & 0.262 &0.549 & 0.000 &0.000 & 0.000 &    0.000 &   0.000 & 0.498 & 0.035 & 0.481 &0.006 &  0.000 & 0.000 \\ \hline
      \multirow{2}{*}{FH-static} & mean & 0.020 & 0.993 &0.038 & 0.990 &0.100 & 0.999 &    0.531 &   1.000 & 0.011 & 1.000 & 0.010 &1.000 &  0.093 & 1.000 \\
      & std & 0.002 & 0.000 &0.002 & 0.001 &0.000 & 0.000 &    0.000 &   0.000 & 0.002 & 0.000 & 0.004 &0.000 &  0.000 & 0.000 \\ \hline
      \multirow{2}{*}{WO-static} & mean & 0.806 & 0.990 &0.201 & 0.976 &0.062 & 0.999 &    2.437 &   1.000 & 0.036 & 0.793 & 0.040 &0.752 &  0.120 & 1.000 \\
       & std & 0.442 & 0.003 &0.015 & 0.007 &0.000 & 0.000 &    0.000 &   0.000 & 0.029 & 0.288 & 0.025 &0.342 &  0.000 & 0.000 \\ \hline
     \multirow{2}{*}{WOH-static} & mean & 1.473 & 0.984 &0.139 & 0.985 &1.469 & 0.999 &    2.610 &   1.000 & 0.012 & 0.538 & 0.015 &0.538 &  0.208 & 1.000 \\
      & std & 0.264 & 0.008 &0.014 & 0.000 &0.000 & 0.000 &    0.000 &   0.000 & 0.004 & 0.000 & 0.009 &0.000 &  0.000 & 0.000 \\ 
    \end{tabular}
    }
\caption{
ATE RMSE [m] and tracking rate of the GRADE tested sequences (rows) in their re-rendered static versions with the experiment repetition tool. We report mean and standard deviation over ten trials. The columns indicate the evaluated method and the two metrics considered. Each experiment is 60 seconds long and the depth is limited to 3.5 m, without additional noise.}
    \label{tab:slam-3.5m-static-gt}
\end{table*}

\subsubsection*{Visual SLAM performance}
The results are presented in Tables~\ref{tab:slam-3.5m-static-gt},~\ref{tab:slam-3.5m}, and~\ref{tab:slam-5m} (in the Supplementary Materials). The first noticeable thing is how both TartanVO and StaticFusion show consistently high ATE with perfect TR. For TartanVO this can be linked to a poor generalization capability, probably due to the domain-gap with the training data and the use of only RGB information. StaticFusion, instead, is extremely sensitive to parameter tuning procedures, as shown also in other works~\cite{8613746}.
\begin{table*}[ht]
\centering
\resizebox{\textwidth}{!}{
 \begin{tabular}{llr|cc|cc|cc|cc|cc|cc|cc}
& & {} & \multicolumn{2}{c|}{DynaVINS --- VO} & \multicolumn{2}{c|}{DynaVINS --- VIO} & \multicolumn{2}{c|}{StaticFusion} & \multicolumn{2}{c|}{TartanVO} & \multicolumn{2}{c|}{DynaSLAM} & \multicolumn{2}{c|}{ORB-SLAMv2} & \multicolumn{2}{c}{RTABMap} \\
& {} & {} & ATE & TR & ATE & TR & ATE & TR & ATE & TR & ATE & TR & ATE & TR & ATE & TR \\ \hline
 
\multirow{16}{*}{\begin{sideways}Ground-truth data\end{sideways}} & \multirow{2}{*}{D} & 
 mean & 1.450 & 0.888 & 0.192 & 0.989 & 1.212 & 0.999 & 1.264 & 1.000 & 0.042 & 0.830 & 0.283 & 0.981 & 0.417 & 0.891 \\
 & & std & 0.441 & 0.096 & 0.010 & 0.000 & 0.000 & 0.000 & 0.000 & 0.000 & 0.009 & 0.111 & 0.036 & 0.021 & 0.000 & 0.000 \\ \cline{2-17}
 & \multirow{2}{*}{DH} & mean & 1.582 & 0.644 & 8.020 & 0.993 & 1.664 & 0.999 & 1.259 & 1.000 & 0.011 & 0.097 & 0.005 & 0.179 & 0.091 & 0.654 \\
 & & std & 0.468 & 0.301 & 0.282 & 0.000 & 0.000 & 0.000 & 0.000 & 0.000 & 0.007 & 0.028 & 0.001 & 0.007 & 0.000 & 0.000 \\ \cline{2-17}
& \multirow{2}{*}{F} & mean & 1.532 & 0.841 & 2.057 & 0.980 & 2.866 & 0.999 & 4.132 & 1.000 & 0.858 & 0.440 & 0.294 & 0.565 & 0.086 & 0.219 \\
 & & std & 0.504 & 0.230 & 0.478 & 0.001 & 0.000 & 0.000 & 0.000 & 0.000 & 0.184 & 0.115 & 0.151 & 0.228 & 0.000 & 0.000 \\ \cline{2-17}
& \multirow{2}{*}{FH} & mean & 0.220 & 0.993 & 0.075 & 0.989 & 0.085 & 0.999 & 0.551 & 1.000 & 0.258 & 1.000 & 0.295 & 1.000 & 0.324 & 1.000 \\
 & & std & 0.058 & 0.000 & 0.006 & 0.000 & 0.000 & 0.000 & 0.000 & 0.000 & 0.054 & 0.000 & 0.067 & 0.000 & 0.000 & 0.000 \\ \cline{2-17}
 & \multirow{2}{*}{WO} & mean & 1.219 & 0.910 & 0.582 & 0.957 & 2.807 & 0.999 & 2.473 & 1.000 & 0.090 & 0.079 & 0.157 & 0.197 & 0.275 & 0.197 \\
 & & std & 0.291 & 0.092 & 0.072 & 0.003 & 0.000 & 0.000 & 0.000 & 0.000 & 0.011 & 0.000 & 0.007 & 0.000 & 0.000 & 0.000 \\ \cline{2-17}
 & \multirow{2}{*}{WOH} & mean & 1.474 & 0.808 & 0.223 & 0.981 & 1.980 & 0.999 & 2.361 & 1.000 & 0.013 & 0.538 & 0.011 & 0.538 & 0.088 & 0.569 \\
 & & std & 0.548 & 0.120 & 0.027 & 0.002 & 0.000 & 0.000 & 0.000 & 0.000 & 0.002 & 0.000 & 0.002 & 0.000 & 0.000 & 0.000 \\ \cline{2-17}
 & \multirow{2}{*}{S} & mean & 0.036 & 0.993 & 0.222 & 0.991 & 7.919 & 0.999 & 1.205 & 1.000 & 0.011 & 1.000 & 0.011 & 1.000 & 0.084 & 1.000 \\
 & & std & 0.003 & 0.000 & 0.010 & 0.000 & 0.000 & 0.000 & 0.000 & 0.000 & 0.001 & 0.000 & 0.001 & 0.000 & 0.000 & 0.000 \\\cline{2-17}
 & \multirow{2}{*}{SH} & mean & 0.029 & 0.993 & 0.119 & 0.991 & 0.594 & 0.999 & 2.395 & 1.000 & 0.011 & 1.000 & 0.012 & 1.000 & 0.089 & 1.000 \\
 & & std & 0.005 & 0.000 & 0.007 & 0.000 & 0.000 & 0.000 & 0.000 & 0.000 & 0.003 & 0.000 & 0.002 & 0.000 & 0.000 & 0.000 \\

\midrule

\multirow{16}{*}{\begin{sideways}Noisy data\end{sideways}} & \multirow{2}{*}{D} & 
 mean & 1.362 & 0.899 & 0.693 & 0.989 & 2.278 & 0.999 & 1.356 & 1.000 & 0.061 & 0.554 & 0.725 & 0.877 & 0.401 & 0.606 \\
& & std & 0.355 & 0.090 & 0.086 & 0.000 & 0.000 & 0.000 & 0.000 & 0.000 & 0.033 & 0.002 & 0.052 & 0.035 & 0.000 & 0.000 \\ \cline{2-17}
 & \multirow{2}{*}{DH} & mean & 1.628 & 0.609 & 1.982 & 0.993 & 1.091 & 0.999 & 1.234 & 1.000 & 0.003 & 0.051 & 0.004 & 0.054 & 0.052 & 0.175 \\
 & & std & 0.642 & 0.338 & 0.268 & 0.000 & 0.000 & 0.000 & 0.000 & 0.000 & 0.001 & 0.000 & 0.000 & 0.000 & 0.000 & 0.000 \\ \cline{2-17}
 &\multirow{2}{*}{F} & mean & 2.039 & 0.855 & 2.431 & 0.975 & 3.992 & 0.999 & 4.223 & 1.000 & 0.212 & 0.255 & 0.142 & 0.258 & 0.047 & 0.191 \\
 & & std & 0.575 & 0.193 & 1.616 & 0.002 & 0.000 & 0.000 & 0.000 & 0.000 & 0.037 & 0.037 & 0.017 & 0.028 & 0.000 & 0.000 \\ \cline{2-17}
 & \multirow{2}{*}{FH} & mean & 0.455 & 0.991 & 0.207 & 0.988 & 0.854 & 0.999 & 0.582 & 1.000 & 0.244 & 0.986 & 0.240 & 1.000 & 0.184 & 1.000 \\
 & & std & 0.199 & 0.003 & 0.104 & 0.001 & 0.000 & 0.000 & 0.000 & 0.000 & 0.062 & 0.042 & 0.081 & 0.000 & 0.000 & 0.000 \\ \cline{2-17}
 & \multirow{2}{*}{WO} & mean & 1.219 & 0.844 & 1.085 & 0.955 & 2.213 & 0.999 & 2.380 & 1.000 & 0.099 & 0.079 & 0.181 & 0.197 & 0.329 & 0.197 \\
 & & std & 0.116 & 0.207 & 0.275 & 0.002 & 0.000 & 0.000 & 0.000 & 0.000 & 0.009 & 0.000 & 0.006 & 0.000 & 0.000 & 0.000 \\ \cline{2-17}
 & \multirow{2}{*}{WOH} & mean & 1.364 & 0.910 & 0.560 & 0.981 & 1.826 & 0.999 & 2.399 & 1.000 & 0.032 & 0.536 & 0.021 & 0.536 & 0.118 & 0.569 \\
 & & std & 0.541 & 0.115 & 0.086 & 0.001 & 0.000 & 0.000 & 0.000 & 0.000 & 0.011 & 0.000 & 0.002 & 0.000 & 0.000 & 0.000 \\\cline{2-17}
 & \multirow{2}{*}{S} & mean & 0.067 & 0.993 & 0.200 & 0.991 & 3.538 & 0.999 & 1.306 & 1.000 & 0.022 & 1.000 & 0.024 & 1.000 & 0.212 & 1.000 \\
 & & std & 0.007 & 0.000 & 0.010 & 0.000 & 0.000 & 0.000 & 0.000 & 0.000 & 0.001 & 0.000 & 0.001 & 0.000 & 0.000 & 0.000 \\\cline{2-17}
 & \multirow{2}{*}{SH} & mean & 0.073 & 0.993 & 0.693 & 0.991 & 4.184 & 0.999 & 2.517 & 1.000 & 0.017 & 1.000 & 0.018 & 1.000 & 0.092 & 1.000 \\
 & & std & 0.008 & 0.000 & 0.408 & 0.000 & 0.000 & 0.000 & 0.000 & 0.000 & 0.001 & 0.000 & 0.002 & 0.000 & 0.000 & 0.000 \\

\end{tabular}
}

\caption{ATE RMSE [m] and tracking rate of the tested sequences (rows) in both their ground-truth and noisy versions. The ground-truth sequences are reported in the upper half of the table, while the sequences with added noise are in the bottom half. We report mean and standard deviation over ten trials. The columns indicate the evaluated method and the two metrics considered. Each experiment is 60 seconds long and the depth is limited to 3.5 m.}
\label{tab:slam-3.5m}
\end{table*}

The results obtained by testing the selected methods on the static sequences are presented in Tab.~\ref{tab:slam-3.5m-static-gt}, i.e. the re-rendered ones, and in Tab.~\ref{tab:slam-3.5m} and Tab.~\ref{tab:slam-5m} (S and SH). We can observe how, generally, all the methods perform well in SH and S, with low ATE and high tracking rates. At the same time, on the other sequences, there is high variability in both metrics. However, at least one of the selected methods for each experiment shows good performance. This, while showing that the rendered data can be used effectively to perform visual odometry, demonstrates also the low adaptation capabilities of some of these algorithms. The low tracking rates of RTABMap on D-static, DH-static, and F-static are to be associated with events in which the system loses track of the odometry and resets, without recovering. Notably, while both ATE and TR vary across methods, standard deviations are generally low.

In general, one should not be misled by the good ATE results of some methods. For example, using DynaSLAM, in four out of eight ground-truth 3.5m limited depth sequences (Tab.~\ref{tab:slam-3.5m}), the method loses track of the trajectory for at least $\sim27$ seconds, and up to $54$ seconds. In general, despite these methods showing compelling results when tested with common datasets like TUM-RGBD or EuRoC, they exhibit several limitations when tested on our data. We can see that the majority of the experiments on noisy data are, as expected, slightly worse than the ones performed on ground truth data. Interestingly, with some combinations, the SLAM methods perform worse in the statically rendered sequences than in the dynamic ones. For example, RTABMap performs better on the D dynamic sequence than on D-static run. Probably, this is because, in some situations, the methods are capable of leveraging features lying over dynamic content to better track the camera movement. ORB-SLAM and DynaSLAM are, for the most part, comparable. DynaVINS VO often performs worse than its VIO counterpart, indicating the reliance on the IMU sensor. StaticFusion shows degrading performance with increased depth data. DynaVINS seems overall the most stable, without suffering depth ranges or noisiness of the data, although it has higher deviations. Overall, DynaVINS VIO is the best-performing method when considering both ATE and TR. However, despite the use of the IMU, the ATE obtained with DynaVINS VIO on the DH non-noisy sequence reported in Tab.~\ref{tab:slam-3.5m} is over $8$ meters for just a 60 s sequence, while for the VO counterpart it is just 1.582 meters. Similar results can be observed in the S and SH sequences, where DynaVINS VO shows 3 to 10 times lower ATE than the one obtained with DynaVINS VIO, indicating how the method does not always benefit from using the IMU sensor.

\subsubsection*{Study on the Deep Learning-SLAM relation}
Here, we evaluate DynaVINS (VO) and DynaSLAM jointly with some of the models we obtained in the previous section from the training of YOLOv5 and Mask R-CNN with synthetic and real-world data. We evaluate their performance on both the synthetic sequences with the depth limited to 3.5 meters of GRADE (without noise) and the TUM \textit{fr3/walking} sequences~\cite{turgbd}. The results are presented in Tab.~\ref{tab:both-nets}. For Mask R-CNN we use the best-performing model on the \textit{segmentation} task. The baseline results on the TUM-RGBD sequences are obtained using the baseline models. We were unable to reproduce the published results for the \textit{rpy} and \textit{static} sequences for DynaVINS and the \textit{walking\_rpy} sequence for DynaSLAM.
\begin{table*}[ht]
\centering
\resizebox{\textwidth}{!}{
\begin{tabular}{lllr|cc|cc|cc|cc|cc|cc|cc|cc|cc|cc|cc|cc}
{} & {} & {} & {Train} & \multicolumn{2}{c|}{S-GRADE+S-CH} & \multicolumn{2}{c|}{A-GRADE} & \multicolumn{2}{c|}{S-GRADE} & \multicolumn{2}{c|}{S-GRADE+CH} & \multicolumn{2}{c|}{S-GRADE} & \multicolumn{2}{c|}{A-GRADE} & \multicolumn{2}{c|}{A-GRADE+CH} & \multicolumn{2}{c|}{A-GRADE} & \multicolumn{2}{c|}{A-GRADE+S-CH} & \multicolumn{2}{c|}{S-GRADE} & \multicolumn{2}{c|}{S-CH} & \multicolumn{2}{c}{\multirow{2}{*}{Baseline}} \\
{} & {} & {} & {Finetuned} & \multicolumn{2}{c|}{} & \multicolumn{2}{c|}{} & \multicolumn{2}{c|}{} & \multicolumn{2}{c|}{} & \multicolumn{2}{c|}{CH} & \multicolumn{2}{c|}{S-CH} & \multicolumn{2}{c|}{} & \multicolumn{2}{c|}{CH} & \multicolumn{2}{c|}{} & \multicolumn{2}{c|}{S-CH} & \multicolumn{2}{c|}{} & {} & {} \\\hline
{} & {} & {}& {}& ATE & TR & ATE & TR & ATE & TR & ATE & TR & ATE & TR & ATE & TR & ATE & TR & ATE & TR & ATE & TR & ATE & TR & ATE & TR& ATE & TR \\ \hline

\multirow{16}{*}{\begin{sideways}{TUM RGBD}\end{sideways}} & \multirow{8}{*}{\begin{sideways}{DynaVINS --- VO}\end{sideways}} 
 & \multirow{2}{*}{halfsphere} & mean & 0.074 & 0.968 & 0.062 & 0.968 & 0.051 & 0.968 & 0.063 & 0.968 & 0.064 & 0.968 & 0.100 & 0.967 & 0.060 & 0.968 & 0.069 & 0.968 & 0.064 & 0.968 & 0.072 & 0.967 & 0.148 & 0.966 & 0.097 & 0.967 \\
 & & & std & 0.022 & 0.000 & 0.026 & 0.000 & 0.008 & 0.000 & 0.007 & 0.000 & 0.013 & 0.000 & 0.074 & 0.004 & 0.006 & 0.000 & 0.020 & 0.000 & 0.023 & 0.000 & 0.034 & 0.003 & 0.178 & 0.004 & 0.080 & 0.004 \\
 & & \multirow{2}{*}{rpy} & mean & 0.166 & 0.975 & 0.155 & 0.975 & 0.230 & 0.975 & 0.129 & 0.974 & 0.169 & 0.967 & 0.108 & 0.975 & 0.142 & 0.972 & 0.115 & 0.972 & 0.143 & 0.975 & 0.131 & 0.975 & 0.136 & 0.975 & 0.130 & 0.972 \\
 & & & std & 0.028 & 0.000 & 0.018 & 0.000 & 0.023 & 0.000 & 0.018 & 0.004 & 0.061 & 0.008 & 0.015 & 0.000 & 0.048 & 0.006 & 0.016 & 0.005 & 0.020 & 0.000 & 0.012 & 0.000 & 0.030 & 0.000 & 0.027 & 0.005 \\
 & & \multirow{2}{*}{static} & mean & 0.357 & 0.981 & 0.135 & 0.972 & 0.279 & 0.978 & 0.050 & 0.970 & 0.141 & 0.972 & 0.091 & 0.972 & 0.123 & 0.855 & 0.071 & 0.970 & 0.212 & 0.980 & 0.325 & 0.976 & 0.291 & 0.919 & 0.542 & 0.968 \\
 & & & std & 0.373 & 0.005 & 0.133 & 0.007 & 0.188 & 0.007 & 0.021 & 0.006 & 0.182 & 0.007 & 0.071 & 0.007 & 0.072 & 0.189 & 0.054 & 0.006 & 0.065 & 0.007 & 0.268 & 0.008 & 0.142 & 0.115 & 0.757 & 0.025 \\
 & & \multirow{2}{*}{xyz} & mean & 0.049 & 0.970 & 0.039 & 0.970 & 0.050 & 0.970 & 0.044 & 0.970 & 0.045 & 0.970 & 0.041 & 0.970 & 0.043 & 0.970 & 0.046 & 0.970 & 0.047 & 0.970 & 0.044 & 0.970 & 0.040 & 0.849 & 0.055 & 0.970 \\
 & & & std & 0.013 & 0.000 & 0.006 & 0.000 & 0.010 & 0.000 & 0.010 & 0.000 & 0.010 & 0.000 & 0.005 & 0.000 & 0.005 & 0.000 & 0.011 & 0.000 & 0.008 & 0.000 & 0.008 & 0.000 & 0.007 & 0.241 & 0.014 & 0.000 \\
\cline{2-28}
& \multirow{8}{*}{\begin{sideways}
 DynaSLAM
\end{sideways}} & \multirow{2}{*}{halfsphere} & mean & 0.031 & 1.000 & 0.029 & 1.000 & 0.029 & 0.999 & 0.029 & 0.999 & 0.030 & 0.999 & 0.029 & 1.000 & 0.029 & 0.999 & 0.028 & 1.000 & 0.029 & 1.000 & 0.031 & 0.971 & 0.031 & 0.999 & 0.028 & 1.000 \\
 & & & std & 0.002 & 0.000 & 0.002 & 0.000 & 0.001 & 0.003 & 0.002 & 0.002 & 0.001 & 0.003 & 0.001 & 0.001 & 0.001 & 0.003 & 0.001 & 0.000 & 0.001 & 0.001 & 0.002 & 0.049 & 0.001 & 0.002 & 0.002 & 0.001 \\
 & & \multirow{2}{*}{rpy} & mean & 0.063 & 0.939 & 0.048 & 0.974 & 0.100 & 0.972 & 0.037 & 0.865 & 0.037 & 0.855 & 0.035 & 0.874 & 0.036 & 0.897 & 0.041 & 0.860 & 0.042 & 0.896 & 0.038 & 0.831 & 0.039 & 0.833 & 0.035 & 0.846 \\
 & & & std & 0.047 & 0.029 & 0.015 & 0.023 & 0.021 & 0.003 & 0.004 & 0.039 & 0.004 & 0.021 & 0.003 & 0.019 & 0.006 & 0.034 & 0.011 & 0.019 & 0.005 & 0.042 & 0.003 & 0.025 & 0.005 & 0.043 & 0.006 & 0.022 \\
 & & \multirow{2}{*}{static} & mean & 0.007 & 1.000 & 0.006 & 1.000 & 0.007 & 0.997 & 0.007 & 0.850 & 0.007 & 0.839 & 0.007 & 0.974 & 0.008 & 0.971 & 0.007 & 0.839 & 0.007 & 0.998 & 0.007 & 1.000 & 0.011 & 1.000 & 0.007 & 0.957 \\
 & & & std & 0.000 & 0.000 & 0.000 & 0.000 & 0.000 & 0.000 & 0.000 & 0.036 & 0.000 & 0.000 & 0.000 & 0.001 & 0.000 & 0.000 & 0.001 & 0.000 & 0.000 & 0.001 & 0.000 & 0.000 & 0.002 & 0.000 & 0.000 & 0.002 \\
 & & \multirow{2}{*}{xyz} & mean & 0.016 & 0.979 & 0.016 & 0.956 & 0.016 & 0.980 & 0.016 & 0.915 & 0.016 & 0.915 & 0.016 & 0.922 & 0.016 & 0.915 & 0.015 & 0.915 & 0.016 & 0.915 & 0.016 & 1.000 & 0.016 & 0.979 & 0.015 & 0.915 \\
 & & & std & 0.001 & 0.000 & 0.001 & 0.000 & 0.000 & 0.015 & 0.000 & 0.000 & 0.001 & 0.000 & 0.000 & 0.000 & 0.001 & 0.000 & 0.001 & 0.000 & 0.001 & 0.000 & 0.001 & 0.000 & 0.000 & 0.024 & 0.000 & 0.001 \\ \midrule

\multirow{24}{*}{\begin{sideways}{GRADE}\end{sideways}} & \multirow{12}{*}{\begin{sideways}{DynaVINS --- VO}\end{sideways}} 
& \multirow{2}{*}{D} & mean & 1.363 & 0.657 & 1.497 & 0.917 & 1.410 & 0.808 & 1.186 & 0.893 & 1.382 & 0.926 & 1.504 & 0.818 & 1.488 & 0.927 & 1.488 & 0.803 & 1.345 & 0.865 & 1.403 & 0.945 & 1.348 & 0.793 & 1.450 & 0.888 \\
& & & std & 0.331 & 0.277 & 0.315 & 0.085 & 0.259 & 0.220 & 0.448 & 0.097 & 0.273 & 0.090 & 0.280 & 0.226 & 0.391 & 0.089 & 0.250 & 0.303 & 0.323 & 0.232 & 0.260 & 0.075 & 0.346 & 0.297 & 0.441 & 0.096 \\
& & \multirow{2}{*}{DH} & mean & 1.621 & 0.712 & 1.368 & 0.691 & 1.501 & 0.770 & 2.330 & 0.416 & 1.348 & 0.590 & 1.412 & 0.685 & 1.380 & 0.659 & 2.109 & 0.577 & 2.296 & 0.676 & 1.949 & 0.598 & 1.345 & 0.640 & 1.582 & 0.644 \\
& & & std & 0.228 & 0.247 & 0.271 & 0.174 & 0.501 & 0.201 & 2.139 & 0.287 & 0.267 & 0.265 & 0.343 & 0.280 & 0.405 & 0.256 & 2.523 & 0.208 & 2.143 & 0.252 & 1.690 & 0.281 & 0.222 & 0.218 & 0.468 & 0.301 \\
& & \multirow{2}{*}{F} & mean & 1.490 & 0.746 & 1.701 & 0.820 & 1.628 & 0.755 & 1.743 & 0.741 & 1.688 & 0.721 & 1.571 & 0.946 & 1.723 & 0.819 & 1.511 & 0.743 & 1.794 & 0.807 & 1.747 & 0.825 & 1.604 & 0.880 & 1.532 & 0.841 \\
& & & std & 0.200 & 0.324 & 0.480 & 0.266 & 0.524 & 0.322 & 0.366 & 0.310 & 0.439 & 0.325 & 0.607 & 0.063 & 0.376 & 0.283 & 0.310 & 0.332 & 0.319 & 0.276 & 0.351 & 0.254 & 0.311 & 0.230 & 0.504 & 0.230 \\
& & \multirow{2}{*}{FH} & mean & 0.382 & 0.991 & 0.355 & 0.991 & 0.541 & 0.991 & 0.236 & 0.993 & 0.287 & 0.992 & 0.233 & 0.992 & 0.404 & 0.991 & 0.414 & 0.963 & 0.221 & 0.948 & 0.263 & 0.966 & 0.258 & 0.992 & 0.220 & 0.993 \\
& & & std & 0.259 & 0.003 & 0.209 & 0.003 & 0.374 & 0.003 & 0.073 & 0.000 & 0.218 & 0.002 & 0.154 & 0.002 & 0.295 & 0.003 & 0.214 & 0.081 & 0.057 & 0.091 & 0.107 & 0.079 & 0.219 & 0.002 & 0.058 & 0.000 \\
& & \multirow{2}{*}{WO} & mean & 1.130 & 0.814 & 1.186 & 0.659 & 1.344 & 0.839 & 1.334 & 0.864 & 1.199 & 0.755 & 1.184 & 0.766 & 1.331 & 0.855 & 1.279 & 0.790 & 1.195 & 0.857 & 1.469 & 0.772 & 1.473 & 0.887 & 1.219 & 0.910 \\
& & & std & 0.242 & 0.287 & 0.294 & 0.278 & 0.303 & 0.208 & 0.433 & 0.207 & 0.198 & 0.304 & 0.236 & 0.289 & 0.506 & 0.229 & 0.445 & 0.202 & 0.225 & 0.203 & 0.399 & 0.264 & 0.289 & 0.102 & 0.291 & 0.092 \\
& & \multirow{2}{*}{WOH} & mean & 1.389 & 0.847 & 1.186 & 0.878 & 1.324 & 0.904 & 1.492 & 0.874 & 1.935 & 0.937 & 1.710 & 0.928 & 1.575 & 0.807 & 1.283 & 0.838 & 1.195 & 0.867 & 1.458 & 0.862 & 1.529 & 0.897 & 1.474 & 0.808 \\
& & & std & 0.522 & 0.146 & 0.506 & 0.104 & 0.340 & 0.143 & 0.572 & 0.139 & 1.117 & 0.075 & 0.509 & 0.071 & 0.725 & 0.158 & 0.479 & 0.156 & 0.415 & 0.177 & 0.429 & 0.162 & 0.418 & 0.137 & 0.548 & 0.120 \\
 \cline{2-28}
& \multirow{12}{*}{\begin{sideways}{DynaSLAM}\end{sideways}} 
 & \multirow{2}{*}{D} & mean & 0.070 & 0.837 & 0.040 & 0.884 & 0.044 & 0.816 & 0.042 & 0.870 & 0.056 & 0.672 & 0.037 & 0.833 & 0.074 & 0.867 & 0.046 & 0.738 & 0.066 & 0.856 & 0.026 & 0.462 & 0.114 & 0.434 & 0.042 & 0.830 \\
 & & & std & 0.049 & 0.085 & 0.004 & 0.037 & 0.008 & 0.058 & 0.007 & 0.026 & 0.044 & 0.098 & 0.006 & 0.100 & 0.077 & 0.083 & 0.011 & 0.054 & 0.071 & 0.110 & 0.006 & 0.064 & 0.105 & 0.113 & 0.009 & 0.111 \\
 & & \multirow{2}{*}{DH} & mean & 0.011 & 0.091 & 0.013 & 0.097 & 0.012 & 0.094 & 0.007 & 0.086 & 0.012 & 0.096 & 0.008 & 0.090 & 0.016 & 0.099 & 0.015 & 0.090 & 0.015 & 0.099 & 0.020 & 0.079 & 0.014 & 0.079 & 0.011 & 0.097 \\
 & & & std & 0.011 & 0.025 & 0.008 & 0.026 & 0.009 & 0.023 & 0.005 & 0.030 & 0.008 & 0.032 & 0.006 & 0.023 & 0.012 & 0.026 & 0.014 & 0.033 & 0.009 & 0.020 & 0.020 & 0.023 & 0.010 & 0.023 & 0.007 & 0.028 \\
 & & \multirow{2}{*}{F} & mean & 0.789 & 0.523 & 0.676 & 0.408 & 0.592 & 0.426 & 0.587 & 0.371 & 0.843 & 0.593 & 0.581 & 0.450 & 0.692 & 0.447 & 0.851 & 0.590 & 0.803 & 0.400 & 0.858 & 0.484 & 0.486 & 0.260 & 0.858 & 0.440 \\
 & & & std & 0.271 & 0.165 & 0.305 & 0.160 & 0.317 & 0.230 & 0.325 & 0.122 & 0.013 & 0.141 & 0.347 & 0.136 & 0.298 & 0.193 & 0.036 & 0.142 & 0.232 & 0.156 & 0.053 & 0.140 & 0.232 & 0.153 & 0.184 & 0.115 \\
 & & \multirow{2}{*}{FH} & mean & 0.193 & 0.987 & 0.250 & 0.999 & 0.177 & 1.000 & 0.274 & 0.995 & 0.211 & 1.000 & 0.219 & 1.000 & 0.304 & 1.000 & 0.209 & 0.999 & 0.242 & 1.000 & 0.054 & 0.455 & 0.215 & 0.946 & 0.258 & 1.000 \\
 & & & std & 0.063 & 0.025 & 0.061 & 0.001 & 0.039 & 0.000 & 0.066 & 0.008 & 0.034 & 0.000 & 0.029 & 0.000 & 0.228 & 0.000 & 0.030 & 0.002 & 0.067 & 0.001 & 0.023 & 0.266 & 0.071 & 0.085 & 0.054 & 0.000 \\
 & & \multirow{2}{*}{WO} & mean & 0.071 & 0.079 & 0.090 & 0.079 & 0.087 & 0.079 & 0.083 & 0.079 & 0.080 & 0.079 & 0.087 & 0.079 & 0.091 & 0.079 & 0.083 & 0.079 & 0.086 & 0.079 & 0.069 & 0.079 & 0.087 & 0.097 & 0.090 & 0.079 \\
 & & & std & 0.005 & 0.000 & 0.006 & 0.000 & 0.010 & 0.000 & 0.012 & 0.000 & 0.009 & 0.000 & 0.008 & 0.000 & 0.012 & 0.000 & 0.013 & 0.000 & 0.014 & 0.000 & 0.011 & 0.000 & 0.020 & 0.000 & 0.011 & 0.000 \\
 & & \multirow{2}{*}{WOH} & mean & 0.014 & 0.538 & 0.012 & 0.538 & 0.011 & 0.535 & 0.013 & 0.538 & 0.014 & 0.537 & 0.013 & 0.526 & 0.012 & 0.538 & 0.014 & 0.538 & 0.012 & 0.538 & 0.043 & 0.104 & 0.017 & 0.520 & 0.013 & 0.538 \\
 & & & std & 0.001 & 0.000 & 0.002 & 0.000 & 0.002 & 0.000 & 0.001 & 0.000 & 0.003 & 0.000 & 0.003 & 0.003 & 0.002 & 0.000 & 0.002 & 0.000 & 0.001 & 0.000 & 0.015 & 0.086 & 0.007 & 0.003 & 0.002 & 0.000 \\
\end{tabular}}
\caption{Evaluations performed using DynaVINS and DynaSLAM while varying the model checkpoint used by the underlying YOLO and MaskRCNN networks (columns). In each row we report the mean and standard deviation of ATE RMSE [m] and tracking rate (TR) over ten trials on the walking sequences of the TUM RGBD dataset and on the GRADE ground-truth sequences with depth limited to 3.5 meters.}
\label{tab:both-nets}
\end{table*}

The results on the TUM data show that changing the model for DynaVINS does not influence the amount of tracked trajectory. DynaSLAM instead, is highly affected by the model used. Surprisingly, many trials show TR that are on par or better than the baseline, despite the underlying model showing lower segmentation performance. For example, by using the network pre-trained on S-GRADE and fine-tuned on S-CH we can reach a tracking time $8.5\%$ higher on the \textit{xyz} sequence. For both SLAM frameworks, the ATE varies based on the model used. However, the best-performing networks are often not associated with the best ATE and TR couple. For example, using the model trained only with S-GRADE, shows already compelling results. When used with DynaVINS it shows the best performance on \textit{halfsphere}, halving the ATE w.r.t. to the baseline. With DynaSLAM, using Mask R-CNN trained on S-GRADE shows ATEs comparable to the baseline results, but with higher tracking times, thus with better overall performance. At the same time, using DynaSLAM on the \textit{rpy} sequence with the model pre-trained on A-GRADE and fine-tuned on CH, i.e. the best performing one, significantly degrades the tracking rate of about $5\%$. On GRADE using different models instead influences both TR and ATE. However, also in this case, there is not a model giving a clear advantage. Still, models performing poorly on the detection and segmentation evaluations are capable of attaining higher TR and ATE.

\section*{Discussion}
% \todo{I'll do this last thing}
% \textit{Also in this case, the tight margin existing between the models trained from scratch solely on synthetic data evaluated in the close indoor domain of TH, shows how the data generated with this strategy can be used to close the syn-to-real gap.}

% \textit{Our results indicate that models trained \textit{only} with our synthetic data have good generalization capabilities on real-world images. Notably, this is possible without any error-prone and time-consuming labeling or manual data collection procedures. Moreover, we provide evidence that using our synthetic data alongside real-world images during training can significantly improve both detection and segmentation results.
% }

GRADE is a novel, versatile, and flexible tool that provides a streamlined approach to online, offline, and repeatable testing and benchmarking of robotics and vision-based methods, producing valuable learning data through a photorealistic engine. In this study, through simulations that closely replicate real-world conditions in both physics and visualization, we reduce the gap between simulation and reality. %We do this by addressing the limitations of existing simulation engines in robotics by providing full simulation control, ROS integration, realistic physics, and photorealism. 
Our customizable framework, built over Isaac Sim, allows for the seamless preparation of assets and the simulation of realistic dynamic environments. This is through a system that allows different control modalities, diverse environment sources, multi-robot scenarios, and low-level simulation control through the same software suite. %We evaluate GRADE in its flexibility, visual realism, and by testing dynamic SLAM algorithms on the data generated. , allowing for the generation of ground truth data, online testing of robots, and post-processing of the produced information
GRADE can be used in different scenarios and adapted successfully for diverse requirements, e.g. to simulate different natively unsupported platforms such as UAVs and three-wheeled omnidirectional robots, as a rendering engine by dismissing robot control, or to test robotics algorithms both online (active SLAM) and offline even in \textit{heterogeneous} multi-robot scenarios. 

Our thorough testing on a pool of state-of-the-art dynamic SLAM methods, both using synthetic and real data, shows the over-reliance of state-of-the-art methods on currently available benchmark datasets and tools. Moreover, we indicate the necessity of reporting the tracking rate alongside the results, as the ATE alone may mislead evaluations. Both aspects become clear after analyzing the extensive experiments we conducted that show how these methods fail either in \textit{correctly} or \textit{completely} tracking trajectories. Those tests showed the limitations of such systems which struggled to generalize well to sequences that are, in principle, usable. Moreover, the testing done on the TUM-RGBD using our trained models exemplifies the necessity of more thorough evaluations and studies in dynamic SLAM, as not always the best-performing human detection or segmentation models yield the best result. For example, better feature-rejection procedures might be useful to improve both the ATE and tracking times of current methods. Overall, better failure recovery procedures and processing of the dynamics in the scene are necessities that are clearly under-explored due to the limitations that existed in old simulation tools and benchmarking datasets, but that are now possible thanks to GRADE.

%, with ours lacking crowded scenes, outdoor scenarios with humans placed far in the background, or clothing variations in the assets we use, i.e. we do not have humans wearing ski suits or helmets or inside buses.
Moreover, the good sim-to-real performance on learned human detection and segmentation tasks shows the effectiveness of the simulation and its visual realism. Through extensive testing on human detection and segmentation, we demonstrate that the generated visual data is realistic enough to train models from scratch, without needing to gather and manually label images. Moreover, the data generated with GRADE, despite the minor effort we put into having hyperrealistic humans with hair, shoes, high-quality textures, or diverse settings, has a considerable impact on both tested networks when used as pre-training data. The results obtained on the two (S-)CH datasets are, in our opinion, mostly driven by the diverse distribution of the data. Nonetheless, we see how our synthetic data generalize well to the real world if we consider the TUM-RGBD dataset. Also, a better-balanced choice of training and validation sets when using S-CH or CH in the mixed models, as well as a better separation of training and validation data on both S-GRADE and GRADE will probably boost the overall performance. 

Surely, adopting commercial solutions for environments and/or dynamic humans will greatly increase the quality of the generated data and, as shown also recently from~\cite{hssd}, will also improve the performance on the learning tasks. Moreover, additional characteristics such as dirtiness of the materials, or fluid and clothing simulations, can be implemented by the end user by integrating specialized systems such as OmniGibson~\cite{li2022behavior}, or by developing custom additional tools either as SIL or based on the Omniverse suite. This is inherently possible in GRADE thanks to the design choices we made in this work, including allowing direct simulation control. 

Additionally, to the best of our knowledge, GRADE introduces the first method for precise programmatic experiment repetition with adaptable surroundings. We believe this is an important step towards better testing and adoption possibilities as data can now be independently modified, explored, or expanded by third parties in simple and effective ways. Finally, different from previous systems such as OmniGibson and BenchBot, GRADE is not focused on providing a closed system for a specific application but is built as close as possible to the low-level APIs of Isaac Sim allowing for a finer control over the experiments. Overall, GRADE provides a comprehensive solution for simulating intelligent robots in dynamic environments, enabling safer and more efficient research and development of robotics systems, and does that through an open-source code.

In short, we provide, thoroughly test, and review a novel framework that supports the inclusion of dynamic animated models, diverse rigid objects, and various environments through the exploitation, integration, and expansions of the capabilities of Isaac Sim via automatic placement procedures, fine simulation control, additional (external) tools to convert (dynamic) assets, and automatic data processing and evaluation procedures. This approach allowed us to employ GRADE in various published and ongoing research related to visual perception and robotics, i.e. detection~\cite{bonetto_detection} and pose estimation~\cite{zebrapose} of Zebras in wild environments and Dynamic V-SLAM~\cite{dynapix}.

\section*{Materials and Methods}
\label{sec:approach}
% To have the necessary flexibility to simulate (dynamic) environments for robotics, we first need a set of assets, a placement procedure for the (pre-)animated objects, and the possibility to randomize their appearance. Then, we must be able to load one or more robots with their specific sensor suites and the ability to control them. Finally, the simulation should be physically enabled and, when visual perception is required, realistically rendered, to close the sim-to-real gap. Ideally, the simulation software itself should be fully controllable and customizable, while maintaining the ability to save and process the generated richly-annotated ground-truth data. In this work, we address all these aspects by creating APIs and tools to ease and customize each step of the pipeline. To do so, we leverage Isaac Sim and the Omniverse suite. Following the logical steps also depicted in Fig.~\ref{fig:pipeline}, we proceed to describe the four core building blocks: asset creation, robots set-up and control, simulation handling, and complementary tools.
Following the logical steps also depicted in Fig.~\ref{fig:pipeline}, we proceed to describe the four core building blocks: asset creation, robots set-up and control, simulation handling, and complementary tools.

% We now proceed to describe our workflow in the next sections, focusing on an overview of the system, possible placement strategies, robot creation and control, simulation management with data generation, and its post-processing. A precise description of the generation procedure for the data that we used in our experiments is provided in the Supplementary Materials. In the supplementary, we also provide a comprehensive survey regarding simulation engines, indoor simulated environments and simulated animated and clothed humans.

\subsection*{Rigid and Non-Rigid Assets}
The environment or any other asset can be created beforehand in various applications before loading them in Isaac Sim. To this end, we leverage Blender, the Omniverse Connectors~\cite{omniverse-connectors}, and the Universal Scene Description (USD) format, adopted by Isaac Sim. The flexibility provided by the USD file format makes it possible to use different sources to generate assets for the simulation. Here, we focus mainly on indoor environments, rigid objects, and animated assets from different sources. We provide the reader with a review of the main sources of environments and animated humans in the Supplementary Materials.

For our experiments, our main source of environments is the 3D-Front~\cite{3d-front} dataset. This is one of the largest available, mesh-based, indoor environments datasets. We chose this also because sources based on scans of the environments and not mesh objects, like Matterport3D~\cite{Matterport3D}, cannot provide realistic lighting interaction. Additionally, in our experiments, we also use one Sketchfab indoor environment~\cite{interior-sketch} and two Unreal Engine freely available worlds~\cite{rural-australia,outdoorcity}. To prepare those, we first modify \textit{BlenderProc}~\cite{blenderproc} to be able to convert both the 3D-Front environments and other file formats (e.g. FBX, GLB) automatically. We modify the 3D-Front processor of \textit{BlenderProc} by fixing both texture generation and the semantic mapping procedure to partially correct wrong semantic classes, a known problem of this dataset~\cite{front3d-wrong}. We also compute a tight rectangle and the approximated enclosing non-convex polygon computation, a necessary step for our placement strategy. To that end, we first assume convex boundaries, find an enclosing convex polygon, and transform it to non-convex if the angle between two subsequent vertices differs from 90 degrees. While this is an approximation not able to capture all situations, such as small alcoves, for our use-case of indoor environments depicting houses it is enough. We then export, apart from the USD file to be loaded within Isaac Sim, the STL and the x3d conversions of the environment. The x3d format can be easily converted to an octomap~\cite{octomap}. The STL file, the polygon vertices, and the inscribing rectangle are used during the assets' placement procedure.

\usetikzlibrary{calc}
\usetikzlibrary{arrows}
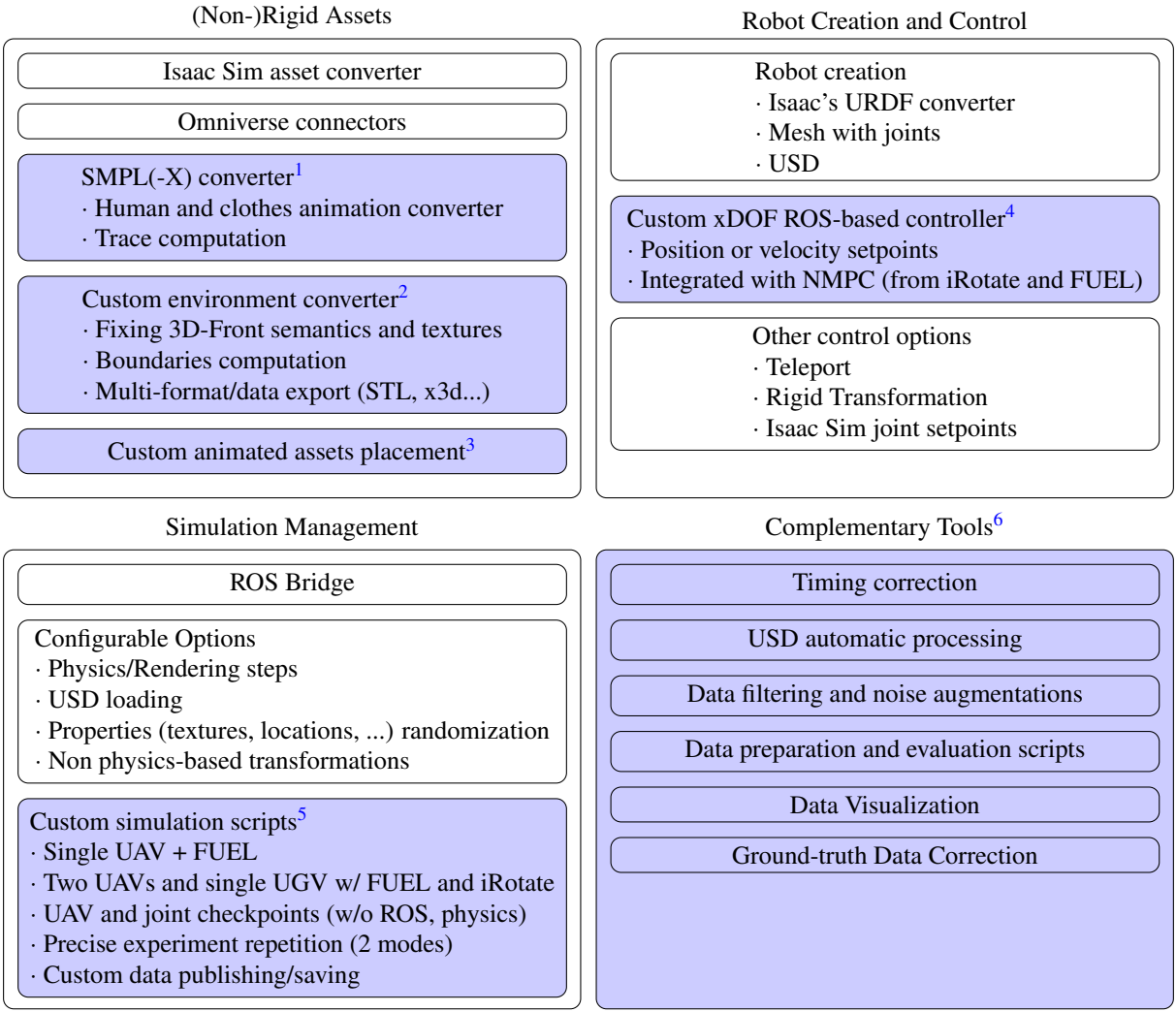
\begin{figure}[!ht]
    \centering
    \begin{tikzpicture}[module/.style={draw, thin, rounded corners,minimum width=0.47\textwidth},
    node distance=0.5cm and 0.5cm,
    on grid]
        % Define the width of each box based on the total text width divided by the number of lists
        
        \newcommand{\boxwidth}{0.48\textwidth} % slightly less than 1/4 of text width to allow spacing
        \newcommand{\boxheight}{6.3cm}
        
        % List 1: Rigid And Non-Rigid Assets
        \node[module,draw, text width=\boxwidth, align=left, anchor=north west, rounded corners, minimum height=\boxheight, label=above:(Non-)Rigid Assets] (list1) at (0, 0) {}; 
        % \node[module,align=left, anchor=north west] (A1) at ($(list1.north west) + (0.2, -0.2)$) {USD assets};        
        \node[module,align=left, anchor=north west] (A2) at ($(list1.north west) + (0.2, -0.2)$) {Isaac Sim asset converter};
        \node[module,align=left, anchor=north west] (A3) at ($(A2.south west) + (0, -0.2)$) {Omniverse connectors};
        \node[module,draw, fill=blue!20, align=left, anchor=north west] (A4) at ($(A3.south west) + (0, -0.2)$)  {SMPL(-X) converter\footnotemark \\$\cdot$ Human and clothes animation converter\\ $\cdot$ Trace computation};
        \node[module,draw, fill=blue!20, align=left, anchor=north west] (A5) at ($(A4.south west) + (0, -0.2)$) {Custom environment converter\footnotemark \\$\cdot$ Fixing 3D-Front semantics and textures\\ $\cdot$ Boundaries computation \\ $\cdot$ Multi-format/data export (STL, x3d...)};
        \node[module,draw, fill=blue!20, align=left, anchor=north west] (A6) at ($(A5.south west) + (0, -0.2)$) {Custom animated assets placement\footnotemark};
            
        % List 2: Robot Creation and Control
        \node[module,draw, text width=\boxwidth, align=left, anchor=north west, rounded corners, minimum height=\boxheight, label=above:Robot Creation and Control] (list2) at ($(list1.north east) + (0.2cm, 0)$) {};
        \node[module,align=left, anchor=north west] (B1) at ($(list2.north west) + (0.2, -0.2)$) {Robot creation\\
        $\cdot$ Isaac's URDF converter\\
        $\cdot$ Mesh with joints\\
        $\cdot$ USD};
        \node[module,draw, fill=blue!20, align=left, anchor=north west] (B2) at ($(B1.south west) + (0, -0.2)$){Custom xDOF ROS-based controller\footnotemark\\$\cdot$ Position or velocity setpoints\\$\cdot$ Integrated with NMPC (from iRotate and FUEL)};
        \node[module,draw, align=left, anchor=north west] (B3) at ($(B2.south west) + (0, -0.2)$){Other control options\\$\cdot$ Teleport\\$\cdot$ Rigid Transformation\\$\cdot$ Isaac Sim joint setpoints};

        % List 3: Simulation Management
        \node[module,draw, text width=\boxwidth, align=left, anchor=north west, rounded corners, minimum height=\boxheight, label=above:Simulation Management] (list3) at ($(list1.south west) - (0cm, .7cm)$) {};
        \node[module,align=left, anchor=north west] (C1) at ($(list3.north west) + (0.2, -0.2)$) {ROS Bridge};
        \node[module,align=left, anchor=north west] (C2) at ($(C1.south west) + (0, -0.2)$) {Configurable Options\\$\cdot$ Physics/Rendering steps\\$\cdot$ USD loading\\$\cdot$ Properties (textures, locations, ...) randomization\\$\cdot$ Non physics-based transformations};
        \node[module,draw, fill=blue!20, align=left, anchor=north west] (C3) at ($(C2.south west) + (0, -0.2)$){Custom simulation scripts\footnotemark\\$\cdot$ Single UAV + FUEL\\$\cdot$ Two UAVs and single UGV w/ FUEL and iRotate
        \\$\cdot$ UAV and joint checkpoints (w/o ROS, physics)\\$\cdot$ Precise experiment repetition (2 modes)\\$\cdot$ Custom data publishing/saving};

        % List 4: Complementary tools
        \node[module,draw, text width=\boxwidth, align=left,fill=blue!20, anchor=north west, rounded corners, minimum height=\boxheight, label={above:Complementary Tools\footnotemark}] (list4) at ($(list3.north east) + (0.2cm, 0)$) {};
        \node[module,draw, fill=blue!20, align=left, anchor=north west] (D3) at ($(list4.north west) + (0.2, -0.2)$){Timing correction};
        \node[module,draw, fill=blue!20, align=left, anchor=north west] (D4) at ($(D3.south west) + (0, -0.2)$){USD automatic processing};
        \node[module,draw, fill=blue!20, align=left, anchor=north west] (D5) at ($(D4.south west) + (0, -0.2)$){Data filtering and noise augmentations};
        \node[module,draw, fill=blue!20, align=left, anchor=north west] (D6) at ($(D5.south west) + (0, -0.2)$){Data preparation and evaluation scripts};
        \node[module,draw, fill=blue!20, align=left, anchor=north west] (D7) at ($(D6.south west) + (0, -0.2)$){Data Visualization};
        \node[module,draw, fill=blue!20, align=left, anchor=north west] (D8) at ($(D7.south west) + (0, -0.2)$){Ground-truth Data Correction};
        
    \end{tikzpicture}
    \caption{Recap of the main components of the GRADE framework. With a blue background, we highlight the software developed within the scope of this work and give reference to the specific repository in the footnotes.}
    \label{fig:pipeline}
\end{figure}

In our work, we use rigid assets from the Google Scanned Objects~\cite{googlescanned} (GSO) and the ShapeNet~\cite{shapenet} datasets. Then we employ pre-animated meshes, i.e. the freely available zebras~\cite{zebra+motions} from the Sketchfab marketplace, and SMPL~\cite{smpl} based animated humans. The humans come from the Cloth3D~\cite{cloth3d} and AMASS~\cite{amass} datasets. While Cloth3D also provides animated clothes for the given animation sequence, 
\footnotetext[1]{\url{https://github.com/eliabntt/animated_human_SMPL_to_USD}}
\footnotetext[2]{\url{https://github.com/eliabntt/Front3D_to_USD/}}
\footnotetext[3]{\url{https://github.com/eliabntt/moveit_based_collision_checker_and_placement/}} 
\footnotetext[4]{\url{https://github.com/eliabntt/custom_6dof_joint_controller}}
\footnotetext[5]{\url{https://github.com/eliabntt/grade-rr} and \url{https://github.com/eliabntt/ros_isaac_drone}}
\footnotetext[6]{\url{https://github.com/robot-perception-group/GRADE_tools}}
AMASS CMU sequences consist of only unclothed fittings. Again, all of these must be converted to the USD format. Isaac Sim provides a converter for the ShapeNet data, and more generally for common file formats like OBJ and FBX. We adapted it to be able to load GSO's objects at runtime, and animate them as flying objects through random, non-physics-enabled, transformations. We do so by defining predefined/random starting locations through the code at runtime and subsequent transformations in scale, position, and orientation. However, the provided converter often fails to correctly load and convert animated assets. While the Omniverse connector can be used for assets like the zebras, it failed to convert our animated humans. Thus, to solve this, we created a new program based on Blender, to automatically convert SMPL, potentially clothed, animated sequences to USDs while automatically saving complementary and necessary information. %Generally, the animations can be exported as deformed vertex mesh sequences and skeleton joint transformations, which can then be loaded as-is or modified during the simulation itself since Isaac Sim supports blendshapes. 
The program, initially based on the Cloth3D demo software, allows us to export both SMPL and SMPL-H animations as either deformed vertex mesh sequences or skeleton joint transformations, and clothes as deformed meshes. While doing so, we randomize the appearance of the assets by using Surreal's SMPL textures~\cite{surreal}. Those are freely available textures, but low-resolution, as can be observed in Fig.~\ref{fig:gendata}. We then load the animation sequence in Blender in batches and export those as USD files. Contextually, we generate an additional 3D trace of the animation and export it as an STL file to be used in our automatic placement procedure. Moreover, information like SMPL parameters, vertices locations, and length of the sequence are also saved at the same time. In our work, we used deforming meshes for the Cloth3D dataset, to solve some co-penetration problems between the clothing and the human assets, and skeletal transformations for AMASS.

Then, given an environment and an animated asset, we want to avoid any physical overlap between the two. To do so, various strategies can be employed. We can place the animated assets manually in the environment, as the zebras in the savanna scenario. However, given the specificity of the indoor set-up, we created a custom procedure for the (clothed) animated humans to avoid collisions. This is based on the 3D occupancy of the environment and the full 3D trace of each animation sequence we exported in the previous steps. Since those come as STL files, we can check for any co-penetrating point between the meshes and the world. We do so through a new MoveIt's~\cite{moveit} interface based on the Flexible Collision Library (FCL). This was more effective than using simple 2D occupancy projections. Indeed, the clothing animations and diversified actions we use (e.g. jumping, wide arms) can greatly increase the footprint without necessarily being in collisions with the environment. Different approaches can be easily introduced through simple Python scripting, thanks to the modularity of our approach. For all assets, we can apply further (selected) randomizations such as scale, orientation, texture, material properties, etc., at runtime.

\subsection*{Robot creation and control}
Robots can be created directly in Isaac Sim by adding joints to `objects' or through the URDF file format and the integrated connector. We first design and create the basic versions of our robots using the original Isaac Sim software. Using teleporting, a readily available option, and unrealistic (for a robot) rigid non-physics transformations, i.e. as a flying object, are not viable options for robot control. Clearly, these modalities can be used when the focus is mostly on the visual data collection, as we do during our experiment repetition and re-rendering.
% Indeed, control can be done \textit{with} or \textit{without} physics, as well as \textit{with} or \textit{without} ROS, increasing the flexibility of our framework which \textit{already} includes all these capabilities. This means that the simulation can be used both as an offline data-generation tool, e.g. by using teleporting and randomization, as well as an online platform to test robotic applications.
Instead, we seek to simulate custom UAVs and UGVs through our system while physics is enabled. However, Isaac Sim does not provide fluid-dynamic physics, necessary for UAVs, and, like Gazebo, it does not model frictionless perpendicular translation movements~\cite{irotate}, necessary for omnidirectional wheels. The recent PegasusSimulator~\cite{pegasus} addresses the control of the PX4 UAV by directly applying force to the drone mesh, without using any actual fluid dynamics simulation. Our approach differs as we control our robot through a PID-based joint-level control. We attach to the Firefly drone model from RotorS~\cite{rotors} and the three-wheeled omnidirectional robot from iRotate~\cite{irotate} one joint for each degree of freedom (six and three, respectively). Then, we use our custom controller by exploiting the ROS communication system and the joint definitions. We receive position or velocity setpoints from other software as input, e.g. a (N)MPC or an active SLAM framework, and convert that to joint commands through a PID formulation. Those are then received by the simulation management software and translated into robot movements. The ROS communication system is crucial in this case to seamlessly communicate with the Isaac Sim framework. Alternatively, instead of using a limited set of commands for a single specific platform like in BenchBot, we can give velocity and position setpoints to each joint separately while remaining agnostic to the underlying robot. We use the ROS-based control in the Active SLAM simulations that generated the data for our experiments, since iRotate and FUEL, both based on ROS, are used as our autonomous exploration frameworks. We then use the internal Isaac control for the experiment repetition tool. The robot's sensors can be defined beforehand, by directly adding them to the USD file of the robot, or by loading and modifying those dynamically at simulation time. Here, we follow the second approach, as it provides more flexibility. We can load heterogeneous robotic platforms and automatically differentiate ROS topics, sensor types, settings, and placement based on a single configuration file. Thus, a single robot model can be easily created at simulation time with different sensor suites without redefining the basic configuration each time, while publishing and listening to the desired topic names, simplifying control.

% \begin{figure*}[!th]
%     \centering
%     \includegraphics[width=\textwidth]{fig/pipeline.png}
%     \caption{A detailed scheme of GRADE, which shows how each tool can be adapted to everyone's need and the possibilities enabled by the tool developed by us.}
%     \label{fig:pipeline}
% \end{figure*}

\subsection*{Simulation management}
% \todo{I believe this is better than before but still a bit unsure about this, but first rewriting}
We manage the simulation through a single Python script with various submodules. The script communicates via ROS with the other software packages to complete the system. The main script, which uses Isaac Sim APIs alongside custom utilities, takes care mainly of i) starting and configuring the simulation environment, ii) loading, placing, and configuring the assets and robots, iii) several randomization procedures, iv) launch, if necessary, the complementary ROS nodes, v) managing the simulation steps (both physics and rendering) and data saving. 

We can save information or any additional data (e.g. skeleton's joint locations) directly to the hard drive from the simulation itself. While this was available through an extension, it was not directly usable through our system and was limited in its application. Indeed, the native ground-truth saving procedure did not allow, for example, custom data to be included or free access to the information during a robotics simulation but was limited to single rendering events. Our modifications allow us to have finer control, solve some failure cases (as segmentation IDs overflow), and access and save additional information such as the vertical field of view of the camera. Naturally, the data published by the sensors through ROS, can be saved also natively by using \textit{rosbags}. However, having two data-saving tools allows us to have, for example, two different camera resolutions: one for the online ROS-based SIL application, and one for training and evaluation of neural networks. We leverage this approach in our data generation procedure by using SIL to autonomously explore the environment using low-resolution images while saving full HD richly annotated ground-truth data. Moreover, having low-level access to the ground truth information means we can modify it before publishing to augment data with noise or custom drop rates. 

% Through the simulation software, we programmatically change environmental conditions, camera/rendering settings, light colors and intensity, materials' reflection parameters, assets' textures, and the time of day.

% Although these are characteristics that are inherited from the software tool that we opted to use, in this work we show how researchers can programmatically do all of this through different functions, automating the process, and allowing them to actively use those. This was possible thanks to our general approach which allowed us not only to randomize GRADE during our data generation and use it for different scenarios but also to expand Isaac Sim's capabilities with additional tools. 

% ROS messages can then be listened and/or published with different and customized rates. 
% The published messages can then be easily used by additional software online, e.g. from Active SLAM, or recorded through \textit{rosbags} for logging or offline processing. 

Through the script we can control various options like the number of dynamic assets, the initial location of the robot, or the size of the physics and rendering steps. This allowed us to have simulations in which the physics is either enabled or not, or ROS is either used or not, showing the flexibility obtainable by GRADE thanks to Isaac Sim APIs. In our work, we explicitly implemented several illustrative simulation control scripts, like Active SLAM-based exploration, ROS-free data collection in a savanna scenario with animated zebras, or experiment repetition. More details about these are given in the Supplementary Materials.
% Thus, customization of our codebase and integration of other software is easy, as our modular approach and the integration with ROS.

% i.e. simulations in which every step is interchangeable with custom implementations and different strategies.

% personalizing various settings and applying desired customizations, manually stepping the simulation with chosen step sizes , launching ROS nodes, including any other custom software or package, and fully supervising when and which information to publish and/or save.

\subsection*{Complementary tools}
We develop several additional tools to correct and use the data in downstream tasks. Indeed, the ground-truth information saved must then be processed to effectively test any robotics method in conditions as close as possible to the real world. Moreover, pre-processing the generated data is fundamental for using it with deep learning networks or during offline robotic testing. We thus add noise to the data by leveraging and expanding codes from RotorS~\cite{rotors} and the work from Zhang et al.~\cite{deblurDatasetIMU}. Specifically, in our work, we apply i) IMU noise and bias, ii) RGB motion blur and rolling shutter noise, iii) depth filtering and noise. We do this both to saved arrays and to \textit{rosbags} independently of Isaac Sim. Therefore, we can add noise to full-resolution images, segmentation masks, and bounding boxes, to be used with neural networks, and to \textit{rosbags}, used during our Dynamic V-SLAM evaluations. The noise augmentation tool can be easily expanded to different sensors (e.g. LRF, odometer), used only on specific topics, or with different noise models. While we apply that to the saved data, it can also be integrated directly into the simulation to publish noisy data alongside the corresponding ground truth. We also correct wrongly generated ground-truth data, such as 3D bounding boxes~\cite{bug-3d}, poses of some animated assets~\cite{bug-pose}, and \textit{rosbags} timings, all errors due to bugs in the Isaac simulator itself (details in the Supplementary Materials). Finally, we provide code and instructions towards a seamless data preparation for training of neural networks (e.g. by selecting object classes, or extracting keypoints), automatic processing to quickly benchmark the used SLAM frameworks, and other utilities such as data coloring or bulked USD file processing.   %\todo{This last sentence required a bit of work, but can't be really a contrib probably.}\todo{we also have a USD-processing script, but that is just text manipulation}
	
	% \balance
	\bibliographystyle{Science}
	\bibliography{biblio}
 
    \newpage
\section*{Acknowledgments}
The authors thank the International Max Planck Research School, Germany for Intelligent Systems (IMPRS-IS) for supporting Elia Bonetto.

\textbf{Author contributions:} Elia Bonetto is the primary author who formulated and implemented the simulation framework, any piece of code related to Isaac Sim, the data generation pipeline, the human assets and environments conversion code, the placement and robot's control software, and wrote the manuscript. He designed, carried out, and analyzed the experiments and their results. He implemented the scenarios and simulations presented (including the repeating experiment tool), and initiated the development of the data processing and augmentation code. Chenghao Xu contributed to the development and execution of the SLAM evaluations, participated in the development of the data post-processing tools, and prepared the GRADE datasets used for the training of the neural networks then carried out by Elia Bonetto. Elia Bonetto and Chenghao Xu contributed equally to the labeling of the TUM-RGBD data. Aamir Ahmad supervised the project and revised and edited the manuscript.

\textbf{Competing interests:} The authors declare that they have no competing interests.

\textbf{Data and materials availability:} All data needed to evaluate the conclusions in the paper are present in the paper or the Supplementary Materials. Other materials and the code can be found at \url{https://grade.is.tue.mpg.de/}.

\section*{Supplementary Materials}
\label{sec:soa}
In this supplementary material, we first present the current state-of-the-art on i) simulated and reconstructed indoor environments and ii) animated humans. Then, we proceed to describe i) a ROS- and physics- free simulation management example and ii) a (Multi-)Robot Active SLAM simulation example. Furthermore, we present in detail the two approaches we developed to perform experiment repetition. Then, we provide additional information on the data generation strategy and give an overview of the data we released with this work. We conclude by describing some issues of the current system which are independent of our framework.

\subsection*{Indoor Environments Survey}
\label{soa:envs}
There are two main ways in which we can represent indoor scenes within a simulation environment~\cite{survey3dcompletion}: scans of real-world environments or posed meshed objects.
Using scans of the real world (e.g. HM3D~\cite{hm3d}, Matterport3D~\cite{Matterport3D}, Gibson Env~\cite{gibsonenv}, SceneNN~\cite{scenenn-3dv16}, Replica~\cite{ replica19arxiv}, or Structured3D~\cite{Structured3D}) poses several issues. First, those are non-interactive environments in which all the objects are non-movable. Second, any \textbf{new} object or asset that is placed within the environment will not be lighted correctly and will not \textit{realistically} affect the scene with shadows or reflections. Finally, many of these present various artifacts, e.g. unrealistic-looking objects, holes in the reconstruction due to reflective surfaces or unmapped areas, and uneven surfaces~\cite{survey3dcompletion}.
Using worlds based on meshed 3D assets addresses these problems while also allowing randomization (e.g. on textures, and object placement) and, eventually, interaction. However, datasets based on those like ML-Hypersim~\cite{hypersim} and InteriorNet~\cite{interiornet}, usually rely on non-freely available elements and only release rendered images, making them unusable for our purposes. These factors limit their adoption, reproduction, and expansion. Furthermore, InteriorNet simulator has not been made available while in HyperSim the engine is not physics-based and its sequences relate only to very short trajectories (just 100 frames). 
ProcThor~\cite{procthor} is a recently developed framework to procedurally generate environments. However, it is limited in the quality of the assets and usable only within the Ai2THOR suite, which is focused on visual-AI rather than robotics and offers no ROS support. OpenRooms~\cite{openrooms} has not yet released any asset or CAD model publicly. 3D-FRONT~\cite{3d-front,3d-future} is a large publicly available dataset with meshed, professionally designed, and semantically annotated room layouts. This is, by far, the largest dataset available nowadays based on meshes that can be adopted. However, the annotations are not perfect and objects sometimes co-penetrate each other~\cite{hssd}. Finally, HSSD~\cite{hssd} is a synthetic matterport-like dataset of indoor scenes. While this is a viable alternative to 3D-FRONT, it does not provide light sources and is still much smaller. The five environments released with BEAR~\cite{bear} in five variations each instead are only slight modifications of worlds that are commercially available from Evermotion~\cite{bear}. By leveraging the integration of NVIDIA Isaac Sim as the main simulation engine within GRADE, and of the Omniverse Connectors, GRADE overcomes the limitations of existing scene representations and provides a platform for generating high-quality, customizable simulations for robotics research. Recently, a mesh-based generation strategy for indoor environments, Infinigen Indoor~\cite{infinigen-indoor}, has been released. It is already usable with Omniverse and, thus, with GRADE.

\subsection*{Dynamic Humans Survey}
% The majority of the dynamic content on a scene comes from humans and dynamic objects.
The most widely adopted method to describe a human body pose and shape is SMPL~\cite{smpl}. Using that, any motion, synthetic or real, can be considered a deformable and untextured mesh. Real-world SMPL fittings are obtainable only in controlled environments, e.g. through a VICON or MOCAP systems~\cite{amass}. These are limited in the number of subjects, clothing variety, and scenarios. Synthetic data is being used to solve these problems~\cite{airpose, peoplesanspeople}. However, such data does not include either the full camera's state, IMU readings, scene depth, LiDAR data, or offers the possibility to easily extend it after the experiment has been recorded (e.g. with additional cameras or sensors), thus is generally unusable for any robotics application. Indeed, synthetic datasets are usually developed by stitching people over image backgrounds~\cite{peoplesanspeople}, statically placing them in some limited environment~\cite{airpose} and often recorded with static monocular cameras that take single pictures~\cite{airpose}. Furthermore, many of those are generated without any clothing information~\cite{surreal}. Few datasets, like Cloth3D~\cite{cloth3d}, provide simulated clothed humans with SMPL fittings usable in other simulations. Commercial solutions like RenderPeople~\cite{renderpeople} or clo$|$3d~\cite{clo3d} exist and would be applicable. However, using such assets limits the possibility of reproduction and re-distributing data. Being able to include such assets in a simulated environment, as we do in GRADE, is of crucial importance to obtain simulations that are closer to reality.

\subsection*{ROS-free simulation in a Savanna environment}
\label{subsec:savana}
We can use GRADE to capture customized and controlled environments without necessarily using the physics engine. In this example, we collect data with UAVs in a savanna environment~\cite{rural-australia} with several simulated animated zebras. The environment is from Unreal Engine and converted to USD before loading it in Isaac Sim. The zebra and the set of animations are taken from a free asset in Sketchfab~\cite{zebra+motions}. Using Blender, we exported four animation sequences, i.e. walking, eating, trotting, and jumping. Then we created three different transitions set between those and manually placed them in the main environment (Fig.~\ref{img:scen_zebra}). The main body of the UAV is controlled without any SIL or ROS package. We experiment with both a scripted sequence of waypoints for each one of the six joints and make the drone a physics-less flying object. In any case, waypoints are then provided to the Isaac engine directly from the main simulation loop as position/orientation goals. The control is performed internally by the simulation engine. In the physics-enabled simulation, the dynamics are dictated by the mass of the drone and the joint characteristics, while in the physics-disabled by an interpolated trajectory between the goal locations. The schematic of this set-up is provided in Fig.~\ref{schema:zebras}.
\usetikzlibrary{arrows.meta}

\newpage

\begin{landscape}
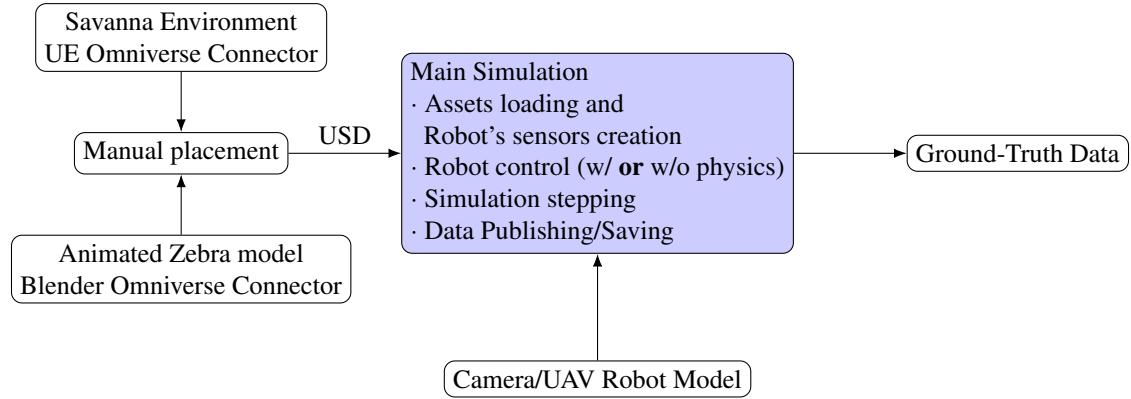
\begin{figure}[ht]
    % \centering
    % \resizebox{\columnwidth}{!}{
    \begin{tikzpicture}[module/.style={draw, thin, rounded corners},arrow/.style={-Latex, rounded corners},
    node distance = 0.0mm, auto,
    on grid]
        % \node[module, left=8mm and 8mm of A1.west,align=center] (B1){};
        \node[module](B3) at (5.5,0) {Camera/UAV Robot Model};

        % \node[module, fill=blue!20, align=left] (pen) at (5.5,2) {Physics enabled\\Joints' waypoints configuration};
        % \node[module, fill=blue!20, align=left] (pdis) at (5.5,4) {Physics disabled\\
                                                    % Flying object waypoints configuration
                                                    % };
        \node[module, fill=blue!20, align=left] (Core) at (5.5, 3) {Main Simulation\\
                                                    $\cdot$ Assets loading and\\~~Robot's sensors creation\\
                                                    $\cdot$ Robot control (w/ \textbf{or} w/o physics)\\
                                                    $\cdot$ Simulation stepping\\
                                                    $\cdot$ Data Publishing/Saving};

        \node[module, left=0mm and 15mm of Core.west] (B2){Manual placement};
        \node[module, above=8mm and 0mm of B2.north,align=center] (A2){Savanna Environment \\ UE Omniverse Connector};
        \node[module,align=center, below=8mm and 0mm of B2.south] (A1){Animated Zebra model \\ Blender Omniverse Connector};

        % \node[module, fill=blue!20, align=left] (ROS) at (5, -2.5) {Custom 6DOF Controller};
        % \node[module, align=left] (aslam) at (0, -2.5) {Active SLAM};

        \node[module, align=center, right=15mm of Core.east] (data) {Ground-Truth Data};
        % \coordinate (dataw) at ([xshift=-0.7cm]data.west);
        % \coordinate (datawdown) at (data |- Core);

        % \node[module, fill=blue!20, align=left] (Tools) at (10, -2) {Complementary tools\\
        % $\cdot$ Fix data\\$\cdot$ Add noise\\$\cdot$ Train/SLAM testing};
        
        \draw[arrow] (B2.east) -- (Core) node[midway,above]{USD};
        \draw[arrow] (B3.north) -- (Core);
        \draw[arrow] (Core.east) -- (data.west);

        \draw[arrow] (A2.south) -- (B2);
        \draw[arrow] (A1.north) -- (B2);
        % \draw[arrow, -] (ROS.east) -- (datawdown);
        % \draw[arrow] (datawdown) -- (data.north);
        % \draw[arrow] (data.south) -- (Tools.north);
        % \draw[arrow,<->] (ROS) -- (Core);
        % \draw[arrow,<->] (aslam) -- (Core);
        % \draw[arrow] (aslam) -- (ROS);
    \end{tikzpicture}%}
    \caption{A schematic of the ROS-free system used in the Savanna simulation. In blue, we highlight our customizations.}
    \label{schema:zebras}
\end{figure}

\begin{figure}[ht]
    % \centering
    \resizebox{\columnwidth}{!}{
    \begin{tikzpicture}[module/.style={draw, thin, rounded corners},arrow/.style={-Latex,rounded corners},
    node distance = 0.0mm, auto,
    on grid]

        \node[module, align=center] (A1) at (0, 0) {Isaac Sim converter};
        \node[module, fill=blue!20,above left=2mm and 0 of A1.north east, align=left] (B1){Custom converter\\
                                                    $\cdot$ Correct textures\\
                                                    $\cdot$ Correct semantic classes\\
                                                    $\cdot$ Extract Borders\\
                                                    $\cdot$ Export USD and STL};
        \node[module, align=center, left=0mm and 10mm of B1.west] (envs) {3D-FRONT, ...};
        \draw[arrow](envs.east) -- (B1);
        \node[module, align=center] (oooo) at (A1-|envs){ShapeNet, GSO, ...};
        
        \node[module, above left=2mm and 0 of B1.north east] (B2){U(A/G)V Robot Model(s)};
        \node[module, fill=blue!20,below left=2mm and 0 of A1.south east, align=left] (C1) {SMPL(-X) converter\\$\cdot$ Build animation\\
                                                    $\cdot$ Export USD\\
                                                    $\cdot$ Build animation trace\\
                                                    $\cdot$ Export trace's STL};
        \node[module, align=center, left=0mm and 10mm of C1.west] (anims) {Cloth3D, AMASS, ...};
        \draw[arrow](anims.east) -- (C1);

        \node[module, fill=blue!20, align=left] (Core) at (7, 0.7) {Main Simulation\\
                                                    $\cdot$ Physics enabled\\
                                                    $\cdot$ Asset loading and placement\\
                                                    $\cdot$ Randomize textures\\
                                                    $\cdot$ Robot's Sensors creation\\
                                                    $\cdot$ Manual simulation stepping\\
                                                    $\cdot$ Data Publishing/Saving};

        \coordinate (merge_arrows) at ([xshift=-1.5cm]Core.west);
        
        \node[module, align=center,right=1.5cm of Core.east] (data) {Ground-truth Data\\(npy, \textit{rosbags})};
        
        \node[module, fill=blue!20, align=left, right=0mm and 15mm of data.east] (Tools) {Complementary tools\\
        $\cdot$ Fix data\\$\cdot$ Add noise\\$\cdot$ Train/SLAM testing};
        
        \node[module, align=left] (aslam) at (Core |- B2) {Active SLAM}; %, above=30mm and 0mm of Core.north
        \node[module, fill=blue!20, align=left, right=10mm of aslam.east] (ROS) {Custom 6DOF Controller};
        \coordinate (tmp_below_ros) at ([yshift=1cm]Core-|ROS);
        \draw[arrow,-] (A1.east) -- (merge_arrows|-A1);
        \draw[arrow,-] (B2.east) -- (merge_arrows|-B2);
        \draw[arrow,-] (B1.east) -- (merge_arrows|-B1);
        \draw[arrow,-] (C1.east) -- (merge_arrows|-C1);
        \draw[arrow,-] (merge_arrows|-B2) -- (merge_arrows|-C1);
        \draw[arrow] (merge_arrows) -- (Core) node[above,midway]{USDs};
        \draw[arrow] (oooo) -- (A1);
        \draw[arrow] (Core) -- (data);
        \draw[arrow] (tmp_below_ros) -- (Core.east|-tmp_below_ros);
        \draw[arrow] (data.east) -- (Tools.west);
        \draw[arrow] (tmp_below_ros) -- (ROS) node[midway,right]{ROS};
        \draw[arrow] (aslam) -- (Core) node[midway,right]{ROS};
        \draw[arrow] (Core) -- (aslam);
        \draw[arrow] (aslam) -- (ROS);
    \end{tikzpicture}}
    \caption{A schematic of the main dataset generation pipeline. In blue, we highlight our customizations.}
    \label{fig:mainschema}
\end{figure}
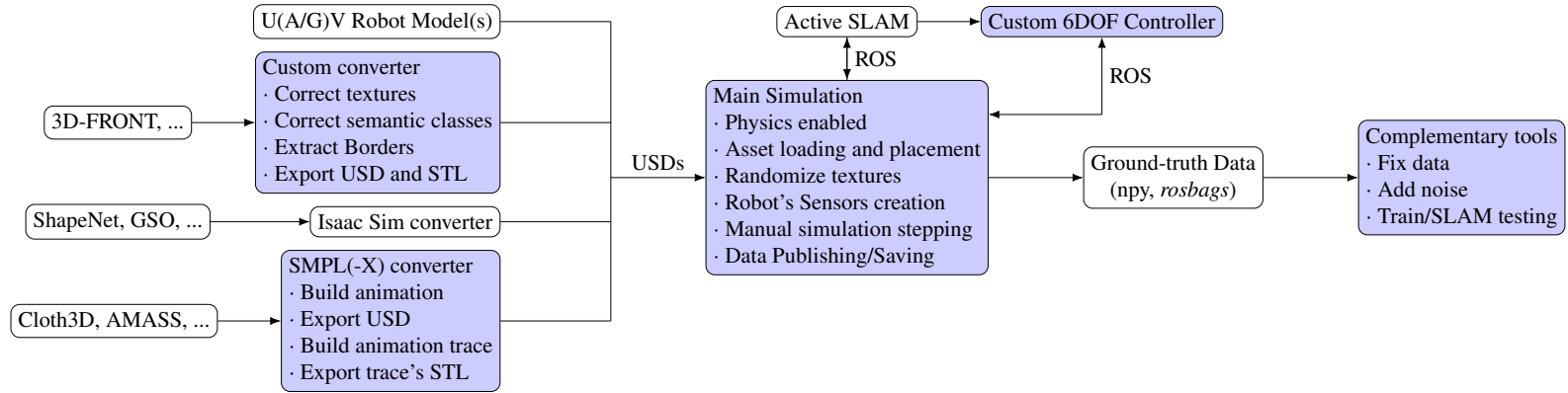
\end{landscape}

\begin{small}
\begin{algorithm}
\SetAlgoLined
  \KwData{\\
  WORLD: folder path of the static environment\\
  ASSETS: list of paths of the assets to be placed\\
  STRICT: boolean, if the random location should be inside the enclosing rectangle or the tight boundaries}
  \KwResult{List of triplets: assets, locations, and orientations}
  \tcp{Preparation}
  environment = load WORLD STL;\\
  boundaries = load WORLD boundaries;\\
  rectangle = load WORLD enclosing rectangle;\\
  placed = []; \\
  isGood = False;\\
  \tcp{Main Loop}
  \ForEach{elements of ASSETS}{
  asset = load element STL;\\
  \For{i = 1 \KwTo 10}{
  \eIf{STRICT}{
  position = random position within boundaries;\\
}{
  position = random position within rectangle;\\
}
orientation = random orientation;\\
  isGood = CheckForCollision(environment, asset, position, orientation, limit=200);\\
  \If{isGood}{
  \tcp{The current placement is collision free}
  break;}
  }
  \If{isGood}{
    \tcp{Update the return vector}
    placed.append((element, position, orientation));\\
    \tcp{Update the environment so that this asset is taken into account in the future}
    environment.update((asset, position, orientation));\\
  }
  }
  \Return placed;

\caption{Pseudocode of the placement procedure.}
\label{alg:placement}
\end{algorithm}
\end{small}

\subsection*{(Multi-)Robot Active SLAM}
\label{sec:gen}

A scheme of the workflow is depicted in Fig~\ref{fig:mainschema}. We use the custom environment converter to prepare the world and extract its STL and boundaries. The robot model comes in the USD format and already has attached one joint for each degree of freedom, necessary to use our controller. With the SMPL animation converter, we prepare the animated assets and extract the STL representing their trajectory. Note that this STL is a static representation of the trajectory that the animated human and its clothes will take in time. For those, we use the placement strategy briefly described in the main paper. The pseudocode for that is provided in Alg.~\ref{alg:placement}. The placement is performed using our FCL-based MoveIt plug-in. From Isaac Sim we publish the list of animated assets that we want to place. Then, the plug-in first loads the STL of the environment and iteratively tries ten times to place each animation one after the other. The initial location is randomized at each iteration, based on the environments' boundaries or the enclosing rectangle. If the asset cannot be placed due to detected collisions, it is removed from the simulation since it will not have a valid location. In doing this, we consider a threshold value below which a collision is not considered, avoiding rejecting intersections with small objects like plant leaves, and set this threshold to 200 contact points. Note how this can work irrespective of the nature of the asset, which can be animated or not. 

We then adapt the FUEL~\cite{zhou2021fuel} ROS-based Active SLAM framework to autonomously explore the environments and interface it with Isaac Sim\footnote{\url{https://github.com/eliabntt/ros_isaac_drone}}. FUEL's exploration goals are then supplied to an NMPC~\cite{kamelmpc2016} to predict a realistic trajectory for the robot. The final predicted state is then provided to our custom controller, which supplies commands to the joints limited in speed and range as in Tab.~\ref{tab:limits}. FUEL uses online RGB-D and odometry data from the simulation to actively provide goals to the controller. The initial location of the robot is randomized using the same placement procedure used for the animated assets to avoid collisions. Furthermore, its initial orientation is pre-optimized to avoid pointing toward empty spaces, e.g. a window. In this scenario, we can also simultaneously manage multiple different robots. The challenge is that we have to dynamically load specific sensor suites and set-up different ROS topics for each one of the robots created. An example is provided in Figures~\ref{img:scen_gr1},~\ref{img:scen_gr2}, and~\ref{img:scen_gr3}. In those, two robots are separate instances of the same UAV, each running a different instance of FUEL, and one is a three-wheeled ground omnidirectional robot. This is interfaced, through our custom controller, with the iRotate framework.

\begin{table}[!h]
\centering
\begin{tabular}{l|c|c|c}
 & x, y, z & roll, pitch & yaw \\
\hline
Position & Env. limits & $[-25,25]$ or $[0,0]$ degrees & $[0,360]$ degrees \\
Speed & $0.5$ m/s & $40$ deg/s & $30$ deg/s
\end{tabular}
\caption{Joint limits}
\label{tab:limits}
\end{table}

\subsection*{Experiment repetition}
\begin{landscape}
\begin{figure*}[h]
    \centering
    \captionsetup[subfigure]{justification=centering}
    {\includegraphics[width=0.198\columnwidth]{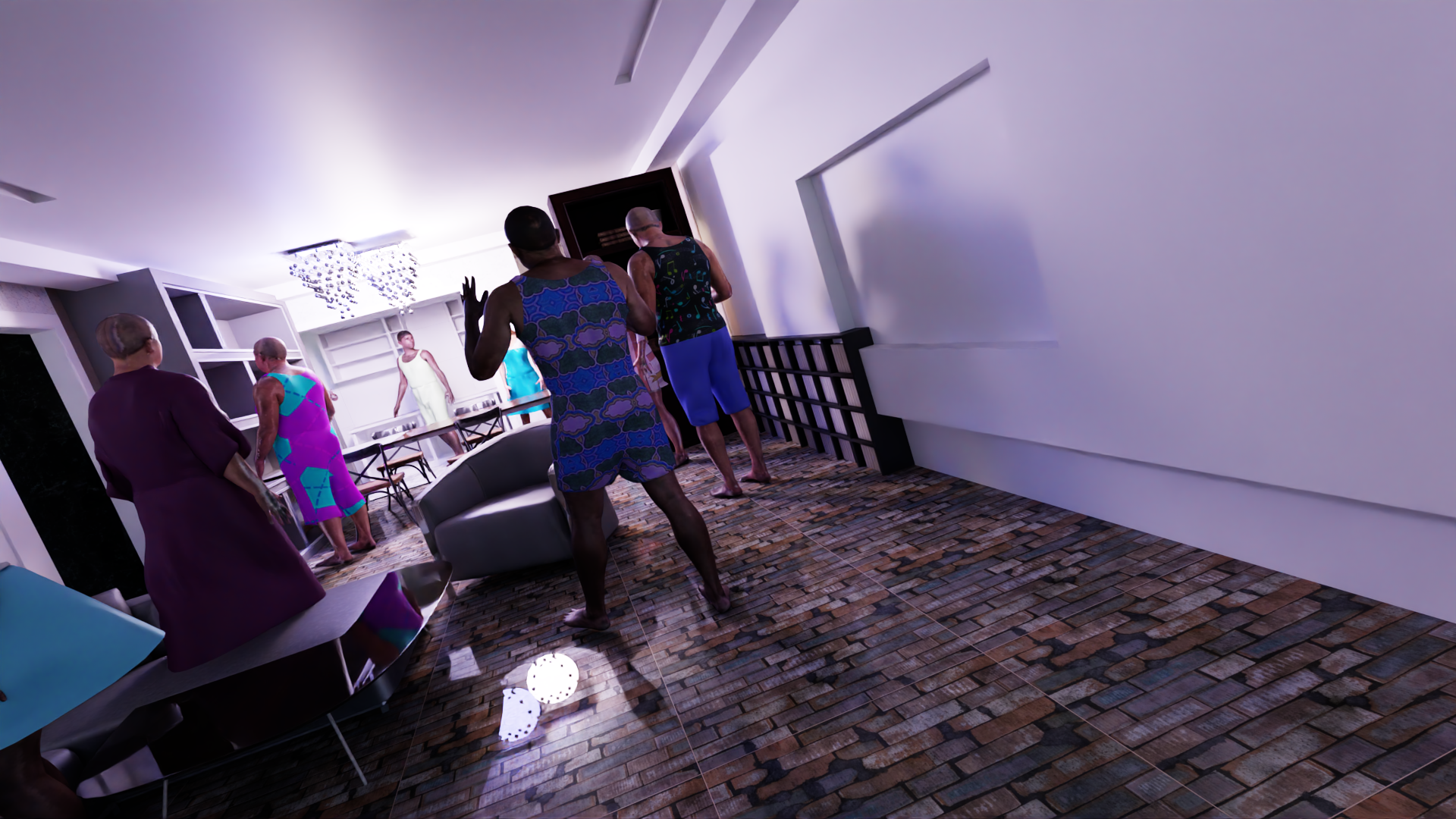}}\hfill
    {\includegraphics[width=0.198\columnwidth]{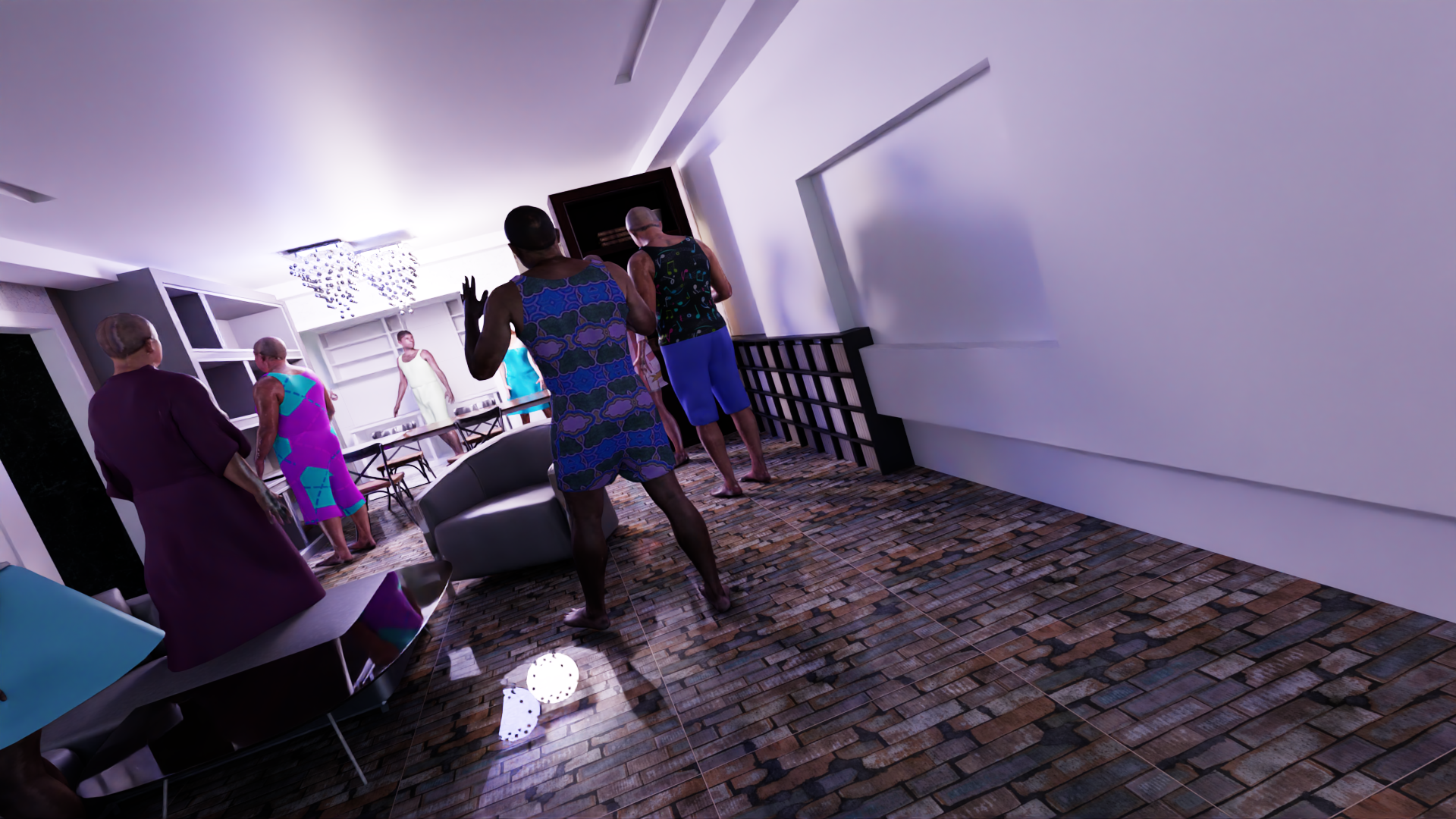}}\hfill
    {\includegraphics[width=0.198\columnwidth]{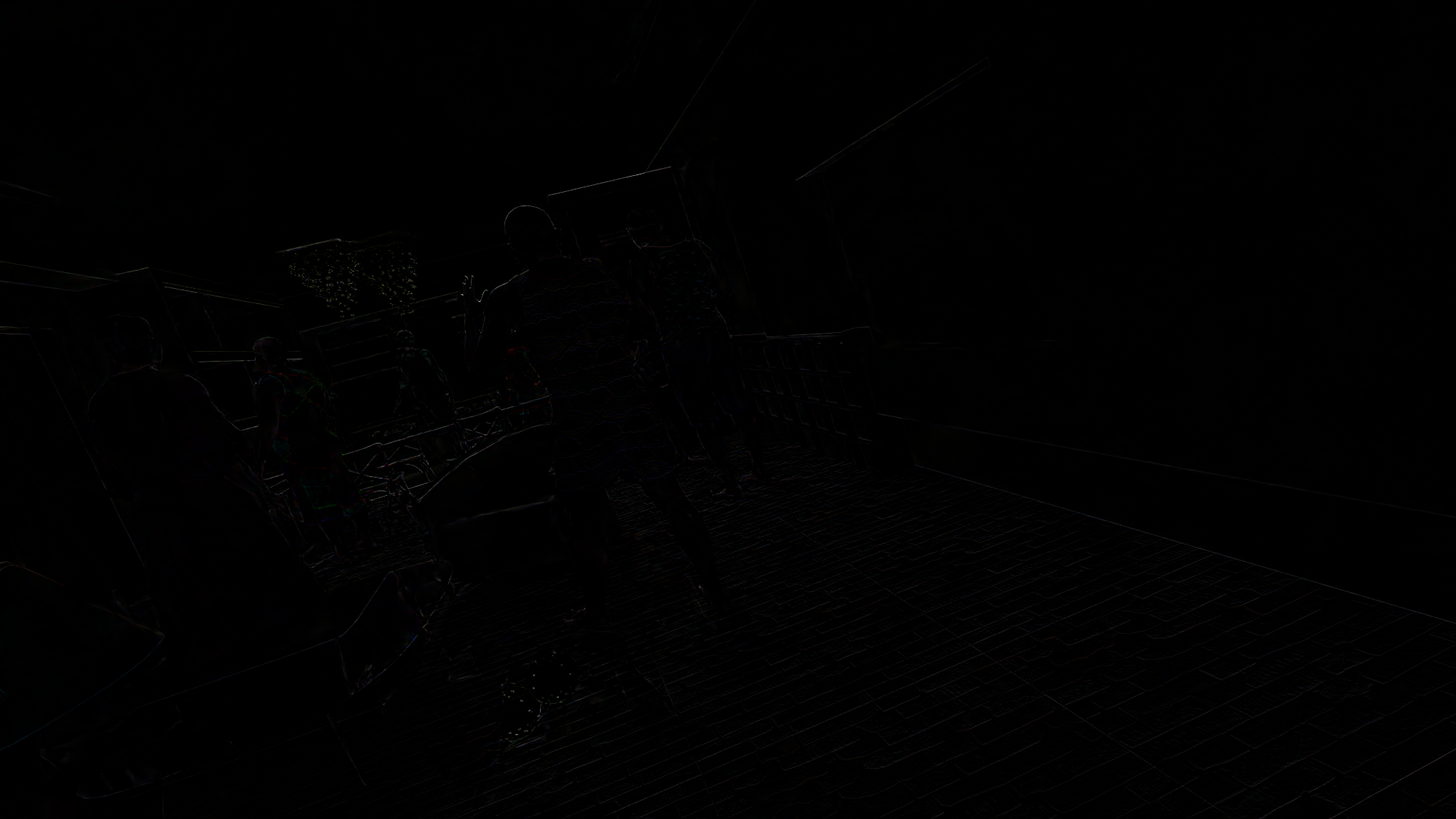}}\hfill
    {\includegraphics[width=0.198\columnwidth]{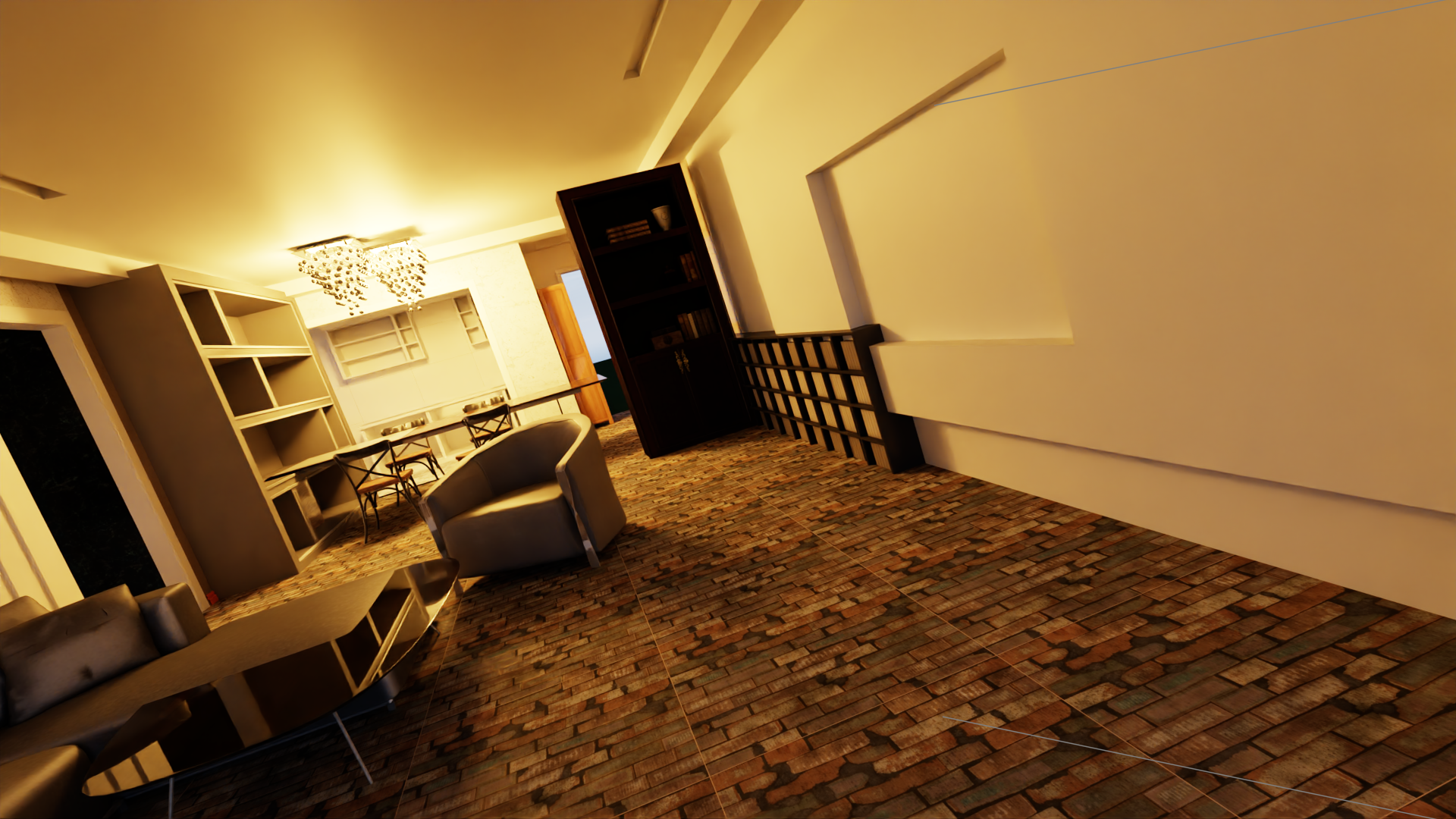}}\hfill
    {\includegraphics[width=0.198\columnwidth]{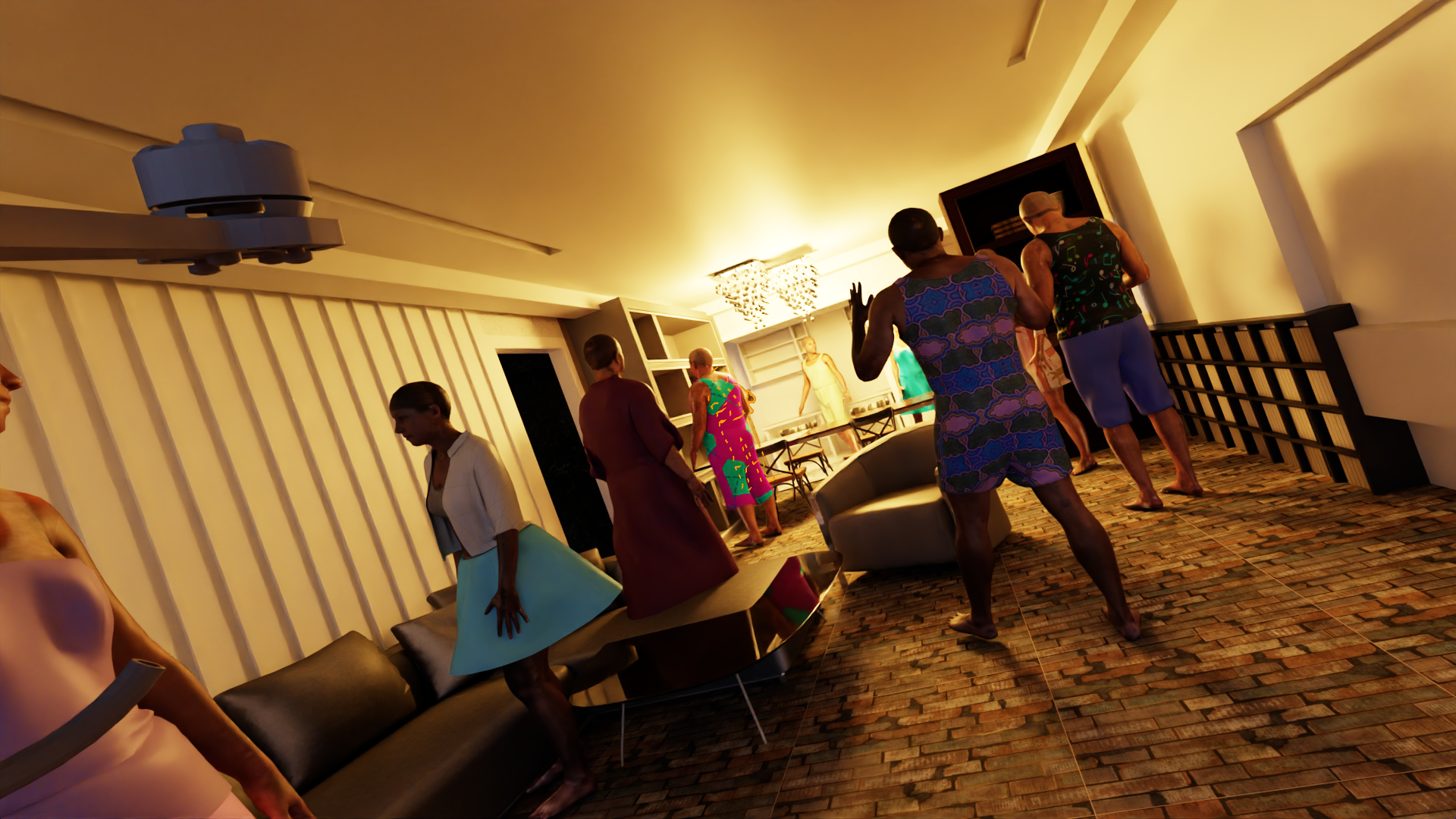}}\hfill
    
    \vspace{3.5pt}

    \subcaptionbox{Original frame}{\includegraphics[width=0.198\columnwidth]{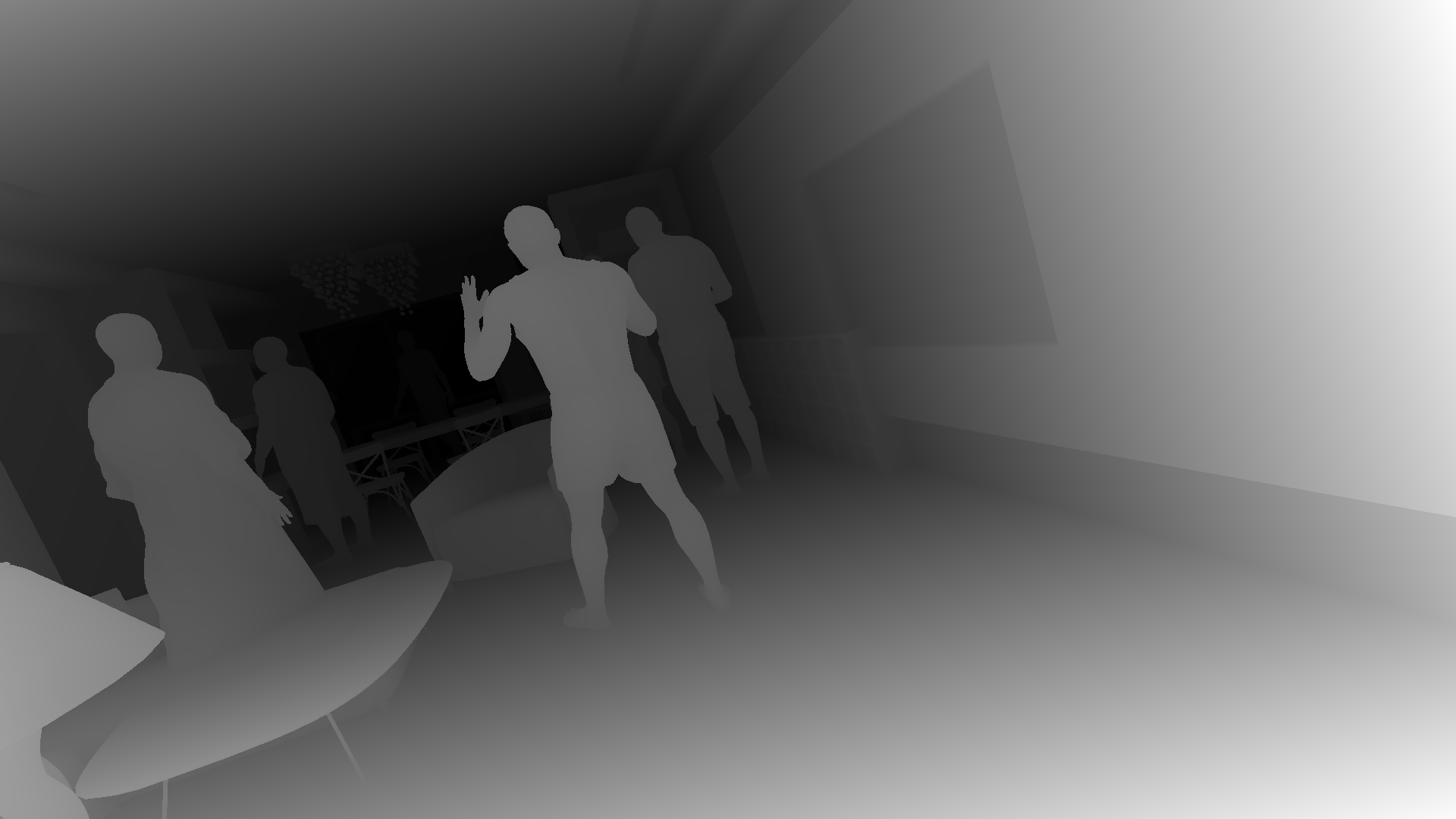}}\hfill
    \subcaptionbox{Re-generated frame}{\includegraphics[width=0.198\columnwidth]{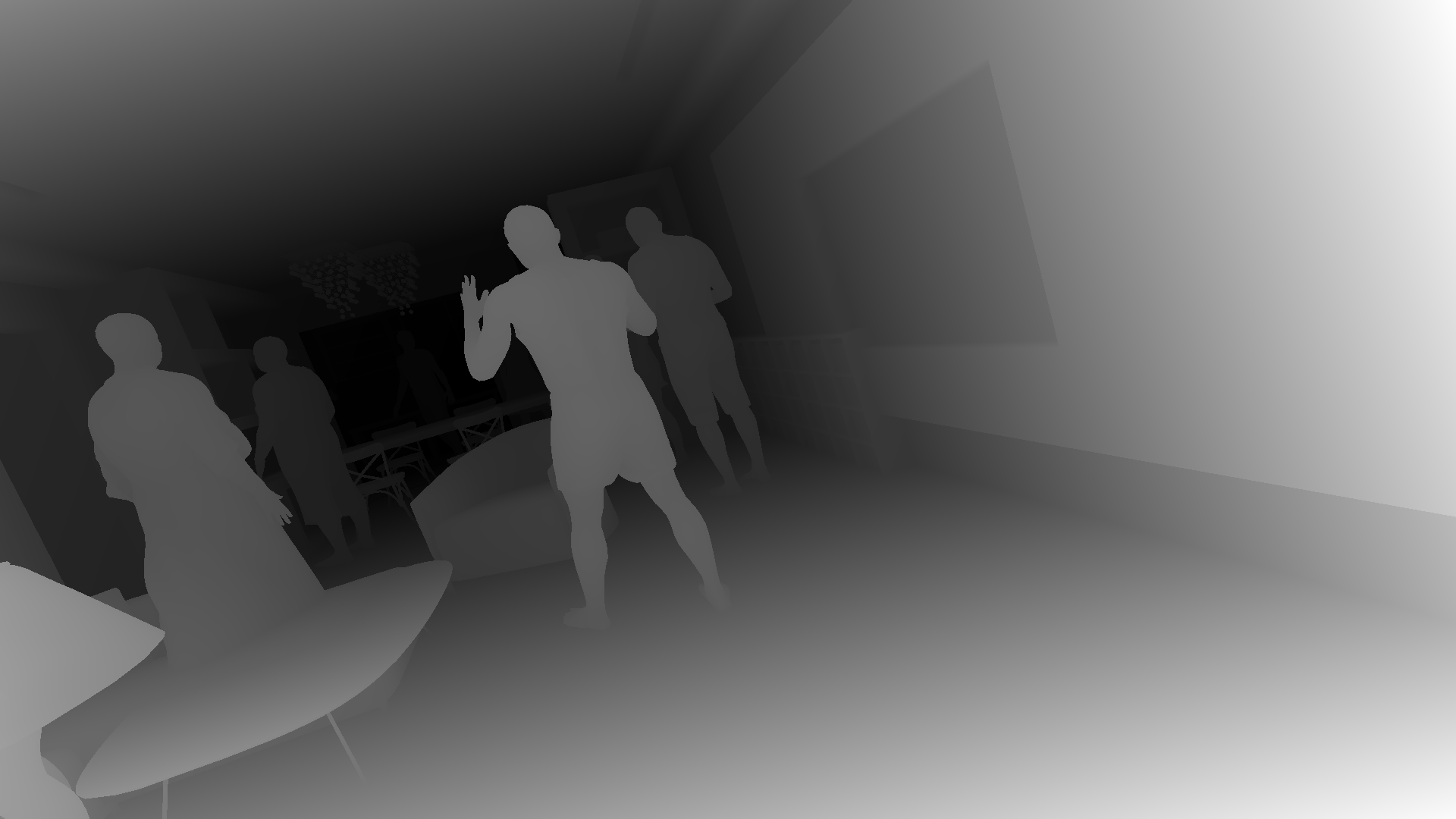}}\hfill
    \subcaptionbox{Difference between \\\textit{(a)} and \textit{(b)}}{\includegraphics[width=0.198\columnwidth]{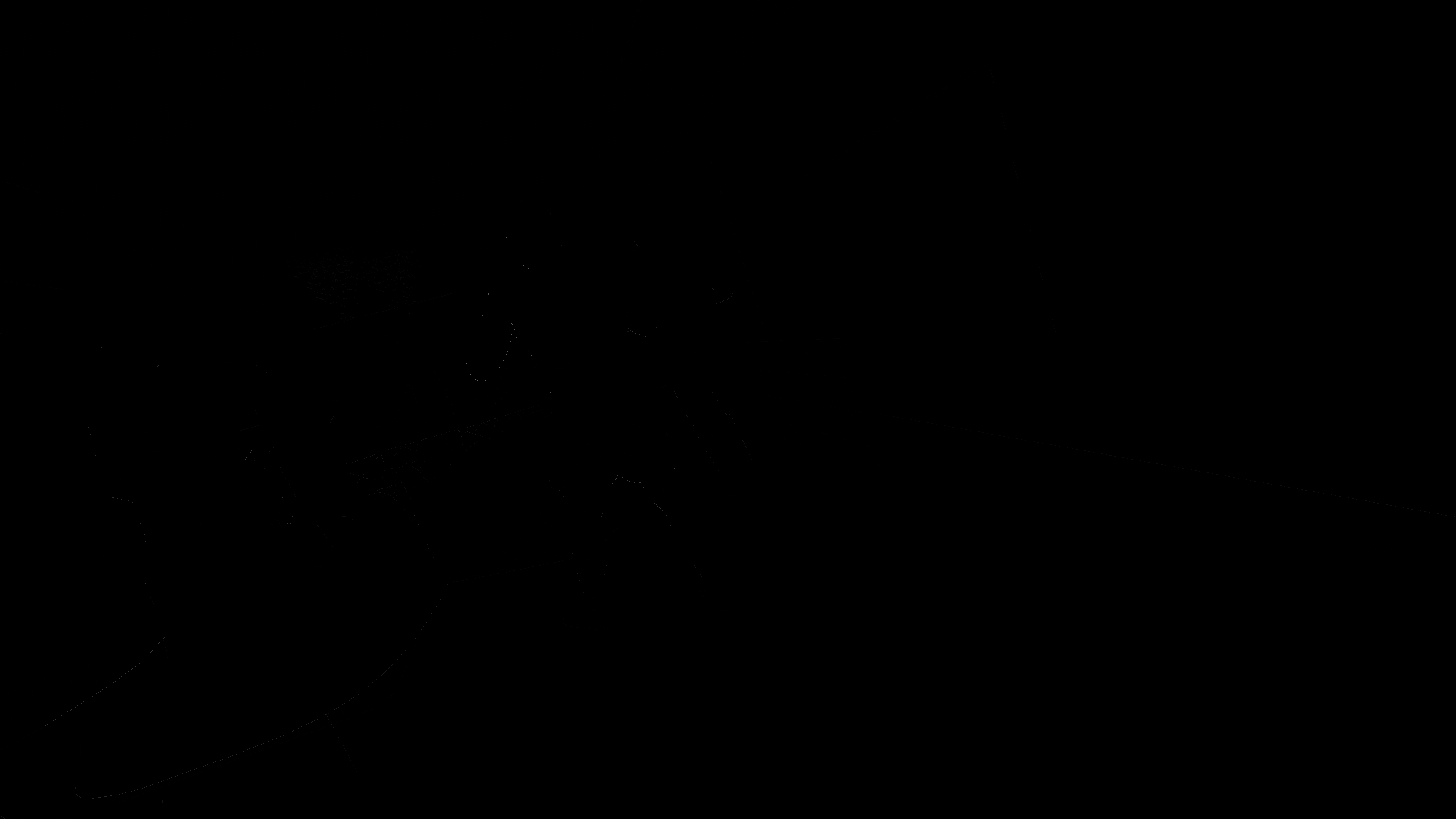}}\hfill
    \subcaptionbox{\textit{(a)} w/o dynamic objects and after a light color change}{\includegraphics[width=0.198\columnwidth]{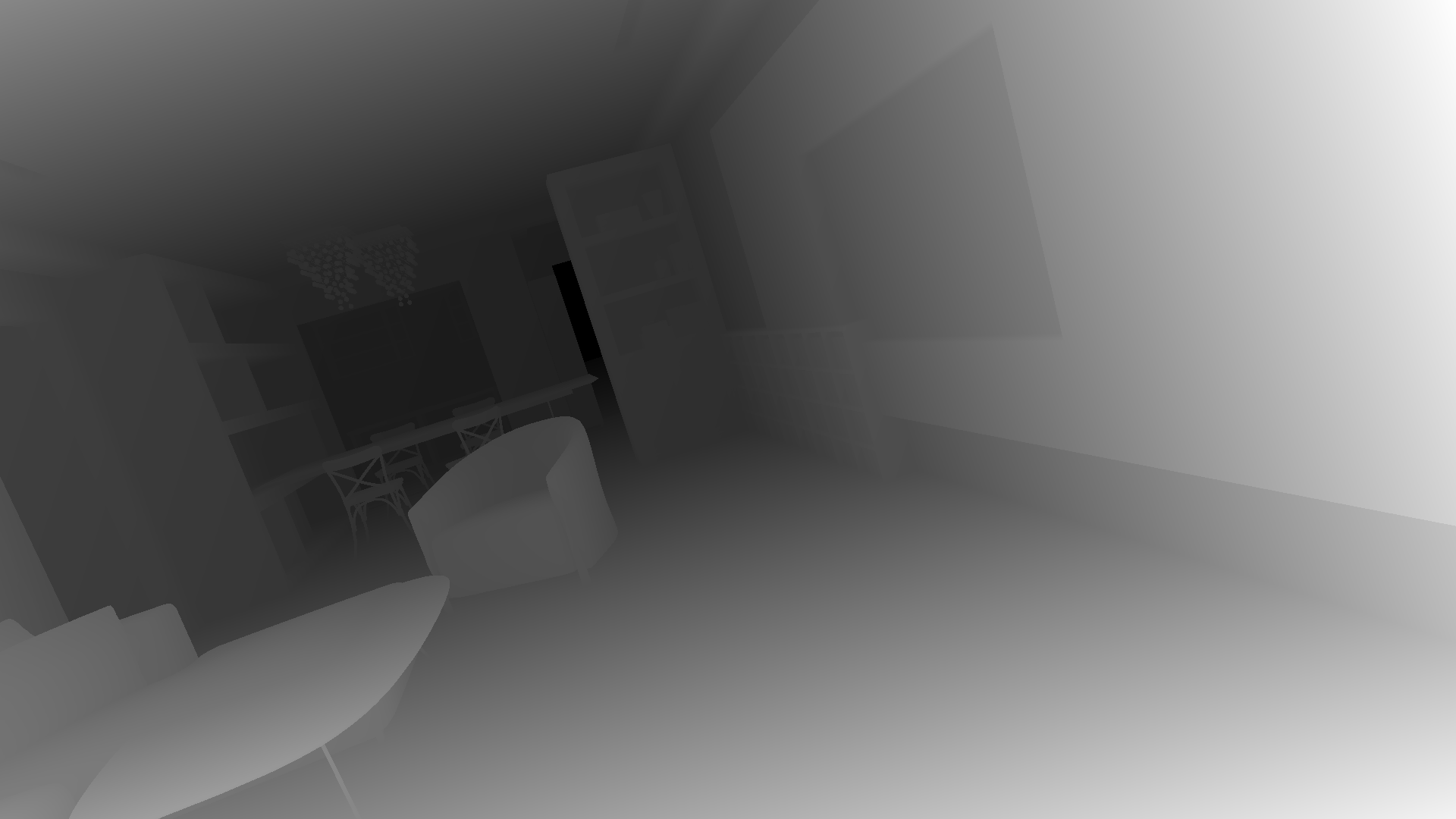}}\hfill
    \subcaptionbox{\textit{(a)} observed from a novel viewpoint}{\includegraphics[width=0.198\columnwidth]{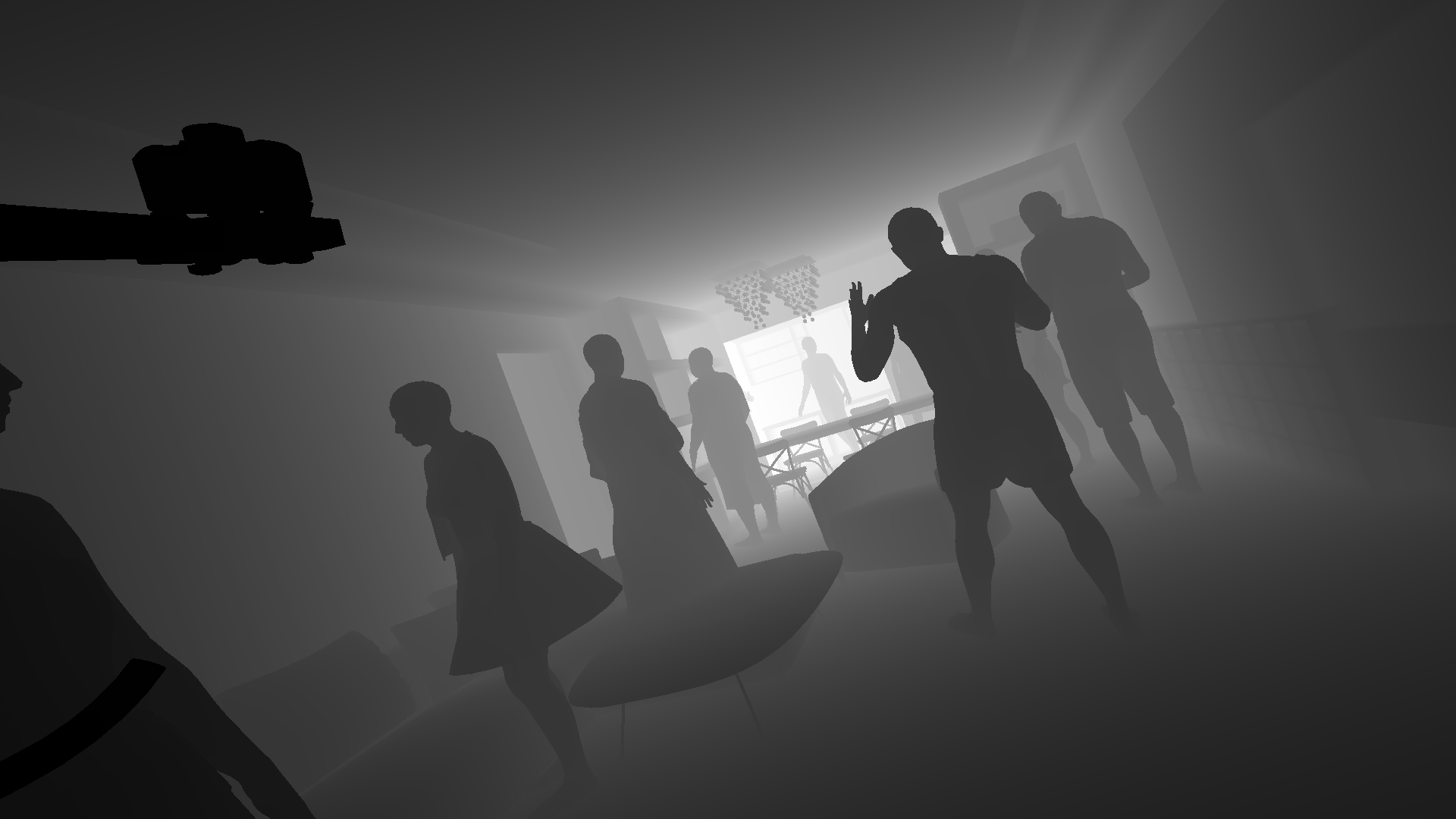}}\hfill

    \caption{Examples of RGB and depths generated using the experiment repetition functionality of GRADE. In the first column, we show the original RGB and depth frames, while the second column depicts the regenerated sensor readings captured at the \textit{same} location. Then, in column \textit{c}, we show the difference between the two corresponding frames in the RGB and depth domains. In this case, the structural similarity of the RGB image is 97.65\% while the MAE on the depth is 0.00066 meters. In columns \textit{d} and \textit{e} we show the same scene with a different lighting condition, first using the same viewpoint but hiding all the dynamic objects, and then changing the camera viewpoint to a different orientation. In column \textit{e} the depth coloring is inverted for better visibility.}
    \label{fig:repeating}
\end{figure*}
\end{landscape}

\begin{figure}[!ht]
    % \centering
    \resizebox{\columnwidth}{!}{
    \begin{tikzpicture}[module/.style={draw, thin, rounded corners},arrow/.style={-Latex, rounded corners},
    node distance = 0.0mm, auto,
    on grid]

        \node[module, align=left] (A1) at (0, 0) {Previous Experiment data\\$\cdot$ Joint positions/velocities\\
        $\cdot$ Simulation configuration\\$\cdot$ Assets locations and animations};

        \node[module, align=left, fill=blue!20, right=0mm and 8mm of A1.east] (data) {Main Code\\
        $\cdot$ Load USDs and settings\\
        $\cdot$ Modify the simulation\\
        ~~according to configuration file\\
        $\cdot$ Main Loop w/ control of physics, rendering\\
        $\cdot$ Save data
        };
        \draw[arrow] (A1.east) -- (data.west);
        
        \node[module, align=left, fill=blue!20, above right=-2mm and 5mm of data.north east] (teleport) {Control Teleport\\
        $\cdot$ Disable physics on main drone OR\\
        ~~Disable physics whole simulation\\
        $\cdot$ Read position, orientation\\
        $\cdot$ Interpolate and Teleport the robot
        };

        \node[module, align=left, fill=blue!20, below=22mm and 0mm of teleport.south] (joints) {Control Joints\\
        $\cdot$ Keep physics active\\
        $\cdot$ Read next position, orientation\\
        $\cdot$ Read next velocity\\
        $\cdot$ Set target speed, position, orientation\\
        $\cdot$ Eventually, teleport at fixed steps
        };

        \coordinate (choice) at (teleport.south |- data);
        \draw[arrow] (choice) -- (data.east);% node[midway, above]{Either one or the other}
        \draw[arrow] (choice) -- (teleport);
        \draw[arrow] (choice) -- (joints);

    \end{tikzpicture}}
    \caption{A schematic of the experiment repetition pipeline. In blue, we highlight our customizations.}
    \label{fig:repeatingschema}
\end{figure}
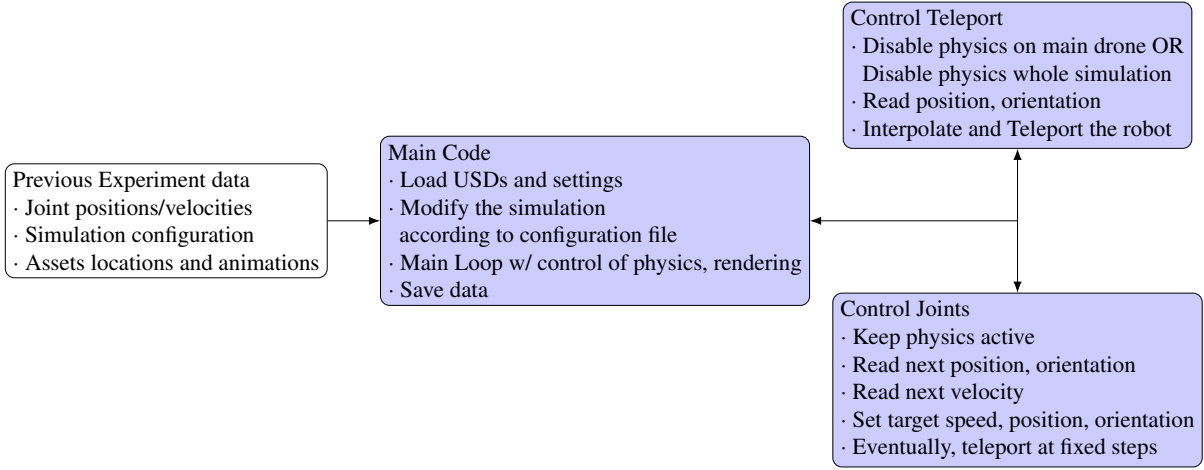

To repeat experiments, we provide two possible solutions. The requirement is that the pose of the robot and the initial simulation conditions must be logged in some form. The first way we implement for repeating experiments is by teleporting the robot to the exact location and re-rendering the scene. Eventually, we can interpolate the missing poses when necessary, e.g. if a newly added sensor has a different rate than the one with which the pose was saved. We also introduce another approach using the physics engine and setting directly the previously logged joint velocities and target positions at every time step. While this could generate some minor deviation due to the unknown acceleration between two timesteps, combining the two strategies can easily mitigate this effect. Then, while the timeline itself controls the animated assets, and the main robot is moved by our tool, the rest of the simulation is still managed by the physics engine. Thus, one can now easily change the experiment and expand the data by adding new physically enabled robots, changing the visual conditions, or adding new sensors. Notably, some information, such as the skeletal configuration at a given timestep, can be extracted without rendering. An example of the re-rendered images during an experiment repetition run is provided in Fig.~\ref{fig:repeating}, while the schematic is in Fig.~\ref{fig:repeatingschema}.

\subsection*{Details on the GRADE Data Generation}
Using the notions introduced in the main paper and in the above (Multi-)Robot Active SLAM scenario, we generate a dataset of richly annotated data collected autonomously in dynamic environments. For the main datasets, i.e. A-GRADE, S-GRADE, and the tested dynamic sequences, we only use worlds from the 3D-Front~\cite{3d-front} dataset, the city environment from the Unreal Engine marketplace~\cite{outdoorcity}, and an indoor environment from SketchFab~\cite{interior-sketch}. The 3D-Front environments are randomized with \textit{ambientCG}~\cite{ambientcg} textures. The flying objects, loaded from GSO and ShapeNet, increase variability in the simulation and introduce more challenging scenarios. For simplicity, we do \textit{not} restrict those objects to \textit{not} collide with other parts of the environment. By design, flying objects are loaded and randomly rigidly animated, without considering any possible collision. While more precise strategies can be implemented, this allows for more variability and challenging scenes. 

The main process is as follows. First, we load the environment, center it w.r.t. the origin, and generate the 2D occupancy map. In this, we also define the size of the physics step, i.e. how much the clock will advance during every loop of the simulation, and rendering parameters, e.g. auto-exposure or path/ray-tracing settings. In our experiments, the physics step is set to 1/240-th of a second. Second, we randomize lights' colors, intensities, and surface roughness (i.e. reflection capabilities). Third, we load the robot, move it to the starting location, and attach its specific set of sensors (including the cameras), ROS publishers, and link it to the motion controller through the correct ROS topics. The robot is controlled with either full 6 DOF capabilities or with a stabilized flight (without roll and pitch) based on the chosen configuration. Then, we import a random number of people and randomly place them within the environment. The number of flying objects loaded varies based on the experiment settings. Once loaded, we randomize their movement patterns based on the environment boundaries. Overall, considering that all simulation have animated humans and accounting for the degrees of freedom of the robot and the presence of flying objects, we obtained six different scenarios. In this work, each experiment sequence can indeed be identified by the usage of N (no) F (few) L (lot) letters, indicating the number of obstacles, and the presence of the letter H, which points to a stabilized flight of the drone. 

After loading and setting up the simulation experiment, we bootstrap the first $1$ second for every experiment to randomize the initial conditions of the robot. When the bootstrap sequence ends, we publish a single message to signal the start of the experiment and record data for one minute. In the main simulation loop we advance the physics one step at a time, manually control the animation timeline, publish the ROS information at the desired rates (based on the physics step), and write data to the disk. The drones are controlled online through the respective active SLAM framework and autonomously explore the environment. A full description of the data that we save with the corresponding FPS is in Tab.~\ref{tab:pubfreq}. The scene is rendered with path tracing and auto exposure. 

\begin{table}[!ht]
\centering
\setlength\tabcolsep{3pt}
\resizebox{0.7\columnwidth}{!}{
\begin{tabular}{l|c|c|c|c|c|c|c|c|c}
\textbf{Sensor} & 
Clock &
\begin{tabular}[c]{@{}c@{}}IMU\\ body/camera\end{tabular} & 
TF  & 
\begin{tabular}[c]{@{}c@{}}Joint\\ state\end{tabular} &
\begin{tabular}[c]{@{}c@{}}Camera\\ pose\end{tabular} & 
Odometry & 
RGB &
Depth &
\begin{tabular}[c]{@{}c@{}}Starting\\ Experiment\end{tabular}   \\ \hline
\textbf{FPS} & 240 & 240 & 120 & 120 & 60 & 60 & 30 & 30 & once 
\end{tabular}}
\caption{Publish frequency for each one of the sensors recorded}
\label{tab:pubfreq}
\end{table}

\begin{table}[!ht]
\centering
\resizebox{0.7\columnwidth}{!}{
\begin{tabular}{l|c|c|c|c|c|c}
& 
Humans & 
GSO & 
ShapeNet &
Horizontal & 
Free & Sequences \\
\hline
N & 7-40 & 0 & 0 & & \tick & 63 \\
\hline
NH & 7-40 & 0 & 0 & \tick & & 77 \\
\hline
F &  7-40 & 5 & 5 & & \tick & 44 \\
\hline
FH & 7-40 & 5 & 5 & \tick & & 63 \\
\hline
 L & 7-40 & 10 & 10 & & \tick & 33 \\
\hline
LH & 7-40 & 10 & 10 & \tick & & 62 \\
\end{tabular}}
\caption{Our data generation summary. N/F/L indicates that at the beginning of the experiment, No/Few/Lot of obstacles are loaded (see the Google Scanned Objects, GSO, and ShapeNet columns). With H we indicate sequences in which the robot is forced to be horizontal, i.e. without roll or pitch movements. The number of humans loaded \textit{before} placement is randomly chosen for each generation between 7 and 40. After placement, this number can be further reduced due to missing space.}
\label{tab:combinations}
\end{table}

\begin{table*}[ht]
\centering
\resizebox{\textwidth}{!}{
 \begin{tabular}{llr|cc|cc|cc|cc|cc|cc|cc}
& & {} & \multicolumn{2}{c|}{DynaVINS --- VO} & \multicolumn{2}{c|}{DynaVINS --- VIO} & \multicolumn{2}{c|}{StaticFusion} & \multicolumn{2}{c|}{TartanVO} & \multicolumn{2}{c|}{DynaSLAM} & \multicolumn{2}{c|}{ORB-SLAMv2} & \multicolumn{2}{c}{RTABMap} \\
& {} & {} & ATE & TR & ATE & TR & ATE & TR & ATE & TR & ATE & TR & ATE & TR & ATE & TR \\ \hline
 
\multirow{16}{*}{\begin{sideways}Ground-truth data\end{sideways}} & \multirow{2}{*}{D} & 
  mean & 1.429 & 0.713 & 0.195 & 0.989 & 22.374 & 0.999 & 1.264 & 1.000 & 0.046 & 0.805 & 0.251 & 0.995 & 0.547 & 0.995 \\
& &  std & 0.275 & 0.265 & 0.009 & 0.000 & 0.000 & 0.000 & 0.000 & 0.000 & 0.009 & 0.070 & 0.053 & 0.000 & 0.000 & 0.000 \\ \cline{2-17}
& \multirow{2}{*}{DH} & mean & 1.692 & 0.655 & 7.926 & 0.993 & 14.938 & 0.999 & 1.259 & 1.000 & 0.014 & 0.087 & 0.006 & 0.176 & 0.097 & 0.646 \\
& &  std & 0.568 & 0.307 & 0.290 & 0.000 & 0.000 & 0.000 & 0.000 & 0.000 & 0.010 & 0.026 & 0.002 & 0.005 & 0.000 & 0.000 \\ \cline{2-17}
& \multirow{2}{*}{F} & mean & 1.733 & 0.789 & 2.262 & 0.981 & 2.781 & 0.999 & 4.132 & 1.000 & 0.708 & 0.420 & 0.408 & 0.487 & 0.091 & 0.219 \\
& &  std & 0.370 & 0.302 & 0.750 & 0.000 & 0.000 & 0.000 & 0.000 & 0.000 & 0.311 & 0.127 & 0.245 & 0.151 & 0.000 & 0.000 \\ \cline{2-17}
& \multirow{2}{*}{FH} & mean & 0.424 & 0.991 & 0.073 & 0.989 & 0.059 & 0.999 & 0.551 & 1.000 & 0.274 & 0.987 & 0.223 & 1.000 & 0.200 & 1.000 \\
& &  std & 0.290 & 0.003 & 0.003 & 0.000 & 0.000 & 0.000 & 0.000 & 0.000 & 0.072 & 0.038 & 0.083 & 0.000 & 0.000 & 0.000 \\ \cline{2-17}
& \multirow{2}{*}{WO} & mean & 1.315 & 0.772 & 0.585 & 0.963 & 1.418 & 0.999 & 2.473 & 1.000 & 0.094 & 0.079 & 0.134 & 0.197 & 0.210 & 0.197 \\
& &  std & 0.361 & 0.259 & 0.068 & 0.003 & 0.000 & 0.000 & 0.000 & 0.000 & 0.013 & 0.000 & 0.007 & 0.000 & 0.000 & 0.000 \\ \cline{2-17}
& \multirow{2}{*}{WOH} & mean & 1.683 & 0.900 & 0.234 & 0.982 & 4.926 & 0.999 & 2.361 & 1.000 & 0.012 & 0.538 & 0.011 & 0.538 & 0.087 & 0.569 \\
& &  std & 0.452 & 0.092 & 0.027 & 0.001 & 0.000 & 0.000 & 0.000 & 0.000 & 0.001 & 0.000 & 0.001 & 0.000 & 0.000 & 0.000 \\ \cline{2-17}
& \multirow{2}{*}{S} & mean & 0.044 & 0.993 & 0.226 & 0.991 & 22.282 & 0.999 & 1.205 & 1.000 & 0.015 & 1.000 & 0.013 & 1.000 & 0.081 & 1.000 \\
& &  std & 0.002 & 0.000 & 0.007 & 0.000 & 0.000 & 0.000 & 0.000 & 0.000 & 0.002 & 0.000 & 0.002 & 0.000 & 0.000 & 0.000 \\ \cline{2-17}
& \multirow{2}{*}{SH} & mean & 0.031 & 0.993 & 0.116 & 0.991 & 2.721 & 0.999 & 2.395 & 1.000 & 0.013 & 1.000 & 0.012 & 1.000 & 0.087 & 1.000 \\
 & &  std & 0.006 & 0.000 & 0.005 & 0.000 & 0.000 & 0.000 & 0.000 & 0.000 & 0.003 & 0.000 & 0.002 & 0.000 & 0.000 & 0.000 \\

\midrule
\multirow{16}{*}{\begin{sideways}Noisy data\end{sideways}} & \multirow{2}{*}{D} & 
  mean & 1.438 & 0.826 & 0.601 & 0.989 & 10.123 & 0.999 & 1.350 & 1.000 & 0.057 & 0.686 & 0.683 & 0.835 & 0.529 & 0.902 \\
& &  std & 0.298 & 0.225 & 0.059 & 0.000 & 0.000 & 0.000 & 0.000 & 0.000 & 0.009 & 0.160 & 0.067 & 0.075 & 0.000 & 0.000 \\ \cline{2-17}
& \multirow{2}{*}{DH} & mean & 1.779 & 0.846 & 2.672 & 0.993 & 23.295 & 0.999 & 1.214 & 1.000 & 0.003 & 0.051 & 0.005 & 0.054 & 0.056 & 0.183 \\
 & &  std & 0.407 & 0.118 & 0.646 & 0.000 & 0.000 & 0.000 & 0.000 & 0.000 & 0.001 & 0.000 & 0.000 & 0.000 & 0.000 & 0.000 \\ \cline{2-17}
& \multirow{2}{*}{F} & mean & 1.910 & 0.888 & 5.411 & 0.973 & 2.661 & 0.999 & 4.192 & 1.000 & 0.220 & 0.274 & 0.128 & 0.268 & 0.123 & 0.218 \\
& &  std & 0.405 & 0.199 & 3.811 & 0.002 & 0.000 & 0.000 & 0.000 & 0.000 & 0.034 & 0.025 & 0.023 & 0.030 & 0.000 & 0.000 \\ \cline{2-17}
& \multirow{2}{*}{FH} & mean & 0.411 & 0.993 & 0.150 & 0.988 & 2.379 & 0.999 & 0.568 & 1.000 & 0.273 & 1.000 & 0.279 & 1.000 & 0.158 & 1.000 \\
 & &  std & 0.149 & 0.000 & 0.017 & 0.001 & 0.000 & 0.000 & 0.000 & 0.000 & 0.051 & 0.000 & 0.082 & 0.000 & 0.000 & 0.000 \\ \cline{2-17}
& \multirow{2}{*}{WO} & mean & 1.180 & 0.890 & 1.117 & 0.960 & 1.724 & 0.999 & 2.399 & 1.000 & 0.113 & 0.080 & 0.135 & 0.197 & 0.265 & 0.197 \\
 & &  std & 0.409 & 0.202 & 0.120 & 0.003 & 0.000 & 0.000 & 0.000 & 0.000 & 0.014 & 0.002 & 0.007 & 0.000 & 0.000 & 0.000 \\ \cline{2-17}
& \multirow{2}{*}{WOH} & mean & 1.351 & 0.869 & 0.534 & 0.980 & 2.691 & 0.999 & 2.389 & 1.000 & 0.037 & 0.536 & 0.019 & 0.536 & 0.081 & 0.539 \\
 & &  std & 0.230 & 0.131 & 0.085 & 0.001 & 0.000 & 0.000 & 0.000 & 0.000 & 0.006 & 0.000 & 0.002 & 0.000 & 0.000 & 0.000 \\ \cline{2-17}
& \multirow{2}{*}{S} & mean & 0.052 & 0.993 & 0.202 & 0.991 & 21.558 & 0.999 & 1.259 & 1.000 & 0.026 & 1.000 & 0.027 & 1.000 & 0.088 & 1.000 \\
 & &  std & 0.003 & 0.000 & 0.011 & 0.000 & 0.000 & 0.000 & 0.000 & 0.000 & 0.002 & 0.000 & 0.002 & 0.000 & 0.000 & 0.000 \\ \cline{2-17}
& \multirow{2}{*}{SH} & mean & 0.057 & 0.993 & 0.496 & 0.991 & 5.602 & 0.999 & 2.537 & 1.000 & 0.018 & 1.000 & 0.018 & 1.000 & 0.082 & 1.000 \\
 & &  std & 0.006 & 0.000 & 0.298 & 0.000 & 0.000 & 0.000 & 0.000 & 0.000 & 0.002 & 0.000 & 0.002 & 0.000 & 0.000 & 0.000 \\

\end{tabular}
}

 \caption{ATE RMSE [m] and tracking rate of the tested sequences in both their ground-truth and noisy versions. Each experiment is 60 seconds long and the depth is limited to 5 m. \bigskip}
 \label{tab:slam-5m}
\end{table*}

The rendering speed greatly depends on the number of lights, reflections, assets in the scene, and cameras. We use two cameras with the same horizontal and vertical FOV, one low resolution ($640 \times 480$) and one high resolution ($1920 \times 1080$). Across different architectures, rendering each \textbf{couple} of views took an average of 12 seconds, including the time necessary to get the remaining ground truth information such as the instance segmentation information. However, by tuning the simulation parameters, one can improve that to multiple images per second up to processing times faster than 15 FPS. In this, the physics simulation step causes a noticeable delay~\cite{nvidiaSlowSpeeding} whenever high-frequency messages related to physics need to be published (e.g. the IMU) due to its tiding with the USD files and the necessity to constantly read and write information to them~\cite{nvidiaSlowSpeeding}. The low-resolution RGB and depth are published with ROS and saved with the \textit{rosbag} tool. The high-resolution camera is used to save ground truth data with \textit{numpy} arrays such as 2D and 3D bounding boxes, instance segmentation masks, camera pose, as well as RGB and depth. All of these can also be published as ROS topics. The animations are reversed based on the average movement duration to increase variability and avoid too many static figures. Note that, due to flying objects and the unpredictability of moving subjects, sometimes the drone goes `through' dynamic assets. Although a way to avoid this would be to pre-sample valid trajectories, we see this as an opportunity to develop more reliable systems. Indeed, these events cause situations, such as losing track of the features due to completely black images or sudden changes between frames, which usually remain untested since not present in currently available datasets. However, those are possible scenarios that can occur if one considers real hardware or communication links that might fail or degrade or real-world environments in which animals or anything else could cause occlusions of the camera lens even for a brief moment. Nonetheless, we are not aware of any data currently available that poses such a challenge and, through our repeating experiment tool, one can solve this issue by simply re-generating the occluded frames. 

We augment the images of S-GRADE using a random rolling shutter noise model ($\mu=0.015$, $\sigma=0.006$) and a fixed exposure time of $0.01$ following~\cite{deblurDatasetIMU}. With A-GRADE we use instead a random exposure time between $0$ and $0.1$ seconds for each sequence and update the segmentation masks and bounding boxes to account for the additional blur using the same code we utilize for blurring the images. This is unnecessary for S-GRADE noisy images as the blurring amount is much lower due to the shorter (and fixed) exposure time. 

\subsection*{Summary of Released Data and Code}
\label{subsec:reldata}
We release 342 sequences of 1800 frames each which, at 30 fps, correspond to 342 minutes of videos, i.e. 615K frames. Those are summarized in Tab.~\ref{tab:combinations}. For each one of the 342 experiments we release depth data, instance segmentation (including clothing segmentation), 2D tight and loose bounding boxes~\cite{bounding-boxes}, 3D bounding boxes, and the corresponding camera information and poses. Additionally, we release processed human data in the form of 3D vertex locations and skeletal information. This information is saved with \textit{numpy} arrays. For each sequence, we also release the recorded \textit{rosbags} with IMU readings, TF tree, joint states, low-resolution RGBD images, and robot's state. IMU readings, camera pose, and robot pose are already provided pre-extracted for convenience. For each experiment, we provide the initial configuration of each asset, the state of the random number generator used, the USD file of the simulation, and other accompanying information necessary to replay the experiment. Other data such as normal vectors, optical flow, can be automatically generated by using the experiment repetition tool. 

All the USD files of the animated sequences and environments can be freely downloaded upon acceptance of the necessary licenses, or generated from scratch when necessary (ShapeNet objects). The source code is fully open source. For convenience, we also release the pre-processed data to test Dynamic V-SLAM methods and the data relative to each experimented approach reported in this work in Tables~\ref{tab:slam-3.5m-static-gt},~\ref{tab:slam-3.5m}, and~\ref{tab:both-nets} of the main paper and~\ref{tab:slam-5m} of the supplementary material. Statistics of the average camera (absolute) speed, average (absolute) acceleration, number of dynamic frames, and average portion of the dynamic frame belonging to dynamic people of the S, SH, D, DH, WO, WOH, F, and FH sequences are depicted in Tab.~\ref{tab:stats}. The average speed and acceleration are obtained from the ground truth values of the odometry recorded at 60 Hz. DH shows accelerations and velocities on roll and pitch due to a small collision between the robot and the environment. Finally, we release the checkpoints of the trained networks alongside the data used to train them and the TUM-RGBD labeled data. A summary of these datasets, their training and validation splits, and the notation used in this work for these can be seen in Tab.~\ref{tab:str_dataset_humans}.

\begin{table}[!ht]
   \centering
    \resizebox{0.7\columnwidth}{!}{
\begin{tabular}{l|c|c|c|c}

 & \begin{tabular}{c} Average speed \\ $[$x, y, z$]$\\ $[$roll, pitch, yaw$]$  \\ \end{tabular} & \begin{tabular}{c} Average acceleration \\ $[$x, y, z$]$ \\ $[$roll, pitch, yaw$]$  \\  \end{tabular} & \begin{tabular}{c} Dynamic \\ frames \\ \end{tabular} & \begin{tabular}{c} Covered\\ ratio \\ \end{tabular}\\ \hline

FH &   
\begin{tabular}{c} $[0.084, 0.091, 0.014]$\\$[0.000, 0.000, 0.444]$ \end{tabular} &    \begin{tabular}{c} $[0.425, 0.551, 0.123]$\\$[0.000,  0.000,  0.698]$ \end{tabular} &           1800 & 34.06 \\ \hline
F & \begin{tabular}{c} $[0.244, 0.239, 0.139]$\\$[0.189, 0.184, 0.436]$ \end{tabular} &  \begin{tabular}{c} $[0.891, 0.882, 0.740]$\\$[1.625, 1.564, 1.242]$ \end{tabular} &           1247 &  11.55 \\ \hline
DH & \begin{tabular}{c} $[0.255, 0.185, 0.046]$\\$[0.029, 0.014, 0.476]$ \end{tabular} & \begin{tabular}{c} $[1.011, 0.938, 0.416]$\\$[0.291, 0.158, 1.659]$ \end{tabular} &            959 & 8.06 \\ \hline
D & \begin{tabular}{c} $[0.293, 0.264, 0.127]$\\$[0.288, 0.269, 0.422]$ \end{tabular} &  \begin{tabular}{c} $[1.206, 1.248, 0.895]$\\$[1.895, 1.797, 1.070]$ \end{tabular} &           1196 &  10.01 \\ \hline
WOH &     \begin{tabular}{c} $[0.275, 0.255, 0.086]$\\$[0.000,  0.000,  0.433]$ \end{tabular} &     \begin{tabular}{c} $[0.949, 1.025, 0.553]$\\$[0.000,  0.000,  0.725]$ \end{tabular} &           1181 & 11.33 \\ \hline
WO &  \begin{tabular}{c} $[0.224, 0.249, 0.103]$\\$[0.214, 0.220, 0.416]$ \end{tabular} & \begin{tabular}{c} $[1.011, 1.059, 0.634]$\\$[1.892, 1.564, 1.278]$ \end{tabular} &           1763 &   17.94 \\ \hline
SH &     \begin{tabular}{c} $[0.304, 0.254, 0.047]$\\$[0.000,  0.000,  0.426]$ \end{tabular} &     \begin{tabular}{c} $[0.965, 0.908, 0.273]$\\$[0.000,  0.000,  1.018]$ \end{tabular} &              0 &   --- \\ \hline
S & \begin{tabular}{c} $[0.259, 0.236, 0.114]$\\$[0.213, 0.228, 0.435]$ \end{tabular} & \begin{tabular}{c} $[0.915, 0.822, 0.538]$\\$[1.592, 1.571, 1.013]$ \end{tabular} & 0 &    --- \\
\end{tabular}}
\caption{Statistics of the selected sequences. Average speed is expressed in m/s and rad/s. Average acceleration in m/$\textrm{s}^2$ and rad/$\textrm{s}^2$.}
\label{tab:stats}
\end{table}

\begin{table}[!ht]
    \centering
    \resizebox{0.7\columnwidth}{!}{
\begin{tabular}{l|l|cc}
  Dataset & Description & Train imgs. & Val imgs. \\ \hline
 CH & COCO's images containing humans & 64115 & 2693 \\
 S-CH & COCO random subset containing humans & 1256 & 120 \\
 TH & TUM-RGBD \textit{fr3} walking sequences & --- & 3580 \\
 S-GRADE & Subset of our synthetic data & 16K & 2K \\
 A-GRADE & Our full synthetic data & 473K & 118K \\ \hline
 $\ast$ -EX & \multicolumn{3}{c}{Model saved at Epoch X} \\
 BASELINE & \multicolumn{3}{c}{Official model weights trained on COCO} \\
\end{tabular}}
\caption{Humans datasets legend.}
    \label{tab:str_dataset_humans}
\end{table}

\subsection*{Limitations}
\label{sec:issues}
Semantic segmentation of Front3D assets is notoriously faulty in some instances~\cite{front3d-wrong}. We modified the \textit{blenderproc} loader to have better mappings by assigning additional labels to each object. Still, we recognize that this may not be sufficient. However, humans and flying objects are always labelled correctly.

While GRADE is designed to be modular and easily configurable, visual realism is something that can be improved on our released data. This includes both the quality of the assets/textures (for environments and humans) and the variability of body shapes and movements. Moreover, rooms show only occidental-like designs which are always tidy, squared, and uncluttered. Placement should also be expanded with more advanced strategies, e.g. by considering semantic information to avoid humans sitting on plain air. Furthermore, variability in the camera sensor settings and more diverse camera movements should be used. Considering the code modularity, it would not be hard to include human or robot environment interactions or introduce animated pets (e.g. dogs, cats) and multi-level houses. The inclusion of motion generation software such as~\cite{samp} is also a direction that could be considered to have more realistic human motions.

Notably, we encountered various issues using Isaac Sim. The three main ones are: i) the ROS clock is sometimes not published at the right instant~\cite{clock-issue}, ii) dynamic meshes not fully supported~\cite{3d-not-update,missing-colliders}, iii) rendering not done in one step, iv) the 2D occupancy map might be partially incorrect. The first one causes the subsequent published ROS messages to have the wrong timestamp, especially for messages published at the same rate as the clock time. The re-indexing procedure we implemented uses the message index number and the message rate to correct timestamps. Notably, one can briefly pause the script each time the clock is published to solve this issue. While this strategy is preferable when using the method online for robotic testing, it lowers the simulation real-time factor. The second one causes issues with 3D bounding boxes and poses, which can be both corrected offline using our code, and with animations' collision elements that are not created correctly due to limitations of the PhysX software. The newly introduced RTX LiDAR, while not solving the collision of such assets with the environment, allows the detection of such animations with LRF and LiDAR sensors. Since Isaac Sim uses a raycasting technique to generate the 2D map from a given point, sometimes it can happen that this results in partial 2D maps. This can be solved quickly by re-generating it if necessary.

Finally, due to an overlooked scaling factor, the physics behavior of the drone is more comparable to a handheld camera device rather than showing smooth state transitions, with spikes in the acceleration components when movement changes. However, data is still physically plausible and usable.

\end{document}